\documentclass[10pt,twocolumn,letterpaper]{article}

\usepackage{cvpr}              %

\usepackage[T1]{fontenc}

\usepackage{amsthm}
\usepackage{capt-of}  %
\usepackage{tikz}
\usetikzlibrary{positioning}

\DeclareRobustCommand{\anchortext}{{%
  \font\anchorfont="[DejaVuSans.ttf]" at 1.25em\relax\anchorfont ⚓%
}}

\newtheorem{proposition}{Proposition}

\theoremstyle{remark}
\newtheorem{remark}{Remark}

\newcommand{\defres}[2]{\expandafter\def\csname res@#1\endcsname{#2}}
\newcommand{\res}[1]{%
  \ifcsname res@#1\endcsname\csname res@#1\endcsname
  \else\PackageError{numbers}{Result `#1' is not defined in generated/numbers.tex}{}\fi}

\definecolor{cvprblue}{rgb}{0.21,0.49,0.74}
\usepackage[pagebackref,breaklinks,colorlinks,allcolors=cvprblue]{hyperref}
\usepackage[capitalize]{cleveref}

\defres{tests.ct.images}{1800}
\defres{tests.ct.anomalous}{1080}
\defres{tests.ct.free}{720}
\defres{tests.ct.patients}{45}
\defres{tests.ct.types}{12}
\defres{intro.amplitude.base}{25--41\%}
\defres{intro.amplitude.anchored}{61--83\%}
\defres{intro.fa.base}{63\%}
\defres{ct.recovery.base}{61--85\%}
\defres{ct.recovery.anchored}{78--88\%}
\defres{intro.fa.anchored}{2\%}
\defres{ct.cov0.base}{66--92\%}
\defres{ct.cov.base}{61--85\%}
\defres{ct.cov.anchored}{83--97\%}
\defres{ct.pairs}{6}
\defres{ct.pairs.mismatch}{0}
\defres{ct.auroc.task}{10 to 12}
\defres{ct.auroc.zero}{$-$1 to $+$1}
\defres{ct.det5.task}{24}
\defres{ct.det5.zero}{$-$4 to $+$2}
\defres{ct.scale.zero}{1.4--5.7}
\defres{ct.scale.anchored}{0.88--1.03}
\defres{ct.fa.anchored.but.ddnm}{1--3\%}
\defres{ct.f1.better}{6}
\defres{ct.f1.worse}{none}
\defres{ct.fa.better}{6}
\defres{ct.fa.worse}{none}
\defres{ct.cov.better}{5}
\defres{ct.cov.worse}{1}
\defres{ct.amplitude.better}{6}
\defres{ct.amplitude.worse}{none}
\defres{ct.rmse.better}{2}
\defres{ct.rmse.worse}{4}
\defres{ct.auroc.better}{2}
\defres{ct.auroc.worse}{none}
\defres{ct.det5.better}{2}
\defres{ct.det5.worse}{1}
\defres{ct.teaser.only_anchored_detects}{97}
\defres{ct.teaser.only_base_detects}{35}
\defres{ct.teaser.only_base_flags}{341}
\defres{ct.teaser.only_anchored_flags}{10}
\defres{ct.kappa.64}{10}
\defres{ct.nodraw.fa.64}{99\%}
\defres{ct.ddnm.fa.64}{2\%}
\defres{ct.anchorflow.f1.64}{$-$0.8}
\defres{ct.anchorflow.f1ci.64}{[$-$2.5, $+$0.9]}
\defres{ct.observer.det5.64}{76\%}
\defres{ct.observer.fa.64}{0.8\%}
\defres{ct.gp.det5.64}{55\%}
\defres{ct.gp.fa.64}{2.8\%}
\defres{ct.gp.exact.det5.64}{61\%}
\defres{ct.gp.exact.fa.64}{9.7\%}
\defres{ct.fbp.det5.64}{38\%}
\defres{ct.fbp.fa.64}{90.0\%}
\defres{ct.det5.anchored.64}{56--59\%}
\defres{ct.lock.fa.64}{99--100\%}
\defres{ct.lock.det5.64}{47--55\%}
\defres{ct.lockknown.fa.64}{4--100\%}
\defres{ct.lockknown.det5.64}{47--67\%}
\defres{ct.known.fa.64}{16--51\%}
\defres{ct.known.det5.64}{56--63\%}
\defres{ct.quality.gp.64}{78}
\defres{ct.quality.anchorflow.64}{49}
\defres{ct.quality.task.64}{47--49}
\defres{ct.quality.ddrm.base.64}{56--73}
\defres{ct.quality.ddrm.anchored.64}{110--137}
\defres{ct.sigma.est.64}{1.048}
\defres{ct.sigma.est.range.64}{0.97--1.13}
\defres{ct.problem.rows.64}{20925}
\defres{ct.problem.rank.64}{20527}
\defres{ct.seconds.anchorflow.64}{1.6}
\defres{ct.seconds.trained.64}{1.6--2.3}
\defres{ct.seconds.zero.64}{3.5--36}
\defres{ct.seconds.ratio.64}{1.01--1.05}
\defres{ct.draw.det5}{$-$10 to $+$8}
\defres{ct.draw.det5.pairs}{6}
\defres{ct.draw.det5.up}{1}
\defres{ct.draw.det5.down}{4}
\defres{ct.draw.det5.task}{$-$10 to $-$8}
\defres{ct.draw.det5.task.pairs}{3}
\defres{ct.draw.det5.task.up}{none}
\defres{ct.draw.det5.task.down}{3}
\defres{ct.draw.det5.zero}{$-$7 to $+$8}
\defres{ct.draw.det5.zero.pairs}{3}
\defres{ct.draw.det5.zero.up}{1}
\defres{ct.draw.det5.zero.down}{1}
\defres{ct.draw.fa}{$-$98 to $-$2}
\defres{ct.draw.fa.pairs}{6}
\defres{ct.draw.fa.up}{none}
\defres{ct.draw.fa.down}{6}
\defres{ct.draw.fa.task}{$-$4 to $-$2}
\defres{ct.draw.fa.task.pairs}{3}
\defres{ct.draw.fa.task.up}{none}
\defres{ct.draw.fa.task.down}{3}
\defres{ct.draw.fa.zero}{$-$98 to $-$36}
\defres{ct.draw.fa.zero.pairs}{3}
\defres{ct.draw.fa.zero.up}{none}
\defres{ct.draw.fa.zero.down}{3}
\defres{ct.lockknown.fa.task.64}{4--7\%}
\defres{ct.lockknown.fa.zero.64}{37--100\%}
\defres{ct.anchorflow.vs.gp.64}{$+$2.1}
\defres{ct.anchorflow.vs.gp.ci.64}{[$-$0.1, $+$5.5]}
\defres{ct.prop.samplers}{7}
\defres{ct.prop.quantile}{2.0\%}
\defres{ct.two.fa.task.base}{0\%}
\defres{ct.two.fa.task.anchored}{0--1\%}
\defres{ct.two.det5.task}{$-$0.1 to $+$0.1}
\defres{ct.two.f1.task}{3 to 5}
\defres{ct.two.fa.zero.base}{27--100\%}
\defres{ct.two.fa.zero.anchored}{0\%}
\defres{ct.two.det5.zero}{$-$3.3 to $-$1.6}
\defres{ct.two.f1.zero}{5 to 19}
\defres{ct.two.fa.anchored.0.5mm}{0--1\%}
\defres{ct.two.fa.base.0.5mm}{0--100\%}
\defres{ct.shift.fa.anchored.0.5mm}{2--4\%}
\defres{ct.two.fa.anchored.1mm}{4--7\%}
\defres{ct.two.fa.base.1mm}{6--100\%}
\defres{ct.shift.fa.anchored.1mm}{14--22\%}
\defres{ct.two.fa.anchored.2mm}{40--53\%}
\defres{ct.two.fa.base.2mm}{64--100\%}
\defres{ct.shift.fa.anchored.2mm}{72--84\%}
\defres{ct.two.gp.det5}{52\%}
\defres{ct.two.observer.det5}{69\%}
\defres{ct.two.anchorflow.det5}{55\%}
\defres{ct.two.fbp.det5}{51\%}
\defres{ct.two.anchorflow.vs.gp}{$+$2.6}
\defres{ct.two.anchorflow.vs.gp.ci}{[$-$0.7, $+$5.6]}
\defres{ct.recal.fa.base}{3--7\%}
\defres{ct.recal.scale.base}{1.3--5.4}
\defres{ct.recal.scale.anchored}{0.84--0.93}
\defres{ct.recal.vs.task}{22 to 23}
\defres{ct.recal.vs.task.pairs}{2}
\defres{ct.recal.vs.task.sig}{2}
\defres{ct.recal.vs.task.up}{2}
\defres{ct.recal.vs.task.down}{none}
\defres{ct.recal.vs.zero}{$-$2 to $+$2}
\defres{ct.recal.vs.zero.pairs}{4}
\defres{ct.recal.vs.zero.sig}{1}
\defres{ct.recal.vs.zero.up}{none}
\defres{ct.recal.vs.zero.down}{1}
\defres{ct.recal.vs.all}{$-$2 to $+$23}
\defres{ct.recal.vs.all.pairs}{6}
\defres{ct.recal.vs.all.sig}{3}
\defres{ct.recal.vs.all.up}{2}
\defres{ct.recal.vs.all.down}{1}
\defres{ct.held.base}{20--93\%}
\defres{ct.held.anchored}{90\%}
\defres{ct.fheld.median}{71\%}
\defres{ct.aupro.task}{6 to 7}
\defres{ct.aupro.zero}{0 to $+$1}
\defres{ct.pixel_auroc.task}{2}
\defres{ct.pixel_auroc.zero}{0 to $+$1}
\defres{ct.dice_max.task}{1}
\defres{ct.dice_max.zero}{1}

\defres{tests.ct128.images}{2880}
\defres{tests.ct128.anomalous}{2160}
\defres{tests.ct128.free}{720}
\defres{tests.ct128.patients}{45}
\defres{tests.ct128.types}{12}
\defres{ct128.intro.amplitude.base}{26--50\%}
\defres{ct128.intro.amplitude.anchored}{63--89\%}
\defres{ct128.intro.fa.base}{75\%}
\defres{ct128.recovery.base}{43--87\%}
\defres{ct128.recovery.anchored}{71--94\%}
\defres{ct128.intro.fa.anchored}{2\%}
\defres{ct128.cov0.base}{52--94\%}
\defres{ct128.cov.base}{50--84\%}
\defres{ct128.cov.anchored}{81--96\%}
\defres{ct128.pairs}{15}
\defres{ct128.pairs.mismatch}{0}
\defres{ct128.auroc.task}{5 to 21}
\defres{ct128.auroc.zero}{$-$1 to $+$3}
\defres{ct128.det5.task}{17 to 45}
\defres{ct128.det5.zero}{$-$2 to $+$8}
\defres{ct128.scale.zero}{0.9--8.1}
\defres{ct128.scale.anchored}{0.83--1.15}
\defres{ct128.fa.anchored.but.ddnm}{1--6\%}
\defres{ct128.f1.better}{15}
\defres{ct128.f1.worse}{none}
\defres{ct128.fa.better}{13}
\defres{ct128.fa.worse}{none}
\defres{ct128.cov.better}{13}
\defres{ct128.cov.worse}{2}
\defres{ct128.amplitude.better}{15}
\defres{ct128.amplitude.worse}{none}
\defres{ct128.rmse.better}{5}
\defres{ct128.rmse.worse}{10}
\defres{ct128.auroc.better}{7}
\defres{ct128.auroc.worse}{none}
\defres{ct128.det5.better}{9}
\defres{ct128.det5.worse}{none}
\defres{ct128.teaser.only_anchored_detects}{167}
\defres{ct128.teaser.only_base_detects}{104}
\defres{ct128.teaser.only_base_flags}{476}
\defres{ct128.teaser.only_anchored_flags}{9}
\defres{ct128.kappa.64}{10}
\defres{ct128.nodraw.fa.64}{100\%}
\defres{ct128.posterior.f1.64}{$-$0.5}
\defres{ct128.posterior.f1ci.64}{[$-$2.6, $+$1.5]}
\defres{ct128.ddnm.fa.64}{2\%}
\defres{ct128.anchorflow.f1.64}{$+$0.7}
\defres{ct128.anchorflow.f1ci.64}{[$-$1.0, $+$2.2]}
\defres{ct128.kappa.256}{10}
\defres{ct128.nodraw.fa.256}{100\%}
\defres{ct128.posterior.f1.256}{$+$0.3}
\defres{ct128.posterior.f1ci.256}{[$-$2.1, $+$2.7]}
\defres{ct128.ddnm.fa.256}{4\%}
\defres{ct128.anchorflow.f1.256}{$+$1.7}
\defres{ct128.anchorflow.f1ci.256}{[$-$1.3, $+$4.7]}
\defres{ct128.observer.det5.64}{72\%}
\defres{ct128.observer.fa.64}{1.9\%}
\defres{ct128.gp.det5.64}{49\%}
\defres{ct128.gp.fa.64}{2.5\%}
\defres{ct128.gp.exact.det5.64}{59\%}
\defres{ct128.gp.exact.fa.64}{5.0\%}
\defres{ct128.fbp.det5.64}{22\%}
\defres{ct128.fbp.fa.64}{100.0\%}
\defres{ct128.det5.anchored.64}{48--53\%}
\defres{ct128.lock.fa.64}{100\%}
\defres{ct128.lock.det5.64}{39--51\%}
\defres{ct128.lockknown.fa.64}{9--100\%}
\defres{ct128.lockknown.det5.64}{33--59\%}
\defres{ct128.known.fa.64}{15--42\%}
\defres{ct128.known.det5.64}{47--57\%}
\defres{ct128.quality.gp.64}{72}
\defres{ct128.quality.anchorflow.64}{46}
\defres{ct128.quality.task.64}{44--47}
\defres{ct128.quality.ddrm.base.64}{77--108}
\defres{ct128.quality.ddrm.anchored.64}{146--189}
\defres{ct128.observer.det5.256}{86\%}
\defres{ct128.observer.fa.256}{2.2\%}
\defres{ct128.gp.det5.256}{73\%}
\defres{ct128.gp.fa.256}{4.4\%}
\defres{ct128.gp.exact.det5.256}{78\%}
\defres{ct128.gp.exact.fa.256}{7.2\%}
\defres{ct128.det5.anchored.256}{73--75\%}
\defres{ct128.lock.fa.256}{100\%}
\defres{ct128.lock.det5.256}{57--74\%}
\defres{ct128.lockknown.fa.256}{4--100\%}
\defres{ct128.lockknown.det5.256}{65--79\%}
\defres{ct128.known.fa.256}{4--66\%}
\defres{ct128.known.det5.256}{73--79\%}
\defres{ct128.quality.gp.256}{51}
\defres{ct128.quality.anchorflow.256}{35}
\defres{ct128.quality.task.256}{31--41}
\defres{ct128.quality.ddrm.base.256}{95--127}
\defres{ct128.quality.ddrm.anchored.256}{169--197}
\defres{ct128.sigma.est.64}{1.011}
\defres{ct128.sigma.est.range.64}{0.95--1.08}
\defres{ct128.problem.rows.64}{10485}
\defres{ct128.problem.rank.64}{10163}
\defres{ct128.sigma.est.256}{1.006}
\defres{ct128.sigma.est.range.256}{1.00--1.02}
\defres{ct128.problem.rows.256}{41997}
\defres{ct128.problem.rank.256}{16384}
\defres{ct128.seconds.anchorflow.64}{0.42}
\defres{ct128.seconds.trained.64}{0.43--0.71}
\defres{ct128.seconds.zero.64}{0.55--20}
\defres{ct128.seconds.ratio.64}{1.01--1.04}
\defres{ct128.seconds.anchorflow.256}{0.44}
\defres{ct128.seconds.trained.256}{0.45--0.71}
\defres{ct128.seconds.zero.256}{0.56--11}
\defres{ct128.seconds.ratio.256}{1.01--1.11}
\defres{ct128.train.anchorflow.64}{0.059}
\defres{ct128.train.anchorflow.256}{0.027}
\defres{ct128.train.cfm.64}{0.058}
\defres{ct128.train.cfm.256}{0.063}
\defres{ct128.train.edm.64}{0.102}
\defres{ct128.train.edm.256}{0.100}
\defres{ct128.draw.det5}{$-$11 to $+$15}
\defres{ct128.draw.det5.pairs}{12}
\defres{ct128.draw.det5.up}{4}
\defres{ct128.draw.det5.down}{8}
\defres{ct128.draw.det5.task}{$-$11 to $-$3}
\defres{ct128.draw.det5.task.pairs}{6}
\defres{ct128.draw.det5.task.up}{none}
\defres{ct128.draw.det5.task.down}{6}
\defres{ct128.draw.det5.zero}{$-$4 to $+$15}
\defres{ct128.draw.det5.zero.pairs}{6}
\defres{ct128.draw.det5.zero.up}{4}
\defres{ct128.draw.det5.zero.down}{2}
\defres{ct128.draw.fa}{$-$98 to $+$1}
\defres{ct128.draw.fa.pairs}{12}
\defres{ct128.draw.fa.up}{none}
\defres{ct128.draw.fa.down}{11}
\defres{ct128.draw.fa.task}{$-$7 to $+$1}
\defres{ct128.draw.fa.task.pairs}{6}
\defres{ct128.draw.fa.task.up}{none}
\defres{ct128.draw.fa.task.down}{5}
\defres{ct128.draw.fa.zero}{$-$98 to $-$40}
\defres{ct128.draw.fa.zero.pairs}{6}
\defres{ct128.draw.fa.zero.up}{none}
\defres{ct128.draw.fa.zero.down}{6}
\defres{ct128.lockknown.fa.task.64}{9--10\%}
\defres{ct128.lockknown.fa.zero.64}{41--100\%}
\defres{ct128.lockknown.fa.task.256}{4--9\%}
\defres{ct128.lockknown.fa.zero.256}{50--100\%}
\defres{ct128.anchorflow.vs.gp.64}{$+$1.9}
\defres{ct128.anchorflow.vs.gp.ci.64}{[$-$1.1, $+$4.4]}
\defres{ct128.anchorflow.vs.gp.256}{$+$0.4}
\defres{ct128.anchorflow.vs.gp.ci.256}{[$-$0.7, $+$1.8]}
\defres{ct128.readings.shared.fa}{28--100\%}
\defres{ct128.readings.own.fa}{0--8\%}
\defres{ct128.ddnm.ratio}{5--17}
\defres{ct128.ddnm.ratio.own}{0.99--1.11}
\defres{ct128.prop.samplers}{15}
\defres{ct128.prop.quantile}{1.7\%}
\defres{ct128.k16.det5}{48--50\%}
\defres{ct128.k64.det5}{55--59\%}
\defres{ct128.two.fa.task.base}{0\%}
\defres{ct128.two.fa.task.anchored}{0--1\%}
\defres{ct128.two.det5.task}{2.4 to 16.9}
\defres{ct128.two.f1.task}{3 to 27}
\defres{ct128.two.fa.zero.base}{35--100\%}
\defres{ct128.two.fa.zero.anchored}{0--4\%}
\defres{ct128.two.det5.zero}{$-$1.8 to $+$2.7}
\defres{ct128.two.f1.zero}{8 to 15}
\defres{ct128.two.fa.anchored.0.5mm}{0--4\%}
\defres{ct128.two.fa.base.0.5mm}{0--100\%}
\defres{ct128.shift.fa.anchored.0.5mm}{2--7\%}
\defres{ct128.two.fa.anchored.1mm}{1--8\%}
\defres{ct128.two.fa.base.1mm}{0--100\%}
\defres{ct128.shift.fa.anchored.1mm}{4--14\%}
\defres{ct128.two.fa.anchored.2mm}{17--36\%}
\defres{ct128.two.fa.base.2mm}{5--100\%}
\defres{ct128.shift.fa.anchored.2mm}{34--70\%}
\defres{ct128.two.gp.det5}{46\%}
\defres{ct128.two.observer.det5}{59\%}
\defres{ct128.two.anchorflow.det5}{48\%}
\defres{ct128.two.fbp.det5}{46\%}
\defres{ct128.two.anchorflow.vs.gp}{$+$1.7}
\defres{ct128.two.anchorflow.vs.gp.ci}{[$+$0.3, $+$4.1]}
\defres{ct128.recal.fa.base}{1--9\%}
\defres{ct128.recal.scale.base}{0.9--7.7}
\defres{ct128.recal.scale.anchored}{0.95--0.98}
\defres{ct128.recal.vs.task}{13 to 43}
\defres{ct128.recal.vs.task.pairs}{6}
\defres{ct128.recal.vs.task.sig}{6}
\defres{ct128.recal.vs.task.up}{6}
\defres{ct128.recal.vs.task.down}{none}
\defres{ct128.recal.vs.zero}{$-$3 to $+$9}
\defres{ct128.recal.vs.zero.pairs}{9}
\defres{ct128.recal.vs.zero.sig}{4}
\defres{ct128.recal.vs.zero.up}{2}
\defres{ct128.recal.vs.zero.down}{2}
\defres{ct128.recal.vs.all}{$-$3 to $+$43}
\defres{ct128.recal.vs.all.pairs}{15}
\defres{ct128.recal.vs.all.sig}{10}
\defres{ct128.recal.vs.all.up}{8}
\defres{ct128.recal.vs.all.down}{2}
\defres{ct128.sigma.samplers}{4}
\defres{ct128.sigma.low}{0.8}
\defres{ct128.sigma.high}{1.2}
\defres{ct128.sigma.fa.stated}{1.9--2.4\%}
\defres{ct128.sigma.det5.max}{1.5}
\defres{ct128.sigma.fa.low}{13.1--18.2\%}
\defres{ct128.sigma.f1.low}{$-$4 to $-$1}
\defres{ct128.sigma.fa.high}{0.0--0.4\%}
\defres{ct128.sigma.f1.high}{$-$6 to $-$5}
\defres{ct128.f1.pairs}{15}
\defres{ct128.f1.pairs.mismatch}{8}
\defres{ct128.held.base}{17--93\%}
\defres{ct128.held.anchored}{87--89\%}
\defres{ct128.fheld.median}{66--90\%}
\defres{ct128.aupro.task}{4 to 17}
\defres{ct128.aupro.zero}{$-$1 to $+$5}
\defres{ct128.pixel_auroc.task}{1 to 7}
\defres{ct128.pixel_auroc.zero}{0 to $+$1}
\defres{ct128.dice_max.task}{1 to 3}
\defres{ct128.dice_max.zero}{0 to 2}

\defres{tests.brain.images}{1362}
\defres{tests.brain.anomalous}{661}
\defres{tests.brain.free}{701}
\defres{tests.brain.patients}{60}
\defres{tests.brain.types}{4}
\defres{intro.brain.pairs}{16}
\defres{brain.rmse.gain}{1.6}
\defres{intro.brain.better}{11}
\defres{intro.brain.worse}{none}
\defres{brain.f1.zero.pairs}{12}
\defres{brain.f1.zero.better}{12}
\defres{brain.fa.task.base}{15--49\%}
\defres{brain.fa.task.anchored}{14--51\%}
\defres{brain.fa.zero.base}{16--100\%}
\defres{brain.fa.zero.anchored}{7--71\%}
\defres{brain.f1.better}{14}
\defres{brain.f1.worse}{none}
\defres{brain.fa.better}{13}
\defres{brain.fa.worse}{none}
\defres{brain.cov.better}{13}
\defres{brain.cov.worse}{3}
\defres{brain.rmse.better}{11}
\defres{brain.rmse.worse}{none}
\defres{brain.auroc.better}{12}
\defres{brain.auroc.worse}{none}
\defres{brain.finder.f1}{$-$2.3 to $+$1.6}
\defres{brain.finder.f1.sig}{6}
\defres{brain.finder.pairs}{16}
\defres{brain.readings.f1}{$-$0.7 to $+$0.9}
\defres{brain.readings.f1.sig}{none}
\defres{brain.readings.pairs}{16}
\defres{brain.recal.fa.base}{2--8\%}
\defres{brain.recal.scale.base}{1.1--3.4}
\defres{brain.recal.scale.anchored}{1.01--2.88}
\defres{brain.recal.vs.task}{$-$2 to $+$11}
\defres{brain.recal.vs.task.pairs}{6}
\defres{brain.recal.vs.task.sig}{3}
\defres{brain.recal.vs.task.up}{3}
\defres{brain.recal.vs.task.down}{none}
\defres{brain.recal.vs.zero}{$-$2 to $+$31}
\defres{brain.recal.vs.zero.pairs}{10}
\defres{brain.recal.vs.zero.sig}{5}
\defres{brain.recal.vs.zero.up}{5}
\defres{brain.recal.vs.zero.down}{none}
\defres{brain.recal.vs.all}{$-$2 to $+$31}
\defres{brain.recal.vs.all.pairs}{16}
\defres{brain.recal.vs.all.sig}{8}
\defres{brain.recal.vs.all.up}{8}
\defres{brain.recal.vs.all.down}{none}
\defres{brain.held.base}{31--91\%}
\defres{brain.held.anchored}{90\%}
\defres{brain.fheld.median}{76--87\%}
\defres{brain.aupro.task}{$-$3 to $+$10}
\defres{brain.aupro.zero}{6 to 29}
\defres{brain.pixel_auroc.task}{$-$3 to $+$7}
\defres{brain.pixel_auroc.zero}{4 to 26}
\defres{brain.dice_max.task}{$-$2 to $+$2}
\defres{brain.dice_max.zero}{2 to 23}

\defres{tests.pat.images}{156}
\defres{tests.pat.anomalous}{78}
\defres{tests.pat.free}{78}
\defres{tests.pat.patients}{27}
\defres{tests.pat.types}{2}
\defres{pat.fa.base}{73--100\%}
\defres{pat.fa.anchored}{3--5\%}
\defres{pat.f1.gain}{52 to 88}
\defres{pat.rmse.base}{49.4--52.1}
\defres{pat.rmse.anchored}{24.3--25.6}
\defres{pat.rmse.worse}{none}
\defres{pat.cov.base}{22--37\%}
\defres{pat.cov.anchored}{65--69\%}
\defres{pat.pairs}{3}
\defres{pat.f1.better}{3}
\defres{pat.rmse.better}{2}
\defres{pat.recal.fa.base}{1--15\%}
\defres{pat.recal.scale.base}{2.4--6.6}
\defres{pat.recal.scale.anchored}{0.90--1.05}
\defres{pat.recal.vs.task}{65 to 92}
\defres{pat.recal.vs.task.pairs}{2}
\defres{pat.recal.vs.task.sig}{2}
\defres{pat.recal.vs.task.up}{2}
\defres{pat.recal.vs.task.down}{none}
\defres{pat.recal.vs.zero}{$-$3}
\defres{pat.recal.vs.zero.pairs}{1}
\defres{pat.recal.vs.zero.sig}{1}
\defres{pat.recal.vs.zero.up}{none}
\defres{pat.recal.vs.zero.down}{1}
\defres{pat.recal.vs.all}{$-$3 to $+$92}
\defres{pat.recal.vs.all.pairs}{3}
\defres{pat.recal.vs.all.sig}{3}
\defres{pat.recal.vs.all.up}{2}
\defres{pat.recal.vs.all.down}{1}
\defres{pat.held.base}{39--90\%}
\defres{pat.held.anchored}{90\%}
\defres{pat.fheld.median}{82\%}
\defres{pat.aupro.task}{3 to 68}
\defres{pat.aupro.zero}{0}
\defres{pat.pixel_auroc.task}{1 to 30}
\defres{pat.pixel_auroc.zero}{0}
\defres{pat.dice_max.task}{$-$1 to $+$9}
\defres{pat.dice_max.zero}{0}

\defres{fig.teaser.readings}{64 views}
\defres{fig.teaser.method}{conditional FM}

\defres{fig.teaser.lesion.base}{0.98}
\defres{fig.teaser.lesion.anchored}{1.12}
\defres{fig.teaser.free.base}{1.18}
\defres{fig.teaser.free.anchored}{0.86}
\defres{fig.teaser.free.image}{2072}
\defres{fig.teaser.free.disagreements}{341}
\defres{fig.teaser.free.interior}{77}
\defres{fig.teaser.free.margin}{64}

\defres{tests.pat128.images}{156}
\defres{tests.pat128.anomalous}{78}
\defres{tests.pat128.free}{78}
\defres{tests.pat128.patients}{27}
\defres{tests.pat128.types}{2}
\defres{pat128.recal.fa.base}{6--10\%}
\defres{pat128.recal.scale.base}{2.9--4.8}
\defres{pat128.recal.scale.anchored}{0.86--1.09}
\defres{pat128.recal.vs.task}{45 to 85}
\defres{pat128.recal.vs.task.pairs}{2}
\defres{pat128.recal.vs.task.sig}{2}
\defres{pat128.recal.vs.task.up}{2}
\defres{pat128.recal.vs.task.down}{none}
\defres{pat128.recal.vs.zero}{$-$2}
\defres{pat128.recal.vs.zero.pairs}{1}
\defres{pat128.recal.vs.zero.sig}{none}
\defres{pat128.recal.vs.zero.up}{none}
\defres{pat128.recal.vs.zero.down}{none}
\defres{pat128.recal.vs.all}{$-$2 to $+$85}
\defres{pat128.recal.vs.all.pairs}{3}
\defres{pat128.recal.vs.all.sig}{2}
\defres{pat128.recal.vs.all.up}{2}
\defres{pat128.recal.vs.all.down}{none}
\defres{pat128.held.base}{50--92\%}
\defres{pat128.held.anchored}{89\%}
\defres{pat128.fheld.median}{83\%}
\defres{pat128.aupro.task}{3 to 50}
\defres{pat128.aupro.zero}{0}
\defres{pat128.pixel_auroc.task}{1 to 20}
\defres{pat128.pixel_auroc.zero}{0}
\defres{pat128.dice_max.task}{$-$1 to $+$8}
\defres{pat128.dice_max.zero}{0}
\defres{summary.brain.fa.base}{96\%}
\defres{summary.brain.fa.recal}{5\%}
\defres{summary.brain.fa.anchored}{13\%}
\defres{summary.brain.det.base}{100\%}
\defres{summary.brain.det.anchored}{100\%}
\defres{summary.pat.det.base}{20\%}
\defres{summary.pat.det.anchored}{98\%}

\defres{brain.prior.exposure}{66\%}
\defres{brain.prior.ddrm.free}{5.9 points [4.8, 7.1]}
\defres{brain.prior.ddrm.exposed}{2.4 points [1.8, 3.1]}
\defres{brain.noise.ddrm.growth}{8.7 points [7.4, 9.9]}
\defres{brain.nfe.anchorflow.low}{5}
\defres{brain.nfe.anchorflow.full}{100}
\defres{brain.nfe.anchorflow.rmse.low}{12.54\%}
\defres{brain.nfe.anchorflow.rmse.full}{12.38\%}
\defres{ct.efficiency.f1.anchorflow}{56.0}
\defres{ct.efficiency.f1.cfm}{25.8}
\defres{ct.efficiency.f1.anchored_cfm}{56.8}
\defres{ct.efficiency.recovery.anchorflow}{79\%}
\defres{ct.efficiency.recovery.ddnm}{88\%}
\defres{ct.runtime.images}{432}
\defres{ct.runtime.patients}{45}
\defres{ct.runtime.anchorflow.nfe.ultra}{1}
\defres{ct.runtime.anchorflow.seconds.ultra}{5.9\,ms}
\defres{ct.runtime.anchorflow.rmse.ultra}{65.5\,HU}
\defres{ct.runtime.anchorflow.contrast-error.ultra}{26.8\%}
\defres{ct.runtime.anchorflow.coverage-error.ultra}{9.4 points}
\defres{ct.runtime.anchorflow.interval.ultra}{321.7\,HU}
\defres{ct.runtime.anchorflow.crps.ultra}{37.7\,HU}
\defres{ct.runtime.cfm.seconds.ultra}{5.1\,ms}
\defres{ct.runtime.cfm.rmse.ultra}{59.3\,HU}
\defres{ct.runtime.cfm.crps.ultra}{35.5\,HU}
\defres{ct.runtime.cfm.seconds.full}{0.436\,s}
\defres{ct.runtime.cfm.rmse.full}{60.1\,HU}
\defres{ct.runtime.cfm.contrast-error.full}{27.9\%}
\defres{ct.runtime.cfm.coverage-error.full}{2.5 points}
\defres{ct.runtime.cfm.interval.full}{280.7\,HU}
\defres{ct.runtime.cfm.crps.full}{33.7\,HU}
\defres{ct.runtime.anchorflow.vs-cfm-full.speedup}{$74\times$}
\defres{ct.runtime.anchorflow.vs-cfm-full.rmse-gap}{5.4\,HU}
\defres{ct.runtime.anchorflow.vs-cfm-full.contrast-gain}{1.1 points}
\defres{ct.runtime.anchorflow.vs-cfm-full.coverage-gap}{6.9 points}
\defres{ct.runtime.anchorflow.vs-cfm-full.interval-gap}{41.0\,HU}
\defres{ct.runtime.anchorflow.vs-cfm-full.crps-gap}{4.0\,HU}
\defres{ct.runtime.anchorflow.vs-best.rmse-relative}{8.9\%}
\defres{ct.runtime.anchorflow.vs-best.interval-relative}{14.6\%}
\defres{ct.runtime.anchorflow.vs-best.crps-relative}{11.8\%}
\defres{ct.runtime.anchorflow.vs-best.contrast-gap}{0.7 points}
\defres{ct.runtime.anchorflow.vs-diffpir-full.speedup}{$996\times$}
\defres{ct.runtime.ddrm.coverage-error.full}{0.04 points}
\defres{ct.runtime.anchorflow.vs-ddrm-full.speedup}{$99\times$}
\defres{ct.runtime.anchorflow.nfe.fast}{10}
\defres{ct.runtime.anchorflow.seconds.fast}{0.046\,s}
\defres{ct.runtime.anchorflow.rmse.fast}{65.8\,HU}
\defres{ct.runtime.anchorflow.contrast.fast}{88.3\%}
\defres{ct.runtime.anchorflow.coverage.fast}{87.5\%}
\defres{ct.runtime.anchorflow.contrast-error.fast}{27.0\%}
\defres{ct.runtime.anchorflow.coverage-error.fast}{2.5 points}
\defres{ct.runtime.anchorflow.speedup.fast}{$9.7\times$}
\defres{ct.runtime.anchorflow.nfe.calibrated}{20}
\defres{ct.runtime.anchorflow.seconds.calibrated}{0.090\,s}
\defres{ct.runtime.anchorflow.rmse.calibrated}{66.5\,HU}
\defres{ct.runtime.anchorflow.contrast.calibrated}{88.4\%}
\defres{ct.runtime.anchorflow.coverage.calibrated}{91.1\%}
\defres{ct.runtime.anchorflow.contrast-error.calibrated}{27.0\%}
\defres{ct.runtime.anchorflow.coverage-error.calibrated}{1.1 points}
\defres{ct.runtime.anchorflow.speedup.calibrated}{$4.9\times$}
\defres{ct.runtime.anchorflow.nfe.full}{100}
\defres{ct.runtime.anchorflow.seconds.full}{0.441\,s}
\defres{ct.runtime.anchorflow.rmse.full}{67.3\,HU}
\defres{ct.runtime.anchorflow.contrast.full}{88.4\%}
\defres{ct.runtime.anchorflow.coverage.full}{94.1\%}
\defres{ct.runtime.anchorflow.contrast-error.full}{27.1\%}
\defres{ct.runtime.anchorflow.coverage-error.full}{4.1 points}
\defres{ct.runtime.cfm.rmse.fast}{59.5\,HU}
\defres{ct.runtime.cfm.contrast.fast}{82.9\%}
\defres{ct.runtime.cfm.coverage.fast}{77.8\%}
\defres{ct.runtime.cfm.contrast-error.fast}{27.9\%}
\defres{ct.runtime.cfm.coverage-error.fast}{12.2 points}
\defres{brain.ref.oracle.gain}{9.1 to 40.8}
\defres{brain.ref.readings.det5}{73\%}
\defres{brain.ref.blind.det5}{63\%}
\defres{brain.ref.readings.truth.det5}{72.5\%}
\defres{brain.ref.blind.truth.det5}{63.1\%}
\defres{brain.ref.blind.contrast}{25\%}

\newcommand{\T}{^{\top}}
\newcommand{\N}{\mathcal{N}}
\newcommand{\E}{\mathbb{E}}
\newcommand{\Var}{\operatorname{Var}}

\newcommand{\lift}{\operatorname{lift}}
\newcommand{\xmap}{\hat{x}}
\newcommand{\Hs}{\mathcal{H}}

\title{Do Not Let the Prior Heal the Anomaly:\\Selective \anchortext\ Anchoring for Detection from Incomplete Measurements}

\author{%
  Nicolas Tricard\textsuperscript{1}\quad
  Stefan Brandstätter\textsuperscript{2,3}\quad
  Jason Chen\textsuperscript{1}\quad
  Zituo Chen\textsuperscript{1}\\
  Linzheng Wang\textsuperscript{1}\quad
  Georg Langs\textsuperscript{3}\quad
  Sili Deng\textsuperscript{1}\\[3pt]
  \textsuperscript{1}Massachusetts Institute of Technology\quad \\
  \textsuperscript{2}Massachusetts General Hospital, Harvard University\quad \\
  \textsuperscript{3}Medical University of Vienna\quad
  \\{\small\texttt{silideng@mit.edu}}
}

\begin{document}
\raggedbottom

\twocolumn[{%
  \renewcommand\twocolumn[1][]{#1}%
  \maketitle
  \centering
  \begin{tikzpicture}

  \node[inner sep=0pt] (fig) {%
    \includegraphics[width=\linewidth]{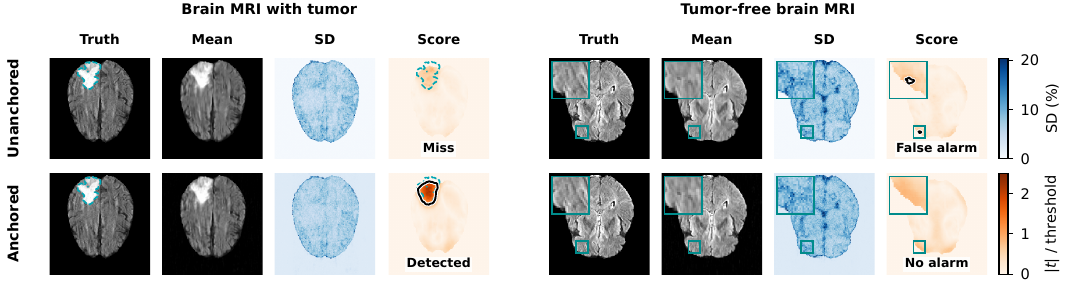}};
    
\end{tikzpicture}

\par\vspace{6pt}

  \captionof{figure}{
  Selective anchoring detects a tumor that unanchored conditional FM misses (left) and
  removes a false alarm (right) on brain MRI from 64 k-space lines. Anchoring preserves the lesion and its uncertainty better. Left: the unanchored score stays below its validation threshold at the tumor, although the tumor is visible in the ensemble mean. With anchoring the same region exceeds the detection score threshold. Right: the unanchored samples agree more closely than the measurement noise warrants,
  so a small deviation appears statistically significant. Columns: truth, ensemble mean, 
  standard deviation (SD, \% of median brain intensity), and Student's t-test score from \cref{eq:t} over its decision threshold. Dashed: expert tumor boundary; solid: score of one; insets enlarge the boxed region.}
\label{fig:teaser}

  \par\vspace{10pt}
}]
\begin{abstract}
Generative priors can recover anatomical detail from incomplete medical measurements, but may erase the very lesions that imaging aims to detect, even when the measurements contain evidence of them.
We introduce \emph{selective anchoring} that protects reliably measured image components from prior-driven changes, while allowing expressive generative models to reconstruct unresolved structure.
Acquisition uncertainty in the protected components is either added deterministically to the downstream test or carried by the ensemble through stochastic perturbations of the readings.
We implement this principle as a wrapper for pretrained diffusion and flow samplers and as \emph{AnchoredFM}, a fast constrained flow model.
Under a linear-Gaussian measurement model, the protected coefficients are unbiased in expectation and achieve nominal interval coverage across repeated acquisitions, independently of the learned prior.
Across sparse-view CT, accelerated brain MRI, and photoacoustic tomography, selective anchoring improves lesion contrast preservation and uncertainty reliability, and can reduce reconstruction error within lesions.
For tumor detection in accelerated MRI, with a healthy reference estimated from the
measurements alone, deterministic anchoring raises mean AUROC across eight samplers
from 78.4\% to 84.6\% and lesion sensitivity at 0.25 false-positive marks per slice
from 38.8\% to 47.5\%, with gains in all four predefined tumor-size groups. At
thresholds frozen on healthy validation data, mean sensitivity rises from 20.2\% to
44.1\% (37.6\% with stochastic anchoring).
These findings show how expressive generative priors can reconstruct unresolved anatomy without healing away the abnormalities that the measurements support.

\end{abstract}
\section{Introduction}
\label{sec:intro}

Sparse-view CT, accelerated MRI, and photoacoustic tomography use fewer measurements to reduce radiation exposure, scan time, or detector requirements in medical imaging.
With incomplete data, direct inversion can amplify noise or produce streaking and aliasing.
Learned generative priors help reconstruct anatomical structures that the measurements cannot fully determine~\cite{chung2023dps,kawar2022ddrm,song2022medical}.
However, when these priors are trained predominantly on normal images, they may favor healthy-looking anatomy and weaken or remove lesions, even when the acquired measurements contain evidence of them~\cite{bhadra2021hallucinations}.

A common anomaly-detection approach compares a patient's reconstructed ensemble mean, $\bar{x}$, obtained from incomplete measurements $y$, with a healthy reference, $r$, representing the same anatomy without pathology~\cite{schlegl2019fanogan,wyatt2022anoddpm,bercea2025normative}.
The subtraction residual is
\begin{equation}
    \Delta x =
    \bar{x} - r.
    \label{eq:subtraction}
\end{equation}
If reconstruction weakens the contrast of a lesion, the difference between the reconstructed image and the healthy reference may be too small to detect it.
A second problem arises when the reconstruction underestimates its uncertainty.
In this case, differences in healthy tissue may appear statistically significant and lead to false alarms.

Incomplete measurements do not constrain every part of an image equally.
Some image components can be estimated reliably from the observations, while others are weakly measured or entirely unobserved.
Existing generative methods were designed for image quality, not for anomaly preservation and detection.
Samplers that guide or condition on the measurements~\cite{chung2023dps,song2022medical,martin2025pnpflow}
let the prior alter reliably measured components, which can weaken an anomaly and cause missed lesions. Samplers that enforce the measurements exactly~\cite{wang2023ddnm,lee2026msc,shi2026nullflow} give every sample the same measured values, so a downstream test reads their agreement as certainty although the readings are noisy, causing false alarms. 
A reconstruction meant for detection should therefore take both the evidence and its noise from the measurements, and leave to the prior only what they do not resolve.

We introduce \textit{selective anchoring} to preserve reliably measured image components while allowing the generative model to reconstruct the remainder.
The protected components correspond to image-space singular-vector directions of the known imaging operator.
We select directions whose measurement uncertainty is small relative to normal-image variability and constrain their coefficients to measurement-derived values.
Weakly measured directions and the measurement null space remain available for generative completion.
We consider two ways of accounting for measurement uncertainty.
\emph{Deterministic anchoring} fixes the protected coefficients at their measurement-derived values and adds their known variance analytically to downstream detection tests.
\emph{Stochastic anchoring} instead assigns each reconstruction sample an independently perturbed target, allowing the ensemble to represent acquisition uncertainty and potentially propagate its effects into the generated components.
For a known linear measurement operator $A$ with additive Gaussian noise, $y=Ax+\varepsilon$ and $\varepsilon\sim\N(0,\sigma^2 I)$, the protected coefficient estimates are unbiased over repeated acquisitions, and their analytic intervals achieve nominal coverage independently of the learned model for the remaining components~(\cref{prop:held}).
We implement selective anchoring as a training-free wrapper for pretrained generative samplers and as AnchoredFM, a flow-matching model that evolves only the unconstrained components~\cite{lipman2023flow}.

We evaluate selective anchoring on inserted CT lesions, real brain tumors under simulated accelerated MRI, and photoacoustic breast phantoms.
Across these tests, anchoring better preserves measurement-supported lesion contrast, improves uncertainty calibration in the protected directions, and often reduces reconstruction error within lesions~(\cref{tab:brain,sec:exp-calibration}).
For primary MRI detection, we construct a healthy reference $r$ by checkerboard inpainting from the measurements and fit thresholds using healthy validation slices.
Across eight matched sampler pairs, anchoring raises mean AUROC from 78.4\% to 84.6\%
and lesion sensitivity at 0.25 false-positive marks per slice from 38.8\% to 47.5\%
(\cref{tab:brain,tab:size,supp:tab:practical-fixed64}). At validation-frozen
thresholds, mean sensitivity rises from 20.2\% to 44.1\%. 
An additional oracle test provides a controlled reference comparison~(\cref{supp:fixed-reference}).
At low sampling budgets, Anchored Flow Matching (AnchoredFM) achieves competitive lesion preservation and uncertainty performance while remaining substantially faster than full-budget generative samplers.
\section{Related work}
\label{sec:related}

\paragraph{Image-based anomaly detection.}
Image-based anomaly detection often compares a test image with a pseudo-healthy reference learned from normal anatomy~\cite{schlegl2019fanogan,baur2021autoencoders}.
Diffusion-based methods generate these references through noising and denoising or guided reconstruction~\cite{wyatt2022anoddpm,pinaya2022fast,wolleb2022diffusion,bercea2024thor}.
Other approaches detect simulated defects, as in DRAEM~\cite{zavrtanik2021draem}, or use pretrained features, as in PatchCore and Dinomaly~\cite{roth2022patchcore,guo2025dinomaly}.
These methods generally assume that the test image is available, whereas tomographic anomaly detection requires reconstructing it from incomplete measurements.

\paragraph{Pathology preservation in tomography.}
The reconstruction step introduces a separate failure mode for downstream detection, as learned priors can suppress clinically relevant features while producing plausible images. Bhadra et al.~\cite{bhadra2021hallucinations} analyzed hallucinations by distinguishing measurement-supported and null-space components, while accelerated MRI studies have shown that rare pathology may be missed despite favorable reconstruction metrics~\cite{cheng2020falsenegative,vanLohuizen2024Assessing,Jensen2022}. To address missed pathology, Morshuis et al.~\cite{morshuis2025mind} generated semantically diverse, measurement-consistent MRI reconstructions. Other studies examine distribution shifts from incomplete measurements~\cite{kim2025distribution,demircan2026ood,sprunck2026bayesian}, including KLIP~\cite{kheirandish2026klip}, which detects localized abnormalities by comparing diffusion prior and posterior distributions. These approaches reveal reconstruction failures or identify abnormalities from the measurements. Our focus is on retaining measurement-supported lesion contrast and its associated uncertainty in the reconstruction samples used for downstream detection.

\paragraph{Measurement consistency and uncertainty.}
Preserving measured information depends on how generative reconstruction incorporates
the observations. Diffusion and flow priors use learned conditioning or
measurement-guided sampling~\cite{song2022medical,chung2023dps,martin2025pnpflow}.
Several methods instead fix the measured components, within each sampling step
(DDNM~\cite{wang2023ddnm}), as a final lock (Measured-Subspace
Consistency~\cite{lee2026msc}), or by confining a flow to the null space
(NullFlow~\cite{shi2026nullflow}). These hold every measured direction, whereas
truncated-SVD learning~\cite{schwab2019tsvd} holds only those with large singular
values. Fixing the measured components limits the prior, but giving every sample the
same values removes their ensemble variation even though the readings are noisy.
Other work handles noise during sampling~\cite{kawar2021snips,dou2024fpssmc}, by
calibration on held-out data~\cite{angelopoulos2022image,teneggi2023trust}, or with
perturbed observations in Kalman and Gaussian posterior
sampling~\cite{burgers1998enkf,bardsley2014rto}. Selective anchoring holds only the
directions whose noise is small relative to normal-image variability and keeps their
acquisition variance, analytically in the test or through perturbed readings. Unlike
DDRM and Posterior-DDIM~\cite{kawar2022ddrm,jiao2026posteriorddim}, which weigh prior
against measurement in each singular direction, we remove the prior from the held
directions, so their coefficients are unbiased with nominal coverage for each fixed
image, not only on average over the prior; both are among our baselines
(\cref{tab:brain}).

\section{Anomaly-preserving reconstruction}
\label{sec:problem}
We seek to reconstruct an image while retaining lesion evidence supported by incomplete measurements. Write $x=b+\ell$, with healthy anatomy $b$ and lesion signal $\ell$, and assume
\begin{equation}
y=Ax+\varepsilon,\qquad \varepsilon\sim\N(0,\sigma^2I_m),
\label{eq:model}
\end{equation}
where $A\in\mathbb R^{m\times d}$ and $\sigma>0$ are known. A prior trained on normal images helps infer anatomy that the measurements do not resolve. Its preference for normal anatomy can also attenuate lesions, even in components measured reliably. Our aim is to protect those components while using the prior to reconstruct the remainder.

The reconstructor returns $K\geq2$ samples $x_j$, whose mean $\bar x=K^{-1}\sum_jx_j$ and spread describe reconstructed evidence and its uncertainty. Their distribution need not equal the true posterior. For downstream subtraction against a healthy reference $r$ estimating $b$,
\begin{equation*}
\bar x-r=\ell+(\bar x-x)-(r-b).
\label{eq:reconvsref_err}
\end{equation*} 
This separates reconstruction error from reference error. Anchoring addresses the former; the experiments compare reconstructors using a shared reference for each image (\cref{sec:exp-reference}).

\section{Selective anchoring}
\label{sec:method}
\paragraph{Selecting what to preserve.}
A component can be protected directly when its measurement is sufficiently precise. For the compact SVD $A=\sum_i s_i u_i v_i\T$, $s_i>0$, the coefficient $z_i=v_i\T x$ has reading
\begin{equation}
\tilde y_i=u_i\T y/s_i=z_i+e_i,\quad
 e_i\sim\N(0,n_i),\quad n_i=\sigma^2/s_i^2.
\label{eq:reading}
\end{equation}
Small singular values amplify measurement noise, so preserving every measured direction would also retain unreliable components. To select reliable readings, we compare their noise with normal-image variability. Fit a stationary Gaussian surrogate $\N(\mu,C)$ to normal training images and define
\begin{equation}
r_i=\frac{v_i\T Cv_i}{n_i},\qquad \Hs=\{i:r_i>\kappa\},\quad \kappa\geq1.
\label{eq:ratio}
\end{equation}
Large $r_i$ identifies readings precise relative to normal variability. In a scalar Gaussian model, prior shrinkage attenuates departures from the normal mean by $r_i/(1+r_i)$. Using the reading directly avoids this attenuation at a small relative variance cost when $r_i$ is large (\cref{supp:selection}).

Let $V_\Hs$ contain the selected right singular vectors, $P_\Hs=V_\Hs V_\Hs\T$, and $\Pi=I-P_\Hs$. The lift assembles the held coefficients into an image component:
\begin{equation}
\lift(y)=\sum_{i\in\Hs}\tilde y_i v_i.
\label{eq:lift}
\end{equation}
The model generates the free component in the range of $\Pi$, which includes weakly measured directions and the null space. An anomaly can occupy both subspaces, and direct preservation concerns its held component.

\usetikzlibrary{arrows.meta}
\definecolor{heldc}{HTML}{35588F}
\definecolor{freec}{HTML}{C9782B}
\DeclareRobustCommand{\legendline}[1]{%
  \tikz[baseline=-0.6ex]{\draw[line width=0.8pt, -{Latex[length=4.6pt, width=3.8pt]}, #1] (0, 0) -- (17pt, 0);}}
\DeclareRobustCommand{\legendanchor}{\legendline{draw=heldc}}
\DeclareRobustCommand{\legendconditioning}{\legendline{draw=black, densely dashed}}
\DeclareRobustCommand{\legendguidance}{\legendline{draw=freec, dash pattern=on 4pt off 1.2pt on 0.8pt off 1.2pt}}
\begin{figure*}[t]
  \centering
  \input{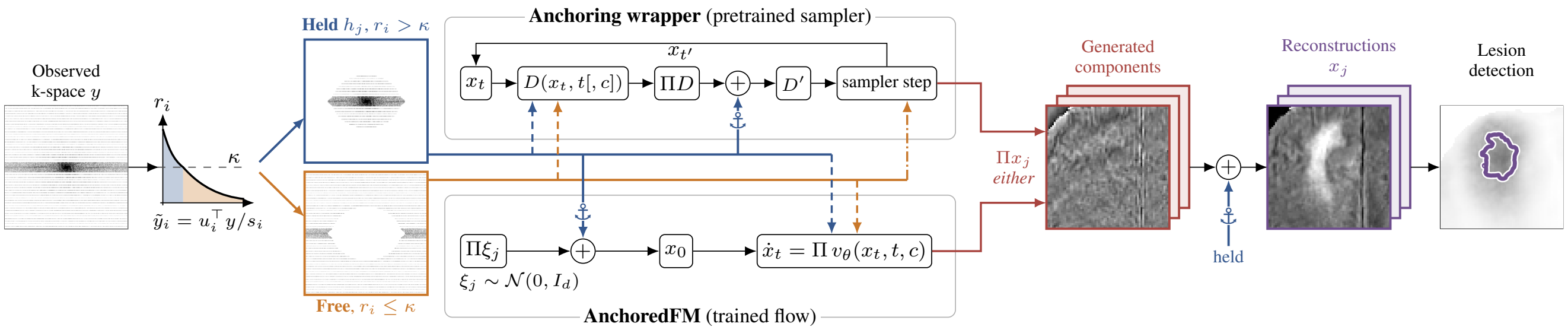}
\caption{
Selective anchoring on a 64-line MRI example ($\kappa=1$). Readings of the observed k-space $y$ (\cref{eq:reading}) are split and ordered by their information ratio $r_i$ (\cref{eq:ratio}; schematic curve) into held and free sets using cutoff $\kappa$. \legendanchor~The anchor $h_j$, deterministic or stochastic (\cref{eq:AnchoredFM-sampling}), enters the wrapper's clean prediction (\cref{eq:wrap}), AnchoredFM's initial state, and the output sum. \legendconditioning~Conditioning on $c=\xmap(y)$, computed from all readings. \legendguidance~Optional guidance uses free readings (\cref{supp:method-details}). Either implementation returns $K=16$ samples $x_j=h_j+\Pi x_j$; AnchoredFM samples under deterministic anchoring are shown. Right: subtraction score $|t|$ (\cref{eq:t}) over its healthy-validation threshold, with the detected lesion outlined.
}
\label{fig:method}
\end{figure*}

\paragraph{Representing measurement uncertainty.}
The anchoring constraint is shared by two uncertainty treatments. \emph{Deterministic anchoring} sets $h_j=\lift(y)$ for every sample and adds the known held-direction variance to downstream tests. \emph{Stochastic anchoring} draws $\eta_j\sim\N(0,I_m)$ independently across samples and sets $h_j=\lift(y+\sigma\eta_j)$, so the ensemble carries acquisition uncertainty. In either case, sample $j$ keeps the same target $h_j$ throughout generation.
We enforce $P_\Hs x_j=h_j$, equivalently $x_j=h_j+\Pi x_j$, so the held component equals its sampled target exactly. 
Under stochastic anchoring, held coefficient $i$ conditional on $y$ follows $\N(\tilde y_i,n_i)$; deterministic anchoring uses this same known measurement law in downstream uncertainty calculations.

\begin{proposition}[Held directions]
\label{prop:held}
Under \cref{eq:model}, fix the held subspace and an image $x=b+\ell$. 
Averaging over acquisition noise for deterministic anchoring, and additionally over the independent target draws for stochastic anchoring, gives
\begin{equation*}
\E[P_\Hs(\bar x-b)]=P_\Hs\ell.
\label{eq:expected_val_lesion}
\end{equation*}
For every held $i$ and chosen nominal coverage level $\gamma\in(0,1)$, the central $\gamma$-interval of $\N(\tilde y_i,n_i)$ covers the held coefficient $z_i$ with probability $\gamma$ over repeated noisy acquisitions of the same fixed image. Both guarantees hold regardless of the model used for the free part.
\end{proposition}

Pixel values combine held and generated components, so their uncertainty also depends on the generator. We therefore evaluate full-image coverage empirically; the proposition concerns analytic held-coordinate intervals under \cref{eq:model}.

We enforce the same target through two implementations: a wrapper for pretrained samplers and AnchoredFM's projected dynamics (\cref{fig:method}).

\paragraph{Wrapper for pretrained samplers.}
The wrapper acts on a sampler's clean-image prediction $D(x_t,t)$. For a straight-path flow, $D=x_t+(1-t)v_\theta(x_t,t,c)$, with conditioning $c$. Replace this prediction by
\begin{equation}
D'(x_t,t)=h_j+\Pi D(x_t,t).
\label{eq:wrap}
\end{equation}
This retains the predicted free component and replaces its held component with the target. The sampler uses $D'$ in its update and projects the returned image to the same anchor. The network retains its conditioning, and additional data updates use the remaining measurements (\cref{supp:method-details,supp:readings}).

\paragraph{AnchoredFM.}
AnchoredFM learns to evolve only the free component. For a normal training image $x_1$, construct $x_0=P_\Hs x_1+\Pi\xi$ and $x_t=(1-t)x_0+tx_1$, with $\xi\sim\N(0,I_d)$. The endpoints share their held component, so the target velocity is $x_1-x_0=\Pi(x_1-\xi)$. With uniform $t\in[0,1]$, flow matching~\cite{lipman2023flow,liu2023rectified} fits this velocity, conditioned on the surrogate MAP image $\xmap(y)$ of noisy measurements:
\begin{equation}
\mathcal L_{\mathrm{AF}}=\E\|v_\theta(x_t,t,\xmap(y))-(x_1-x_0)\|_2^2.
\label{eq:AnchoredFM-loss}
\end{equation}
Training uses clean held coefficients. At inference, initialize with the measurement-based target and project every velocity onto the free subspace:
\begin{equation}
\begin{aligned}
h_j&=\begin{cases}\lift(y),&\text{deterministic},\\ \lift(y+\sigma\eta_j),&\text{stochastic},\end{cases}\\
x_0^{(j)}&=h_j+\Pi\xi_j,\quad \xi_j\sim\N(0,I_d),\\
\dot x_t^{(j)}&=\Pi v_\theta(x_t^{(j)},t,\xmap(y)).
\end{aligned}
\label{eq:AnchoredFM-sampling}
\end{equation}
Because $P_\Hs\Pi=0$, the evolving state retains $P_\Hs x_t^{(j)}=h_j$, independently of network accuracy. We integrate with Euler steps, which require no products with $A$ or $A\T$ after computing the lift and MAP image (\cref{supp:AnchoredFM-training}).

We assess downstream detection conditional on a shared, fixed healthy-reference estimate $r$. We normalize subtraction contrast using reconstruction uncertainty, isolating the effect of anchoring for the same reference. 
For a candidate template $w_p$, define its sample amplitude and the treatment-specific normalized scores by
\begin{equation}
\label{eq:t}
\begin{aligned}
a_j(p)&=\frac{\langle w_p,x_j-r\rangle}{\langle w_p,w_p\rangle},\\
t_{\rm stoch}(p)&=\frac{\bar a(p)}{s_a(p)\sqrt{1+1/K}},\\
t_{\rm det}(p)&=\frac{\bar a(p)}{\sqrt{\tau^2(p)+(1+1/K)s_a^2(p)}}.
\end{aligned}
\end{equation}
Here $\bar a$ and $s_a$ are the sample mean and standard deviation, with variance denominator $K-1$.
For mixed templates, CT uses Satterthwaite degrees of freedom for $t_\mathrm{det}$. MRI uses $K-1$ degrees of freedom for its core normalization and validation-frozen thresholds for practical references (\cref{supp:proof-test,supp:fixed-reference}).

\begin{proposition}[Held-template subtraction]
\label{prop:test}
Under \cref{eq:model}, fix a nonzero held-only template $w$ with $\Pi w=0$, assume the exact reference $r=b$ and $K\geq2$, and let
$\tau^2=\sum_{i\in\Hs}n_i\langle v_i,w\rangle^2/\langle w,w\rangle^2>0$.
Under $\ell=0$, marginally over acquisition noise and, in part (a), the independent anchoring perturbations:
\begin{enumerate}[(a)]
\item For stochastic anchoring, the statistic in \cref{eq:t} follows $t_{K-1}$.
\item For deterministic anchoring, the known-variance statistic $z_{\rm det}=\langle w,\lift(y)-r\rangle/(\langle w,w\rangle\tau)$ follows $\N(0,1)$.
\end{enumerate}
Both null laws are independent of the model used for the free component.
\end{proposition}

Before evaluating the test images, we set $\alpha=5\%$ as the target upper bound false alarm rate, and enforce this image-level bound under the null laws above using a Bonferroni threshold over all position--lesion size pairs. The exact rule is given in \cref{supp:detector-analysis}.

For tests involving free template components or an estimated healthy reference, the false-alarm rate must be checked empirically. The thresholding rule, noncentral extension, mixed-template diagnostics, and proofs are given in \cref{sec:anchored-detection,supp:detector-analysis,supp:proof-test}.

\section{Experiments}
\label{sec:experiments}

\subsection{Setup}
\label{sec:exp-setup}

\paragraph{Tests.}

For real and synthetic pathology cases, all acquisitions are simulated, and the models train on normal reconstructions.
\emph{CT}: LoDoPaB-CT chest slices~\cite{leuschner2021lodopab} on a 256 grid, with 64 parallel-beam views, Gaussian noise at 1\% of the median central reading, and twelve inserted lesion types. Measurements use a finer-grid projector than reconstruction to avoid artificially favorable results from identical discretizations (\cref{supp:problem}).
\emph{Brain MRI}: BraTS FLAIR scans from the Medical Segmentation Decathlon~\cite{antonelli2022msd} on a 256 grid, with 32 or 64 of 256 k-space lines and real tumors with expert masks. Training uses tumor-free slices from training patients.
\emph{Photoacoustic tomography}: VICTRE breast phantoms~\cite{park2026pat,badano2018victre} on a 256 grid, with 64 ring detectors and lesion-bearing/healthy twins.

\paragraph{Samplers.}

We compare conditional flow matching (cFM)~\cite{lipman2023flow} and EDM~\cite{karras2022edm}, both conditioned on $\xmap(y)$, with six zero-shot methods sharing one unconditional prior: DPS~\cite{chung2023dps}, $\Pi$GDM~\cite{song2023pigdm}, DDRM~\cite{kawar2022ddrm}, DDNM~\cite{wang2023ddnm}, DiffPIR~\cite{zhu2023diffpir}, and Posterior-DDIM~\cite{jiao2026posteriorddim}.
Each returns $K=16$ samples, with and without anchoring. 
MRI uses all eight. The 256-grid CT test omits DPS and $\Pi$GDM for cost (both run on
the 128 grid; \cref{supp:tab:ct128}). Photoacoustics uses cFM, DPS and DDNM.
Anchored CT data steps use per-sample readings at $\sqrt2\sigma$. MRI and photoacoustics use shared readings (\cref{supp:readings}).

\paragraph{Protocol.}
We split by patient, select settings on part 0, and evaluate parts 1--3 using paired patient-bootstrap 95\% intervals.
Base settings optimize preset-threshold F1 in CT and lesion error in MRI/photoacoustics. Anchored pairs retain these settings.
CT retains its healthy-validation choice of $\kappa$ when changing the uncertainty treatment, transferring it from the 128 to the 256 grid.
MRI and photoacoustics select $\kappa$ using labeled lesion error.
In the primary MRI tables, \emph{+ anchoring} denotes deterministic anchoring. Subtraction scores include analytic protected acquisition variance. P-DDIM uses $\eta_0=0$, $\eta_1=16$ in the primary MRI comparisons. Table significance uses bootstrap $p<0.05$, Holm-adjusted across sampler pairs. Intervals quoted in the text are unadjusted.

\paragraph{Controls.}

Controls include a measurement-domain matched filter (the direct measurement control, labeled ``ideal observer'' in the tables), and the exact Gaussian surrogate posterior with sampled or analytic covariance.
Using the same lesion templates and Bonferroni tests, the matched filter provides a detection upper bound for a known healthy reference, template, and position under the stated Gaussian model.
Post-hoc rescaling fits one spread factor per base CT/MRI sampler on healthy part 0 and fixes it for testing.
We also adapt KLIP~\cite{kheirandish2026klip} using the shared DPS prior, with guidance and readout selected on validation patients (\cref{supp:klip}).
Acquisition, sampler, selection and control details are in
\cref{supp:baselines,supp:setup-details}.

\paragraph{Detection and reference images.}
All subtraction-based methods use the same healthy-reference estimate for a test image. KLIP is reference-free. Primary MRI detection uses mask-free checkerboard inpainting: a total-variation reconstruction is regenerated in alternating 16\,mm squares, and the mean of 16 healthy inpaintings is treated as fixed. We average ten reference seeds. Thresholds are fitted only on healthy part 0 and frozen for parts 1--3. The expert-mask oracle is a supplementary control (\cref{supp:fixed-reference}). CT uses the lesion-free image, and photoacoustics uses the healthy twin.

\paragraph{Metrics.}
We evaluate anomaly preservation through recovered lesion amplitude in CT and 
lesion-to-background contrast recovery in MRI. Both are expressed relative to 
the ground truth: 100\% indicates full recovery, while values near zero indicate 
suppression. Contrast error is the mean absolute departure from full recovery. Reconstruction fidelity is measured by the root mean squared error 
(RMSE) of the ensemble mean, reported separately within lesions and over the body 
or normal anatomy to distinguish preservation from overall image quality. RMSE is 
expressed in HU for CT and as a percentage of median intensity for MRI and 
photoacoustics. For CT and MRI, whole-image SSIM of the ensemble mean measures structural fidelity (dimensionless; \cref{supp:ssim}). 
For downstream detection, area under the receiver operating 
characteristic curve (AUROC) ranks reconstructions by their largest detection score.
Subtraction uses
threshold-normalized $|t|$ scores and KLIP uses likelihood-update scores. 
Det$_5$ measures sensitivity at a separately
swept threshold giving each method a matched 5\% false-alarm rate on healthy test slices. An anomaly 
counts as detected only when a flag overlaps it. In MRI, Sens.$_{\rm val}$ and FA$_{\rm val}$ report test sensitivity and false alarms at thresholds fitted to a 5\% healthy-validation false-alarm target and frozen before testing. MRI AUROC and slice sensitivity average four predefined tumor-size classes. False alarms (FA) at the
preset Bonferroni threshold ($\alpha=5\%$) assess performance without threshold adjustment.
Free-response receiver operating characteristic (FROC) analysis instead counts localized lesion detections and false marks on all evaluated reconstructions;
Sens.@0.25 is lesion sensitivity at 0.25 false-positive marks per evaluated image, or per evaluated slice for MRI. This operating point is selected on the test FROC curve.

\subsection{Anomaly preservation and reconstruction fidelity}
\label{sec:exp-preservation}

The size of the effect depends on the sampler and the measurement. 
For inserted CT lesions, with known amplitude and acquisition uncertainty, cFM and EDM recover 61--62\% of lesion contrast without anchoring and 78--79\% with deterministic
anchoring. The four data-step samplers already recover 72--85\% and gain 0.4--2.5
points. Lesion RMSE falls for five of the six pairs and is unchanged for P-DDIM
(\cref{supp:tab:ct}). In photoacoustics, stochastic anchoring halves lesion error for
cFM and DPS and leaves DDNM, whose error was already low, unchanged
(\cref{supp:tab:pat}).

In MRI at nominal noise the base samplers already retain 89--93\% of tumor 
contrast and anchoring adds about one point on average (\cref{supp:tab:brain-det-reconstruction}). Even so, mean tumor RMSE across the eight sampler pairs falls from 11.5\% to 11.0\% of median brain intensity and mean absolute contrast error from 8.4\% to 7.3\% although DPS contrast error rises from 7.2\% to 7.7\% (\cref{tab:brain}).
The benefit grows with acquisition noise: for the four zero-shot
samplers tested under stochastic anchoring, contrast recovery improves by 
0.0--1.4 points at the nominal level and by 1.8--10.0 points at eight 
times that level, where the base samplers retain only 72--84\% 
(\cref{supp:tab:brain-noise}). 
Anchoring also lowers mean whole-image SSIM from
0.950 to 0.804 (\cref{supp:tab:brain-det-reconstruction}). The drop is the 
measurement noise kept in the held coefficients. Putting the noisy held readings 
into the unanchored reconstructions gives 0.797 and into the true images 0.843, 
with SSIM of 0.990 inside the brain and 0.810 in the empty background 
(\cref{supp:tab:ssim-noise}).

\begin{table}[t]
  \caption{Brain MRI, 256 grid and 64 lines: 661 tumor and 701 healthy slices from 60 patients. Contrast error is absolute departure from full recovery. Reconstruction and detection metrics are in \%, with RMSE relative to median intensity. Subtraction detection uses shared checkerboard references over ten seeds. The reference-free KLIP panel reports AUROC, retrospective Det$_5$ and validation-frozen FA (\cref{supp:klip}). Bold/italic mark better/worse reconstruction (Holm-adjusted); detection columns are unmarked. AnchoredFM is compared with cFM.}
  \label{tab:brain}
  \centering
  \footnotesize
  \setlength{\tabcolsep}{2pt}
  \begin{tabular}{lrrrrr}
\toprule
Sampler & RMSE & Contrast error & Sens.$_{\rm val}$ & Sens.@0.25 & FA$_{\rm val}$ \\
\midrule
cFM & 13.9 & 11.1 & 23.5 & 40.8 & 4.9 \\
\quad + anchoring & \textbf{12.2} & \textbf{8.4} & 46.8 & 49.0 & 3.4 \\
EDM & 11.7 & 9.4 & 41.3 & 45.6 & 3.5 \\
\quad + anchoring & \textbf{10.8} & \textbf{7.1} & 45.1 & 47.7 & 3.4 \\
\midrule
DPS & 11.0 & 7.2 & 15.8 & 34.6 & 2.6 \\
\quad + anchoring & 11.0 & \textit{7.7} & 40.0 & 45.1 & 3.1 \\
$\Pi$GDM & 10.9 & 7.6 & 22.6 & 41.7 & 3.5 \\
\quad + anchoring & \textbf{10.8} & 7.6 & 44.0 & 46.8 & 3.4 \\
DDRM & 11.6 & 9.3 & 15.7 & 38.1 & 2.4 \\
\quad + anchoring & \textbf{11.2} & \textbf{7.8} & 43.0 & 47.3 & 3.2 \\
DDNM & 11.1 & 7.0 & 2.3 & 29.0 & 0.6 \\
\quad + anchoring & 11.0 & \textbf{6.6} & 43.5 & 47.9 & 3.4 \\
DiffPIR & 11.0 & 7.7 & 28.9 & 41.7 & 5.0 \\
\quad + anchoring & \textbf{10.6} & \textbf{6.3} & 45.0 & 48.6 & 3.4 \\
P-DDIM & 11.0 & 8.3 & 11.6 & 38.7 & 3.4 \\
\quad + anchoring & \textbf{10.8} & \textbf{7.3} & 45.0 & 48.0 & 3.3 \\
\midrule
AnchoredFM & \textbf{12.3} & \textbf{9.3} & 46.4 & 49.6 & 3.4 \\
\midrule
\multicolumn{3}{l}{Reference-free KLIP control} & AUROC & Det$_5$ & FA$_{\rm val}$ \\
\multicolumn{3}{l}{KLIP} & 83.7 & 55.1 & 6.7 \\
\multicolumn{3}{l}{\quad + deterministic anchoring} & 93.7 & 71.2 & 4.9 \\
\bottomrule
\end{tabular}

\end{table}

\subsection{Measurement-based uncertainty calibration}
\label{sec:exp-calibration}

Across the eight matched MRI sampler pairs at 64 lines, nominal 90\% ensemble
intervals cover 68.3\% of the held coefficients on average without anchoring and
90.0\% with it, and every pair moves closer to nominal
(\cref{supp:tab:held-cover}). The largest corrections occur for samplers with severe
undercoverage (DDNM 32.0\%, P-DDIM 57.2\%, DDRM 64.3\%). cFM starts at 84.2\%, and EDM
is already near nominal at 90.6\%. Anchoring therefore brings different uncertainty
behaviors to one measurement-defined level. These directions span healthy and
anomalous anatomy, so the check extends beyond lesions and tests the prediction of
\cref{prop:held}. The analytic intervals of deterministic anchoring cover
90.0--90.1\% at every $\kappa$ (\cref{supp:held-coverage}).

Inside lesions, mean coverage under stochastic anchoring rises from 72.9\% to 88.2\%
across the six CT pairs and from 28.1\% to 67.9\% across the three photoacoustic pairs,
with every pair improving (\cref{supp:tab:ct,supp:tab:pat}). In MRI it rises for seven of the
eight samplers, falling only for DPS. It also stays below nominal at 59.2--84.2\%
(\cref{supp:tab:brain-full}). Pixels combine held and generated components, so
nominal held-coordinate coverage does not imply nominal coverage throughout a lesion.
For the same reason, the unadjusted Bonferroni threshold flags every healthy MRI slice
under the checkerboard reference, requiring threshold fitting on healthy validation slices
(\cref{supp:fixed-reference}).

\subsection{Downstream detection}
\label{sec:exp-reference}

Better lesion preservation retains the signal needed for detection, while more reliable uncertainty helps distinguish that signal from reconstruction error. Across all eight matched MRI sampler pairs with the checkerboard healthy reference at 64 lines, deterministic anchoring raises mean sensitivity at validation-frozen thresholds from 20.2\% to 44.1\%, while realized false alarms remain similar (3.2\% and 3.3\%). AUROC rises from 78.4\% to 84.6\% (\cref{supp:tab:practical-fixed64}). The largest gains occur for samplers that fail without anchoring (DDNM 2.3\%, P-DDIM 11.6\%). The gap between the best and worst of the eight samplers shrinks from
39.0 to 6.8 points. EDM, the strongest base sampler, still gains 3.9 points (\cref{supp:tab:practical-fixed-contrasts64}).

Anchoring also helps a detector from prior work. With the reference-free
KLIP readout~\cite{kheirandish2026klip} on the shared DPS prior, deterministic
anchoring raises AUROC from 83.7\% to 93.7\% and Det$_5$ from 55.1\% to 71.2\%
(\cref{tab:brain,supp:klip}), the highest of any pipeline in the MRI test (anchored
subtraction averages 84.6\% and 49.0\%; \cref{supp:tab:practical-fixed64}).

Detection performance varies with the quality of the healthy reference. In the oracle comparison, access to the true image and lesion masks yields near-perfect AUROC and Det$_5$ even without anchoring. Anchoring still improves lesion localization: at 64 MRI lines, mean sensitivity across eight sampler pairs rises from 67.4\% to 82.2\% at 0.25 false-positive marks per slice. Preset false alarms fall from 67.9\% to 9.0\%, although they remain above the nominal 5\% target (\cref{supp:tab:oracle64}). Checkerboard inpainting builds the reference without lesion masks and provides a more practical test. It can still alter healthy anatomy or retain lesion signal, leading to false alarms or missed lesions.

On CT, deterministic anchoring raises Det$_5$ from 33.1--59.6\% to 
65.5--66.9\% across the six samplers and lowers preset false alarms from 
30.7--100\% to 1.9--7.1\% (\cref{supp:tab:ct,supp:tab:draw}). Stochastic anchoring reaches 55.5--59.2\%, within $-3.6$ to $+0.6$ points of base for the four data-step samplers, so its gain there is calibration. 
Two controls separate these effects from simpler explanations
(\cref{supp:tab:summary}). Post-hoc rescaling multiplies every score by one factor fitted on healthy validation images, so it corrects false alarms but does not change AUROC or Det$_5$. For CT samplers without a measurement update it lowers median preset false alarms from 40\% to 6\% and leaves Det$_5$ at 34\%. 
The Gaussian surrogate posterior uses no learned model yet matches stochastic anchoring on detection (Det$_5$ 55.2\%, or 60.7\% with its exact test) and falls below deterministic anchoring, while its reconstructions are worse with SSIM 0.724 vs. 0.740--0.850 (\cref{supp:tab:ct}). Under stochastic anchoring the measurements supply the detection evidence and the learned prior supplies the image.

\begin{table*}[t]
  \caption{Lesion sensitivity (\%) by tumor size in brain MRI at 64 of 256 lines, using the fixed
  checkerboard-inpainting reference averaged over ten seeds on parts 1--3 (791 lesions in 661 tumor slices and 701 tumor-free slices of 60 patients).
  Values average the eight matched sampler pairs in \cref{tab:brain}. Size is the predefined equivalent diameter
  of each tumor component in the slice. For each sampler and operating point, one score threshold is set over all
  1362 evaluated slices and held fixed across size groups. Differences are anchored minus base in percentage
  points, with 95\% intervals from paired patient-and-reference-seed bootstrap draws.}
  \label{tab:size}
  \centering
  \small
  \setlength{\tabcolsep}{4pt}
  \resizebox{\textwidth}{!}{\begin{tabular}{lrrrrrrrrr}
\toprule
 & & & & \multicolumn{3}{c}{0.25 false positives per slice} & \multicolumn{3}{c}{1 false positive per slice} \\
\cmidrule(lr){5-7}\cmidrule(lr){8-10}
Tumor size & Lesions & Slices & Patients & base & anchored & difference & base & anchored & difference \\
\midrule
11--23\,mm & 217 & 173 & 60 & 6.9 & 10.1 & +3.2 [+1.0, +5.5] & 20.5 & 29.2 & +8.7 [+5.6, +11.4] \\
23--34\,mm & 214 & 173 & 59 & 30.1 & 43.0 & +13.0 [+9.5, +16.5] & 55.3 & 63.8 & +8.4 [+5.8, +11.4] \\
34--45\,mm & 195 & 167 & 56 & 55.7 & 67.3 & +11.6 [+8.3, +15.2] & 74.8 & 80.2 & +5.5 [+3.0, +8.4] \\
$\geq$45\,mm & 165 & 148 & 50 & 71.9 & 79.3 & +7.3 [+4.3, +10.4] & 86.1 & 87.4 & +1.3 [-0.1, +2.9] \\
All & 791 & 661 & 60 & 38.8 & 47.5 & +8.8 [+6.9, +10.6] & 57.0 & 63.3 & +6.3 [+4.9, +7.7] \\
\bottomrule
\end{tabular}
}
\end{table*}

\subsection{Detection across tumor sizes}
\label{sec:exp-size}
The detection gain holds across tumor sizes. We divide the MRI lesions into four
predefined equivalent-diameter groups and, for each sampler, apply one globally
selected FROC threshold to every group (\cref{tab:size}). At 0.25 false-positive
marks per slice, deterministic anchoring raises mean sensitivity in all four groups.
For the smallest 11--23\,mm tumors it rises from 6.9\% to 10.1\%, although sensitivity for these tumors remains low, and over
all lesions the gain is $+8.8$ points. Full per-method results and the
oracle control are in \cref{supp:tab:checkerboard-size64,supp:tab:oracle64}.

\subsection{AnchoredFM sampling efficiency}

\begin{figure}[t]
  \centering
  \includegraphics[width=\linewidth]{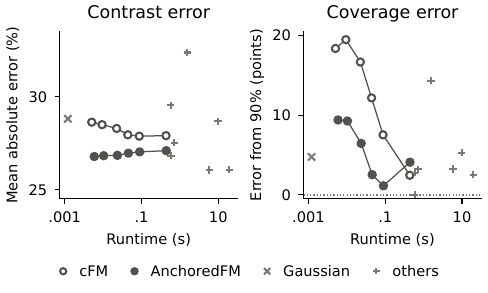}
  \caption{Stochastic-anchoring CT runtime study (128 grid, 64 views): contrast error (left) and coverage error (right) versus runtime for 16 samples, evaluated on \res{ct.runtime.images} lesion reconstructions from \res{ct.runtime.patients} test patients (parts 1--3). AnchoredFM and anchored cFM vary their step budgets. Other samplers use full budgets. Contrast error is the mean absolute deviation from full recovery; coverage error is the absolute deviation from 90\% coverage.}
  \label{fig:efficiency}
\end{figure}

In a CT runtime study on one H100 (128 grid, 64 views), AnchoredFM generates a 16-sample ensemble in \res{ct.runtime.anchorflow.seconds.ultra} with one network evaluation per sample. 
Among learned samplers completing within 12\,ms, it achieves the lowest contrast and coverage errors, while anchored cFM achieves lower lesion RMSE. Compared with the best full-budget learned results, lesion RMSE is \res{ct.runtime.anchorflow.vs-best.rmse-relative} higher and contrast error is \res{ct.runtime.anchorflow.vs-best.contrast-gap} higher, while sampling is \res{ct.runtime.anchorflow.vs-cfm-full.speedup} 
faster than full-budget anchored cFM. A budget of this size matters when reconstruction must keep pace with acquisition, as in real-time or interactive MRI, where a multi-sample ensemble is otherwise too slow to produce for every frame. \Cref{fig:efficiency} shows the runtime--quality trade-off; full numerical comparisons are in \cref{supp:efficiency-details}.

\section{Discussion and limitations}
\label{sec:limitations}

Our results distinguish anatomical fidelity from the preservation of abnormality. Global image similarity and lesion-specific metrics do not always improve together, as clinical reader and detection studies 
have also found~\cite{vanLohuizen2024Assessing,Jensen2022}, motivating separate assessment of anatomical reconstruction and retained lesion evidence. Selective anchoring handles this trade-off by protecting reliable measurement-derived components while using the learned prior to reconstruct unresolved structure.

Lesion contrast and uncertainty play complementary roles in detection. Preserving contrast retains the signal of an abnormality, while uncertainty helps assess whether a residual could be explained by reconstruction error. Underestimated uncertainty in healthy tissue can therefore create false alarms despite visually plausible images. This motivates assessing retained lesion evidence and uncertainty together, with measurement-based guarantees providing a reference for the protected components.
Anchoring's largest detection gains occur for samplers whose data steps reuse the same readings in every sample, although no single mechanism explains all samplers; after anchoring, detection depends far less on which sampler is used.

\paragraph{Choosing $\kappa$.}

Smaller $\kappa$ protects more measured directions but retains more noise; larger $\kappa$ leaves more to the prior.
For independent Gaussian singular coordinates, anchored variance is at most $1+1/\kappa$ times posterior variance (\cref{supp:gaussian}).
This criterion measures relative measurement reliability rather than lesion relevance, and AnchoredFM requires separate training for each $\kappa$.

\paragraph{What anchoring cannot do.}

Anchoring directly protects only well-measured anomaly components, and its uncertainty guarantees apply only to held coordinates.
Overconfidence in free components can still cause false alarms, particularly in MRI, where most directions are free.
Subtraction also depends on a healthy reference that may retain lesions or alter normal anatomy.
Estimating uncertainty from 16 samples further limits detection power.

\paragraph{Assumptions.}

We assume a known linear operator and noise level.
The exact Student $t$ law requires a held-only template, an exact healthy reference, Gaussian acquisition noise, and independent Gaussian anchoring draws (\cref{prop:test}).
Operator error about ten times the noise worsens lesion reconstruction for every anchored photoacoustic sampler (\cref{supp:pat-error}).
Underestimating CT noise by 20\% raises false alarms to \res{ct128.sigma.fa.low} in the 128-grid test. On the 256 grid at 64 views, the residual outside the range of $A$ estimates noise at \res{ct.sigma.est.64} times its true level (\cref{supp:sigma}).
The stationary Gaussian surrogate used for selection ignores correlations between singular directions.
CT data steps use per-sample readings with effective noise $\sqrt2\sigma$; assuming $\sigma$ substantially worsens DDRM and DDNM reconstruction.
MRI and photoacoustics use shared readings, which left these samplers overconfident in CT (\cref{supp:readings}).

\paragraph{Scope of the evaluation.}

All acquisitions are simulated.
MRI readings are generated from magnitude images using the reconstruction operator, an implicit data crime~\cite{shimron2022crimes}.
Performance on raw multi-coil MRI remains untested.

\paragraph{Conclusion.}
\label{sec:conclusion}

Selective anchoring preserves well-measured lesion contrast in expectation while a learned prior reconstructs unresolved anatomy.
Across the tested modalities, anchoring improves anomaly preservation and uncertainty reliability.
With checkerboard-inpainting references in MRI, deterministic anchoring raises mean localized lesion sensitivity from 38.8\% to 47.5\% at 0.25 false-positive marks per slice, with improvements across all four predefined tumor-size groups.
Selective anchoring lets the prior complete what the measurements leave unresolved, without healing away abnormalities they support.

\section*{LLM Usage Statement}
Claude Code (Claude Fable 5-family models) was used to assist with writing and running
the experiment and analysis code, under human direction, and with drafting and
reviewing the manuscript. ChatGPT and OpenAI Codex (Sol-family models) were also used
for writing assistance (comparing drafts, revising text, searching related work) and to
draw the method figure. All research ideation, experimental design, and scientific
conclusions are the authors' own. The authors take full responsibility for all content
in this paper.

\section*{Acknowledgements}
This material is based upon work supported by the U.S. National Science Foundation under Award No. CBET-2143625. J.C. is partially supported by the Jack Kent Cooke Foundation. This research was performed using computational resources sponsored by the U.S. Department of Energy's Office of Critical Minerals and Energy Innovation and located at the National Laboratory of the Rockies. Any opinions, findings, and conclusions or recommendations expressed in this material are those of the authors and do not necessarily reflect the views of the National Science Foundation.

{
    \small
    \bibliographystyle{unsrtnat}
    \bibliography{refs}
}

\clearpage

\setcounter{section}{0}
\setcounter{subsection}{0}
\setcounter{table}{0}
\setcounter{figure}{0}
\setcounter{equation}{0}
\setcounter{proposition}{0}
\setcounter{corollary}{0}
\setcounter{remark}{0}
\renewcommand{\thesection}{S\arabic{section}}
\renewcommand{\thesubsection}{\thesection.\arabic{subsection}}
\renewcommand{\thesubsubsection}{\thesubsection.\arabic{subsubsection}}
\renewcommand{\thetable}{S\arabic{table}}
\renewcommand{\thefigure}{S\arabic{figure}}
\renewcommand{\theequation}{S\arabic{equation}}
\renewcommand{\theproposition}{S\arabic{proposition}}
\renewcommand{\thecorollary}{S\arabic{corollary}}
\renewcommand{\theremark}{S\arabic{remark}}
\renewcommand{\theHsection}{supp.\arabic{section}}
\renewcommand{\theHsubsection}{\theHsection.\arabic{subsection}}
\renewcommand{\theHsubsubsection}{\theHsubsection.\arabic{subsubsection}}
\renewcommand{\theHtable}{supp.\arabic{table}}
\renewcommand{\theHfigure}{supp.\arabic{figure}}
\renewcommand{\theHequation}{supp.\arabic{equation}}
\providecommand{\theHproposition}{}
\providecommand{\theHcorollary}{}
\providecommand{\theHremark}{}
\renewcommand{\theHproposition}{supp.\arabic{proposition}}
\renewcommand{\theHcorollary}{supp.\arabic{corollary}}
\renewcommand{\theHremark}{supp.\arabic{remark}}
\providecommand{\SIfreezecounter}[2]{\setcounter{#1}{\numexpr#2-1\relax}}
\newsavebox{\SIfitbox}
\providecommand{\SIfitwidth}[2]{%
  \sbox{\SIfitbox}{#2}%
  \ifdim\wd\SIfitbox>#1\resizebox{#1}{!}{\usebox{\SIfitbox}}%
  \else\usebox{\SIfitbox}\fi}

\providecommand{\SIcite}[1]{\hyperref[supp:ref:#1]{[\ref*{supp:ref:#1}]}}

\setlength{\textfloatsep}{6pt plus 1pt minus 1pt}
\setlength{\dbltextfloatsep}{6pt plus 1pt minus 1pt}
\setlength{\floatsep}{6pt plus 1pt minus 1pt}
\setlength{\dblfloatsep}{6pt plus 1pt minus 1pt}
\setlength{\intextsep}{6pt plus 1pt minus 1pt}
\setlength{\abovecaptionskip}{3pt}
\setlength{\belowcaptionskip}{0pt}
\setcounter{topnumber}{4}
\setcounter{bottomnumber}{3}
\setcounter{dbltopnumber}{5}
\setcounter{totalnumber}{6}
\renewcommand{\topfraction}{0.95}
\renewcommand{\bottomfraction}{0.95}
\renewcommand{\dbltopfraction}{0.95}
\renewcommand{\textfraction}{0.05}
\renewcommand{\floatpagefraction}{0.75}
\renewcommand{\dblfloatpagefraction}{0.75}
\makeatletter
\setlength{\@fptop}{0pt}
\setlength{\@fpsep}{6pt}
\setlength{\@fpbot}{0pt plus 1fil}
\setlength{\@dblfptop}{0pt}
\setlength{\@dblfpsep}{6pt}
\setlength{\@dblfpbot}{0pt plus 1fil}
\makeatother

\setcounter{page}{1}
\renewcommand{\thepage}{S\arabic{page}}
\maketitlesupplementary

\setlength{\abovedisplayskip}{6pt plus 1pt minus 1pt}
\setlength{\belowdisplayskip}{6pt plus 1pt minus 1pt}
\setlength{\abovedisplayshortskip}{3pt plus 1pt minus 1pt}
\setlength{\belowdisplayshortskip}{4pt plus 1pt minus 1pt}

Deterministic anchoring with analytic held acquisition variance is the default for computed tomography (CT) and magnetic resonance imaging (MRI); labeled stochastic studies are secondary. Numbered citations without an S prefix refer to the main-paper bibliography.

\defres{tests.ct.images}{1800}
\defres{tests.ct.anomalous}{1080}
\defres{tests.ct.free}{720}
\defres{tests.ct.patients}{45}
\defres{tests.ct.types}{12}
\defres{intro.amplitude.base}{25--41\%}
\defres{intro.amplitude.anchored}{61--83\%}
\defres{intro.fa.base}{63\%}
\defres{ct.recovery.base}{61--85\%}
\defres{ct.recovery.anchored}{78--88\%}
\defres{intro.fa.anchored}{2\%}
\defres{ct.cov0.base}{66--92\%}
\defres{ct.cov.base}{61--85\%}
\defres{ct.cov.anchored}{83--97\%}
\defres{ct.pairs}{6}
\defres{ct.pairs.mismatch}{0}
\defres{ct.auroc.task}{10 to 12}
\defres{ct.auroc.zero}{$-$1 to $+$1}
\defres{ct.det5.task}{24}
\defres{ct.det5.zero}{$-$4 to $+$2}
\defres{ct.scale.zero}{1.4--5.7}
\defres{ct.scale.anchored}{0.88--1.03}
\defres{ct.fa.anchored.but.ddnm}{1--3\%}
\defres{ct.f1.better}{6}
\defres{ct.f1.worse}{none}
\defres{ct.fa.better}{6}
\defres{ct.fa.worse}{none}
\defres{ct.cov.better}{5}
\defres{ct.cov.worse}{1}
\defres{ct.amplitude.better}{6}
\defres{ct.amplitude.worse}{none}
\defres{ct.rmse.better}{2}
\defres{ct.rmse.worse}{4}
\defres{ct.auroc.better}{2}
\defres{ct.auroc.worse}{none}
\defres{ct.det5.better}{2}
\defres{ct.det5.worse}{1}
\defres{ct.teaser.only_anchored_detects}{97}
\defres{ct.teaser.only_base_detects}{35}
\defres{ct.teaser.only_base_flags}{341}
\defres{ct.teaser.only_anchored_flags}{10}
\defres{ct.kappa.64}{10}
\defres{ct.nodraw.fa.64}{99\%}
\defres{ct.ddnm.fa.64}{2\%}
\defres{ct.anchorflow.f1.64}{$-$0.8}
\defres{ct.anchorflow.f1ci.64}{[$-$2.5, $+$0.9]}
\defres{ct.observer.det5.64}{76\%}
\defres{ct.observer.fa.64}{0.8\%}
\defres{ct.gp.det5.64}{55\%}
\defres{ct.gp.fa.64}{2.8\%}
\defres{ct.gp.exact.det5.64}{61\%}
\defres{ct.gp.exact.fa.64}{9.7\%}
\defres{ct.fbp.det5.64}{38\%}
\defres{ct.fbp.fa.64}{90.0\%}
\defres{ct.det5.anchored.64}{56--59\%}
\defres{ct.lock.fa.64}{99--100\%}
\defres{ct.lock.det5.64}{47--55\%}
\defres{ct.lockknown.fa.64}{4--100\%}
\defres{ct.lockknown.det5.64}{47--67\%}
\defres{ct.known.fa.64}{16--51\%}
\defres{ct.known.det5.64}{56--63\%}
\defres{ct.quality.gp.64}{78}
\defres{ct.quality.anchorflow.64}{49}
\defres{ct.quality.task.64}{47--49}
\defres{ct.quality.ddrm.base.64}{56--73}
\defres{ct.quality.ddrm.anchored.64}{110--137}
\defres{ct.sigma.est.64}{1.048}
\defres{ct.sigma.est.range.64}{0.97--1.13}
\defres{ct.problem.rows.64}{20925}
\defres{ct.problem.rank.64}{20527}
\defres{ct.seconds.anchorflow.64}{1.6}
\defres{ct.seconds.trained.64}{1.6--2.3}
\defres{ct.seconds.zero.64}{3.5--36}
\defres{ct.seconds.ratio.64}{1.01--1.05}
\defres{ct.draw.det5}{$-$10 to $+$8}
\defres{ct.draw.det5.pairs}{6}
\defres{ct.draw.det5.up}{1}
\defres{ct.draw.det5.down}{4}
\defres{ct.draw.det5.task}{$-$10 to $-$8}
\defres{ct.draw.det5.task.pairs}{3}
\defres{ct.draw.det5.task.up}{none}
\defres{ct.draw.det5.task.down}{3}
\defres{ct.draw.det5.zero}{$-$7 to $+$8}
\defres{ct.draw.det5.zero.pairs}{3}
\defres{ct.draw.det5.zero.up}{1}
\defres{ct.draw.det5.zero.down}{1}
\defres{ct.draw.fa}{$-$98 to $-$2}
\defres{ct.draw.fa.pairs}{6}
\defres{ct.draw.fa.up}{none}
\defres{ct.draw.fa.down}{6}
\defres{ct.draw.fa.task}{$-$4 to $-$2}
\defres{ct.draw.fa.task.pairs}{3}
\defres{ct.draw.fa.task.up}{none}
\defres{ct.draw.fa.task.down}{3}
\defres{ct.draw.fa.zero}{$-$98 to $-$36}
\defres{ct.draw.fa.zero.pairs}{3}
\defres{ct.draw.fa.zero.up}{none}
\defres{ct.draw.fa.zero.down}{3}
\defres{ct.lockknown.fa.task.64}{4--7\%}
\defres{ct.lockknown.fa.zero.64}{37--100\%}
\defres{ct.anchorflow.vs.gp.64}{$+$2.1}
\defres{ct.anchorflow.vs.gp.ci.64}{[$-$0.1, $+$5.5]}
\defres{ct.prop.samplers}{7}
\defres{ct.prop.quantile}{2.0\%}
\defres{ct.two.fa.task.base}{0\%}
\defres{ct.two.fa.task.anchored}{0--1\%}
\defres{ct.two.det5.task}{$-$0.1 to $+$0.1}
\defres{ct.two.f1.task}{3 to 5}
\defres{ct.two.fa.zero.base}{27--100\%}
\defres{ct.two.fa.zero.anchored}{0\%}
\defres{ct.two.det5.zero}{$-$3.3 to $-$1.6}
\defres{ct.two.f1.zero}{5 to 19}
\defres{ct.two.fa.anchored.0.5mm}{0--1\%}
\defres{ct.two.fa.base.0.5mm}{0--100\%}
\defres{ct.shift.fa.anchored.0.5mm}{2--4\%}
\defres{ct.two.fa.anchored.1mm}{4--7\%}
\defres{ct.two.fa.base.1mm}{6--100\%}
\defres{ct.shift.fa.anchored.1mm}{14--22\%}
\defres{ct.two.fa.anchored.2mm}{40--53\%}
\defres{ct.two.fa.base.2mm}{64--100\%}
\defres{ct.shift.fa.anchored.2mm}{72--84\%}
\defres{ct.two.gp.det5}{52\%}
\defres{ct.two.observer.det5}{69\%}
\defres{ct.two.anchorflow.det5}{55\%}
\defres{ct.two.fbp.det5}{51\%}
\defres{ct.two.anchorflow.vs.gp}{$+$2.6}
\defres{ct.two.anchorflow.vs.gp.ci}{[$-$0.7, $+$5.6]}
\defres{ct.recal.fa.base}{3--7\%}
\defres{ct.recal.scale.base}{1.3--5.4}
\defres{ct.recal.scale.anchored}{0.84--0.93}
\defres{ct.recal.vs.task}{22 to 23}
\defres{ct.recal.vs.task.pairs}{2}
\defres{ct.recal.vs.task.sig}{2}
\defres{ct.recal.vs.task.up}{2}
\defres{ct.recal.vs.task.down}{none}
\defres{ct.recal.vs.zero}{$-$2 to $+$2}
\defres{ct.recal.vs.zero.pairs}{4}
\defres{ct.recal.vs.zero.sig}{1}
\defres{ct.recal.vs.zero.up}{none}
\defres{ct.recal.vs.zero.down}{1}
\defres{ct.recal.vs.all}{$-$2 to $+$23}
\defres{ct.recal.vs.all.pairs}{6}
\defres{ct.recal.vs.all.sig}{3}
\defres{ct.recal.vs.all.up}{2}
\defres{ct.recal.vs.all.down}{1}
\defres{ct.held.base}{20--93\%}
\defres{ct.held.anchored}{90\%}
\defres{ct.fheld.median}{71\%}
\defres{ct.aupro.task}{6 to 7}
\defres{ct.aupro.zero}{0 to $+$1}
\defres{ct.pixel_auroc.task}{2}
\defres{ct.pixel_auroc.zero}{0 to $+$1}
\defres{ct.dice_max.task}{1}
\defres{ct.dice_max.zero}{1}

\defres{tests.ct128.images}{2880}
\defres{tests.ct128.anomalous}{2160}
\defres{tests.ct128.free}{720}
\defres{tests.ct128.patients}{45}
\defres{tests.ct128.types}{12}
\defres{ct128.intro.amplitude.base}{26--50\%}
\defres{ct128.intro.amplitude.anchored}{63--89\%}
\defres{ct128.intro.fa.base}{75\%}
\defres{ct128.recovery.base}{43--87\%}
\defres{ct128.recovery.anchored}{71--94\%}
\defres{ct128.intro.fa.anchored}{2\%}
\defres{ct128.cov0.base}{52--94\%}
\defres{ct128.cov.base}{50--84\%}
\defres{ct128.cov.anchored}{81--96\%}
\defres{ct128.pairs}{15}
\defres{ct128.pairs.mismatch}{0}
\defres{ct128.auroc.task}{5 to 21}
\defres{ct128.auroc.zero}{$-$1 to $+$3}
\defres{ct128.det5.task}{17 to 45}
\defres{ct128.det5.zero}{$-$2 to $+$8}
\defres{ct128.scale.zero}{0.9--8.1}
\defres{ct128.scale.anchored}{0.83--1.15}
\defres{ct128.fa.anchored.but.ddnm}{1--6\%}
\defres{ct128.f1.better}{15}
\defres{ct128.f1.worse}{none}
\defres{ct128.fa.better}{13}
\defres{ct128.fa.worse}{none}
\defres{ct128.cov.better}{13}
\defres{ct128.cov.worse}{2}
\defres{ct128.amplitude.better}{15}
\defres{ct128.amplitude.worse}{none}
\defres{ct128.rmse.better}{5}
\defres{ct128.rmse.worse}{10}
\defres{ct128.auroc.better}{7}
\defres{ct128.auroc.worse}{none}
\defres{ct128.det5.better}{9}
\defres{ct128.det5.worse}{none}
\defres{ct128.teaser.only_anchored_detects}{167}
\defres{ct128.teaser.only_base_detects}{104}
\defres{ct128.teaser.only_base_flags}{476}
\defres{ct128.teaser.only_anchored_flags}{9}
\defres{ct128.kappa.64}{10}
\defres{ct128.nodraw.fa.64}{100\%}
\defres{ct128.posterior.f1.64}{$-$0.5}
\defres{ct128.posterior.f1ci.64}{[$-$2.6, $+$1.5]}
\defres{ct128.ddnm.fa.64}{2\%}
\defres{ct128.anchorflow.f1.64}{$+$0.7}
\defres{ct128.anchorflow.f1ci.64}{[$-$1.0, $+$2.2]}
\defres{ct128.kappa.256}{10}
\defres{ct128.nodraw.fa.256}{100\%}
\defres{ct128.posterior.f1.256}{$+$0.3}
\defres{ct128.posterior.f1ci.256}{[$-$2.1, $+$2.7]}
\defres{ct128.ddnm.fa.256}{4\%}
\defres{ct128.anchorflow.f1.256}{$+$1.7}
\defres{ct128.anchorflow.f1ci.256}{[$-$1.3, $+$4.7]}
\defres{ct128.observer.det5.64}{72\%}
\defres{ct128.observer.fa.64}{1.9\%}
\defres{ct128.gp.det5.64}{49\%}
\defres{ct128.gp.fa.64}{2.5\%}
\defres{ct128.gp.exact.det5.64}{59\%}
\defres{ct128.gp.exact.fa.64}{5.0\%}
\defres{ct128.fbp.det5.64}{22\%}
\defres{ct128.fbp.fa.64}{100.0\%}
\defres{ct128.det5.anchored.64}{48--53\%}
\defres{ct128.lock.fa.64}{100\%}
\defres{ct128.lock.det5.64}{39--51\%}
\defres{ct128.lockknown.fa.64}{9--100\%}
\defres{ct128.lockknown.det5.64}{33--59\%}
\defres{ct128.known.fa.64}{15--42\%}
\defres{ct128.known.det5.64}{47--57\%}
\defres{ct128.quality.gp.64}{72}
\defres{ct128.quality.anchorflow.64}{46}
\defres{ct128.quality.task.64}{44--47}
\defres{ct128.quality.ddrm.base.64}{77--108}
\defres{ct128.quality.ddrm.anchored.64}{146--189}
\defres{ct128.observer.det5.256}{86\%}
\defres{ct128.observer.fa.256}{2.2\%}
\defres{ct128.gp.det5.256}{73\%}
\defres{ct128.gp.fa.256}{4.4\%}
\defres{ct128.gp.exact.det5.256}{78\%}
\defres{ct128.gp.exact.fa.256}{7.2\%}
\defres{ct128.det5.anchored.256}{73--75\%}
\defres{ct128.lock.fa.256}{100\%}
\defres{ct128.lock.det5.256}{57--74\%}
\defres{ct128.lockknown.fa.256}{4--100\%}
\defres{ct128.lockknown.det5.256}{65--79\%}
\defres{ct128.known.fa.256}{4--66\%}
\defres{ct128.known.det5.256}{73--79\%}
\defres{ct128.quality.gp.256}{51}
\defres{ct128.quality.anchorflow.256}{35}
\defres{ct128.quality.task.256}{31--41}
\defres{ct128.quality.ddrm.base.256}{95--127}
\defres{ct128.quality.ddrm.anchored.256}{169--197}
\defres{ct128.sigma.est.64}{1.011}
\defres{ct128.sigma.est.range.64}{0.95--1.08}
\defres{ct128.problem.rows.64}{10485}
\defres{ct128.problem.rank.64}{10163}
\defres{ct128.sigma.est.256}{1.006}
\defres{ct128.sigma.est.range.256}{1.00--1.02}
\defres{ct128.problem.rows.256}{41997}
\defres{ct128.problem.rank.256}{16384}
\defres{ct128.seconds.anchorflow.64}{0.42}
\defres{ct128.seconds.trained.64}{0.43--0.71}
\defres{ct128.seconds.zero.64}{0.55--20}
\defres{ct128.seconds.ratio.64}{1.01--1.04}
\defres{ct128.seconds.anchorflow.256}{0.44}
\defres{ct128.seconds.trained.256}{0.45--0.71}
\defres{ct128.seconds.zero.256}{0.56--11}
\defres{ct128.seconds.ratio.256}{1.01--1.11}
\defres{ct128.train.anchorflow.64}{0.059}
\defres{ct128.train.anchorflow.256}{0.027}
\defres{ct128.train.cfm.64}{0.058}
\defres{ct128.train.cfm.256}{0.063}
\defres{ct128.train.edm.64}{0.102}
\defres{ct128.train.edm.256}{0.100}
\defres{ct128.draw.det5}{$-$11 to $+$15}
\defres{ct128.draw.det5.pairs}{12}
\defres{ct128.draw.det5.up}{4}
\defres{ct128.draw.det5.down}{8}
\defres{ct128.draw.det5.task}{$-$11 to $-$3}
\defres{ct128.draw.det5.task.pairs}{6}
\defres{ct128.draw.det5.task.up}{none}
\defres{ct128.draw.det5.task.down}{6}
\defres{ct128.draw.det5.zero}{$-$4 to $+$15}
\defres{ct128.draw.det5.zero.pairs}{6}
\defres{ct128.draw.det5.zero.up}{4}
\defres{ct128.draw.det5.zero.down}{2}
\defres{ct128.draw.fa}{$-$98 to $+$1}
\defres{ct128.draw.fa.pairs}{12}
\defres{ct128.draw.fa.up}{none}
\defres{ct128.draw.fa.down}{11}
\defres{ct128.draw.fa.task}{$-$7 to $+$1}
\defres{ct128.draw.fa.task.pairs}{6}
\defres{ct128.draw.fa.task.up}{none}
\defres{ct128.draw.fa.task.down}{5}
\defres{ct128.draw.fa.zero}{$-$98 to $-$40}
\defres{ct128.draw.fa.zero.pairs}{6}
\defres{ct128.draw.fa.zero.up}{none}
\defres{ct128.draw.fa.zero.down}{6}
\defres{ct128.lockknown.fa.task.64}{9--10\%}
\defres{ct128.lockknown.fa.zero.64}{41--100\%}
\defres{ct128.lockknown.fa.task.256}{4--9\%}
\defres{ct128.lockknown.fa.zero.256}{50--100\%}
\defres{ct128.anchorflow.vs.gp.64}{$+$1.9}
\defres{ct128.anchorflow.vs.gp.ci.64}{[$-$1.1, $+$4.4]}
\defres{ct128.anchorflow.vs.gp.256}{$+$0.4}
\defres{ct128.anchorflow.vs.gp.ci.256}{[$-$0.7, $+$1.8]}
\defres{ct128.readings.shared.fa}{28--100\%}
\defres{ct128.readings.own.fa}{0--8\%}
\defres{ct128.ddnm.ratio}{5--17}
\defres{ct128.ddnm.ratio.own}{0.99--1.11}
\defres{ct128.prop.samplers}{15}
\defres{ct128.prop.quantile}{1.7\%}
\defres{ct128.k16.det5}{48--50\%}
\defres{ct128.k64.det5}{55--59\%}
\defres{ct128.two.fa.task.base}{0\%}
\defres{ct128.two.fa.task.anchored}{0--1\%}
\defres{ct128.two.det5.task}{2.4 to 16.9}
\defres{ct128.two.f1.task}{3 to 27}
\defres{ct128.two.fa.zero.base}{35--100\%}
\defres{ct128.two.fa.zero.anchored}{0--4\%}
\defres{ct128.two.det5.zero}{$-$1.8 to $+$2.7}
\defres{ct128.two.f1.zero}{8 to 15}
\defres{ct128.two.fa.anchored.0.5mm}{0--4\%}
\defres{ct128.two.fa.base.0.5mm}{0--100\%}
\defres{ct128.shift.fa.anchored.0.5mm}{2--7\%}
\defres{ct128.two.fa.anchored.1mm}{1--8\%}
\defres{ct128.two.fa.base.1mm}{0--100\%}
\defres{ct128.shift.fa.anchored.1mm}{4--14\%}
\defres{ct128.two.fa.anchored.2mm}{17--36\%}
\defres{ct128.two.fa.base.2mm}{5--100\%}
\defres{ct128.shift.fa.anchored.2mm}{34--70\%}
\defres{ct128.two.gp.det5}{46\%}
\defres{ct128.two.observer.det5}{59\%}
\defres{ct128.two.anchorflow.det5}{48\%}
\defres{ct128.two.fbp.det5}{46\%}
\defres{ct128.two.anchorflow.vs.gp}{$+$1.7}
\defres{ct128.two.anchorflow.vs.gp.ci}{[$+$0.3, $+$4.1]}
\defres{ct128.recal.fa.base}{1--9\%}
\defres{ct128.recal.scale.base}{0.9--7.7}
\defres{ct128.recal.scale.anchored}{0.95--0.98}
\defres{ct128.recal.vs.task}{13 to 43}
\defres{ct128.recal.vs.task.pairs}{6}
\defres{ct128.recal.vs.task.sig}{6}
\defres{ct128.recal.vs.task.up}{6}
\defres{ct128.recal.vs.task.down}{none}
\defres{ct128.recal.vs.zero}{$-$3 to $+$9}
\defres{ct128.recal.vs.zero.pairs}{9}
\defres{ct128.recal.vs.zero.sig}{4}
\defres{ct128.recal.vs.zero.up}{2}
\defres{ct128.recal.vs.zero.down}{2}
\defres{ct128.recal.vs.all}{$-$3 to $+$43}
\defres{ct128.recal.vs.all.pairs}{15}
\defres{ct128.recal.vs.all.sig}{10}
\defres{ct128.recal.vs.all.up}{8}
\defres{ct128.recal.vs.all.down}{2}
\defres{ct128.sigma.samplers}{4}
\defres{ct128.sigma.low}{0.8}
\defres{ct128.sigma.high}{1.2}
\defres{ct128.sigma.fa.stated}{1.9--2.4\%}
\defres{ct128.sigma.det5.max}{1.5}
\defres{ct128.sigma.fa.low}{13.1--18.2\%}
\defres{ct128.sigma.f1.low}{$-$4 to $-$1}
\defres{ct128.sigma.fa.high}{0.0--0.4\%}
\defres{ct128.sigma.f1.high}{$-$6 to $-$5}
\defres{ct128.f1.pairs}{15}
\defres{ct128.f1.pairs.mismatch}{8}
\defres{ct128.held.base}{17--93\%}
\defres{ct128.held.anchored}{87--89\%}
\defres{ct128.fheld.median}{66--90\%}
\defres{ct128.aupro.task}{4 to 17}
\defres{ct128.aupro.zero}{$-$1 to $+$5}
\defres{ct128.pixel_auroc.task}{1 to 7}
\defres{ct128.pixel_auroc.zero}{0 to $+$1}
\defres{ct128.dice_max.task}{1 to 3}
\defres{ct128.dice_max.zero}{0 to 2}

\defres{tests.brain.images}{1362}
\defres{tests.brain.anomalous}{661}
\defres{tests.brain.free}{701}
\defres{tests.brain.patients}{60}
\defres{tests.brain.types}{4}
\defres{intro.brain.pairs}{16}
\defres{brain.rmse.gain}{1.6}
\defres{intro.brain.better}{11}
\defres{intro.brain.worse}{none}
\defres{brain.f1.zero.pairs}{12}
\defres{brain.f1.zero.better}{12}
\defres{brain.fa.task.base}{15--49\%}
\defres{brain.fa.task.anchored}{14--51\%}
\defres{brain.fa.zero.base}{16--100\%}
\defres{brain.fa.zero.anchored}{7--71\%}
\defres{brain.f1.better}{14}
\defres{brain.f1.worse}{none}
\defres{brain.fa.better}{13}
\defres{brain.fa.worse}{none}
\defres{brain.cov.better}{13}
\defres{brain.cov.worse}{3}
\defres{brain.rmse.better}{11}
\defres{brain.rmse.worse}{none}
\defres{brain.auroc.better}{12}
\defres{brain.auroc.worse}{none}
\defres{brain.finder.f1}{$-$2.3 to $+$1.6}
\defres{brain.finder.f1.sig}{6}
\defres{brain.finder.pairs}{16}
\defres{brain.readings.f1}{$-$0.7 to $+$0.9}
\defres{brain.readings.f1.sig}{none}
\defres{brain.readings.pairs}{16}
\defres{brain.recal.fa.base}{2--8\%}
\defres{brain.recal.scale.base}{1.1--3.4}
\defres{brain.recal.scale.anchored}{1.01--2.88}
\defres{brain.recal.vs.task}{$-$2 to $+$11}
\defres{brain.recal.vs.task.pairs}{6}
\defres{brain.recal.vs.task.sig}{3}
\defres{brain.recal.vs.task.up}{3}
\defres{brain.recal.vs.task.down}{none}
\defres{brain.recal.vs.zero}{$-$2 to $+$31}
\defres{brain.recal.vs.zero.pairs}{10}
\defres{brain.recal.vs.zero.sig}{5}
\defres{brain.recal.vs.zero.up}{5}
\defres{brain.recal.vs.zero.down}{none}
\defres{brain.recal.vs.all}{$-$2 to $+$31}
\defres{brain.recal.vs.all.pairs}{16}
\defres{brain.recal.vs.all.sig}{8}
\defres{brain.recal.vs.all.up}{8}
\defres{brain.recal.vs.all.down}{none}
\defres{brain.held.base}{31--91\%}
\defres{brain.held.anchored}{90\%}
\defres{brain.fheld.median}{76--87\%}
\defres{brain.aupro.task}{$-$3 to $+$10}
\defres{brain.aupro.zero}{6 to 29}
\defres{brain.pixel_auroc.task}{$-$3 to $+$7}
\defres{brain.pixel_auroc.zero}{4 to 26}
\defres{brain.dice_max.task}{$-$2 to $+$2}
\defres{brain.dice_max.zero}{2 to 23}

\defres{tests.pat.images}{156}
\defres{tests.pat.anomalous}{78}
\defres{tests.pat.free}{78}
\defres{tests.pat.patients}{27}
\defres{tests.pat.types}{2}
\defres{pat.fa.base}{73--100\%}
\defres{pat.fa.anchored}{3--5\%}
\defres{pat.f1.gain}{52 to 88}
\defres{pat.rmse.base}{49.4--52.1}
\defres{pat.rmse.anchored}{24.3--25.6}
\defres{pat.rmse.worse}{none}
\defres{pat.cov.base}{22--37\%}
\defres{pat.cov.anchored}{65--69\%}
\defres{pat.pairs}{3}
\defres{pat.f1.better}{3}
\defres{pat.rmse.better}{2}
\defres{pat.recal.fa.base}{1--15\%}
\defres{pat.recal.scale.base}{2.4--6.6}
\defres{pat.recal.scale.anchored}{0.90--1.05}
\defres{pat.recal.vs.task}{65 to 92}
\defres{pat.recal.vs.task.pairs}{2}
\defres{pat.recal.vs.task.sig}{2}
\defres{pat.recal.vs.task.up}{2}
\defres{pat.recal.vs.task.down}{none}
\defres{pat.recal.vs.zero}{$-$3}
\defres{pat.recal.vs.zero.pairs}{1}
\defres{pat.recal.vs.zero.sig}{1}
\defres{pat.recal.vs.zero.up}{none}
\defres{pat.recal.vs.zero.down}{1}
\defres{pat.recal.vs.all}{$-$3 to $+$92}
\defres{pat.recal.vs.all.pairs}{3}
\defres{pat.recal.vs.all.sig}{3}
\defres{pat.recal.vs.all.up}{2}
\defres{pat.recal.vs.all.down}{1}
\defres{pat.held.base}{39--90\%}
\defres{pat.held.anchored}{90\%}
\defres{pat.fheld.median}{82\%}
\defres{pat.aupro.task}{3 to 68}
\defres{pat.aupro.zero}{0}
\defres{pat.pixel_auroc.task}{1 to 30}
\defres{pat.pixel_auroc.zero}{0}
\defres{pat.dice_max.task}{$-$1 to $+$9}
\defres{pat.dice_max.zero}{0}

\defres{fig.teaser.readings}{64 views}
\defres{fig.teaser.method}{conditional FM}

\defres{fig.teaser.lesion.base}{0.98}
\defres{fig.teaser.lesion.anchored}{1.12}
\defres{fig.teaser.free.base}{1.18}
\defres{fig.teaser.free.anchored}{0.86}
\defres{fig.teaser.free.image}{2072}
\defres{fig.teaser.free.disagreements}{341}
\defres{fig.teaser.free.interior}{77}
\defres{fig.teaser.free.margin}{64}

\defres{brain.prior.exposure}{66\%}
\defres{brain.prior.ddrm.free}{5.9 points [4.8, 7.1]}
\defres{brain.prior.ddrm.exposed}{2.4 points [1.8, 3.1]}
\defres{brain.noise.ddrm.growth}{8.7 points [7.4, 9.9]}
\defres{brain.nfe.anchorflow.low}{5}
\defres{brain.nfe.anchorflow.full}{100}
\defres{brain.nfe.anchorflow.rmse.low}{12.54\%}
\defres{brain.nfe.anchorflow.rmse.full}{12.38\%}
\defres{ct.efficiency.f1.anchorflow}{56.0}
\defres{ct.efficiency.f1.cfm}{25.8}
\defres{ct.efficiency.f1.anchored_cfm}{56.8}
\defres{ct.efficiency.recovery.anchorflow}{79\%}
\defres{ct.efficiency.recovery.ddnm}{88\%}
\defres{ct.runtime.images}{432}
\defres{ct.runtime.patients}{45}
\defres{ct.runtime.anchorflow.nfe.ultra}{1}
\defres{ct.runtime.anchorflow.seconds.ultra}{5.9\,ms}
\defres{ct.runtime.anchorflow.rmse.ultra}{65.5\,HU}
\defres{ct.runtime.anchorflow.contrast-error.ultra}{26.8\%}
\defres{ct.runtime.anchorflow.coverage-error.ultra}{9.4 points}
\defres{ct.runtime.anchorflow.interval.ultra}{321.7\,HU}
\defres{ct.runtime.anchorflow.crps.ultra}{37.7\,HU}
\defres{ct.runtime.cfm.seconds.ultra}{5.1\,ms}
\defres{ct.runtime.cfm.rmse.ultra}{59.3\,HU}
\defres{ct.runtime.cfm.crps.ultra}{35.5\,HU}
\defres{ct.runtime.cfm.seconds.full}{0.436\,s}
\defres{ct.runtime.cfm.rmse.full}{60.1\,HU}
\defres{ct.runtime.cfm.contrast-error.full}{27.9\%}
\defres{ct.runtime.cfm.coverage-error.full}{2.5 points}
\defres{ct.runtime.cfm.interval.full}{280.7\,HU}
\defres{ct.runtime.cfm.crps.full}{33.7\,HU}
\defres{ct.runtime.anchorflow.vs-cfm-full.speedup}{$74\times$}
\defres{ct.runtime.anchorflow.vs-cfm-full.rmse-gap}{5.4\,HU}
\defres{ct.runtime.anchorflow.vs-cfm-full.contrast-gain}{1.1 points}
\defres{ct.runtime.anchorflow.vs-cfm-full.coverage-gap}{6.9 points}
\defres{ct.runtime.anchorflow.vs-cfm-full.interval-gap}{41.0\,HU}
\defres{ct.runtime.anchorflow.vs-cfm-full.crps-gap}{4.0\,HU}
\defres{ct.runtime.anchorflow.vs-best.rmse-relative}{8.9\%}
\defres{ct.runtime.anchorflow.vs-best.interval-relative}{14.6\%}
\defres{ct.runtime.anchorflow.vs-best.crps-relative}{11.8\%}
\defres{ct.runtime.anchorflow.vs-best.contrast-gap}{0.7 points}
\defres{ct.runtime.anchorflow.vs-diffpir-full.speedup}{$996\times$}
\defres{ct.runtime.ddrm.coverage-error.full}{0.04 points}
\defres{ct.runtime.anchorflow.vs-ddrm-full.speedup}{$99\times$}
\defres{ct.runtime.anchorflow.nfe.fast}{10}
\defres{ct.runtime.anchorflow.seconds.fast}{0.046\,s}
\defres{ct.runtime.anchorflow.rmse.fast}{65.8\,HU}
\defres{ct.runtime.anchorflow.contrast.fast}{88.3\%}
\defres{ct.runtime.anchorflow.coverage.fast}{87.5\%}
\defres{ct.runtime.anchorflow.contrast-error.fast}{27.0\%}
\defres{ct.runtime.anchorflow.coverage-error.fast}{2.5 points}
\defres{ct.runtime.anchorflow.speedup.fast}{$9.7\times$}
\defres{ct.runtime.anchorflow.nfe.calibrated}{20}
\defres{ct.runtime.anchorflow.seconds.calibrated}{0.090\,s}
\defres{ct.runtime.anchorflow.rmse.calibrated}{66.5\,HU}
\defres{ct.runtime.anchorflow.contrast.calibrated}{88.4\%}
\defres{ct.runtime.anchorflow.coverage.calibrated}{91.1\%}
\defres{ct.runtime.anchorflow.contrast-error.calibrated}{27.0\%}
\defres{ct.runtime.anchorflow.coverage-error.calibrated}{1.1 points}
\defres{ct.runtime.anchorflow.speedup.calibrated}{$4.9\times$}
\defres{ct.runtime.anchorflow.nfe.full}{100}
\defres{ct.runtime.anchorflow.seconds.full}{0.441\,s}
\defres{ct.runtime.anchorflow.rmse.full}{67.3\,HU}
\defres{ct.runtime.anchorflow.contrast.full}{88.4\%}
\defres{ct.runtime.anchorflow.coverage.full}{94.1\%}
\defres{ct.runtime.anchorflow.contrast-error.full}{27.1\%}
\defres{ct.runtime.anchorflow.coverage-error.full}{4.1 points}
\defres{ct.runtime.cfm.rmse.fast}{59.5\,HU}
\defres{ct.runtime.cfm.contrast.fast}{82.9\%}
\defres{ct.runtime.cfm.coverage.fast}{77.8\%}
\defres{ct.runtime.cfm.contrast-error.fast}{27.9\%}
\defres{ct.runtime.cfm.coverage-error.fast}{12.2 points}
\defres{brain.ref.oracle.gain}{9.1 to 40.8}
\defres{brain.ref.readings.det5}{73\%}
\defres{brain.ref.blind.det5}{63\%}
\defres{brain.ref.readings.truth.det5}{72.5\%}
\defres{brain.ref.blind.truth.det5}{63.1\%}
\defres{brain.ref.blind.contrast}{25\%}

\SIfreezecounter{section}{1}
\section{Implementation of selective anchoring}
\label{supp:method-details}
\label{supp:selection}

The ratio in \cref{eq:ratio} selects readings that are precise relative to normal-image variability. The Gaussian calculation in \cref{supp:gaussian} quantifies their shrinkage--variance trade-off. Preservation and analytic coverage concern the selected singular coordinates; pixels also contain generated free components.

\paragraph{Pretrained wrapper.}
The sampler uses the wrapped clean prediction in \cref{eq:wrap} with its original conditioning and projects the returned image to the same anchor. Intermediate noisy states need not satisfy this constraint. For a flow, the wrapped velocity is $(D'-x_t)/(1-t)$ for $t<1$. Additional data updates use only the remaining measurements, as detailed in \cref{supp:readings}.

\paragraph{AnchoredFM.}
\label{supp:AnchoredFM-training}
Training uses clean held coefficients and a Gaussian-surrogate maximum a posteriori (MAP) image of fresh noisy measurements as conditioning. Inference instead uses measurement-derived held inputs: fixed under deterministic anchoring or independently perturbed under stochastic anchoring. Both differ from the clean training inputs. Projected Euler integration preserves the held state exactly and requires no products with $A$ or $A\T$ after computing the lift and MAP image, avoiding the per-step cost of forward-model and adjoint evaluations.

\paragraph{Detection.}
\label{sec:anchored-detection}
The exact held-only null laws for the scores of \cref{eq:t} are proved in \cref{supp:proof-test}. Their Bonferroni guarantee requires those null laws; mixed templates and estimated references require empirical calibration. The practical MRI thresholds are specified in \cref{supp:fixed-reference}.

\SIfreezecounter{section}{2}
\section{Proofs}
\label{supp:proofs}
Write $\tilde y_i=z_i+e_i$, where $e_i=u_i\T\varepsilon/s_i\sim\N(0,n_i)$ and $n_i=\sigma^2/s_i^2$. Orthonormality of the $u_i$ makes the $e_i$ mutually independent and independent of the fixed image.

\SIfreezecounter{subsection}{1}
\subsection{Proof of Proposition 1}
Deterministic anchoring fixes each held coefficient at $\tilde y_i$. Stochastic anchoring adds an independent draw $e'_i\sim\N(0,n_i)$. Write $U_\Hs=[u_i]_{i\in\Hs}$ and $s_\Hs=(s_i)_{i\in\Hs}$ for the selected left singular vectors and singular values. For $\eta\sim\N(0,I_m)$ independent of acquisition noise,
\[
\operatorname{Cov}(\sigma\operatorname{diag}(s_\Hs)^{-1}U_\Hs\T\eta)
=\sigma^2\operatorname{diag}(s_\Hs)^{-2}.
\]
Both therefore give $\E[P_\Hs(\bar x-b)]=P_\Hs\ell$ over acquisition noise and, for stochastic anchoring, the target draws. Write $\Phi$ for the standard-normal CDF, $F_{t_\nu}$ for the Student CDF with $\nu>0$ degrees of freedom, and $t_{p,\nu}=F_{t_\nu}^{-1}(p)$ for $0<p<1$. For coverage level $\gamma\in(0,1)$, let $q_\gamma=\Phi^{-1}((1+\gamma)/2)$. The analytic interval $\tilde y_i\pm q_\gamma\sqrt{n_i}$ contains $z_i$ exactly when $|e_i|\leq q_\gamma\sqrt{n_i}$, an event of probability $\gamma$ for every fixed image. \hfill$\square$

\SIfreezecounter{subsection}{2}
\subsection{A Gaussian prior}
\label{supp:gaussian}
When $A\T A$ and $C$ commute, the ratios $r_i$ are likelihood-informed eigenvalues in a shared eigenbasis~\cite{cui2014lis,spantini2015optimal}.

\SIfreezecounter{proposition}{1}
\begin{proposition}
\label{supp:prop:gaussian}
Suppose also that the prior has independent $z_i\sim\N(m_i,p_i)$, where $m_i=v_i\T\mu$ and $p_i=v_i\T Cv_i>0$, and write $c_i=r_i/(1+r_i)$.
\begin{enumerate}[(a)]
\item In a held direction, the conditional variance of a stochastic anchored sample is $1/c_i=1+1/r_i<1+1/\kappa$ times the exact posterior variance. The same ratio holds for single-sample squared error averaged over the Gaussian prior, acquisition noise and sampling noise, but need not hold for a fixed image.
\item Given $z_i$, the exact posterior's standardized error has mean $\delta_i/\sqrt{1+r_i}$ and variance $c_i$, where $\delta_i=(z_i-m_i)/\sqrt{p_i}$. Its central intervals have nominal prior-averaged coverage, but need not have nominal coverage for a fixed component.
\end{enumerate}
\end{proposition}

\paragraph{Proof.}
The Gaussian posterior mean $M_i$ and variance satisfy
\[
\Var(z_i\mid y)=c_i n_i,\quad
M_i:=v_i\T\xmap(y)=m_i+c_i(\tilde y_i-m_i).
\]
The anchored draw has conditional variance $n_i$, giving (a). Its prior-averaged squared error is $2n_i$, while a posterior draw has error $2c_i n_i$ because truth and draw are conditionally independent posterior samples. For (b),
\[
z_i-M_i=(1-c_i)(z_i-m_i)-c_i e_i.
\]
Dividing by $\sqrt{c_i n_i}$ gives the stated fixed-component mean and variance. Averaging over $\delta_i\sim\N(0,1)$ gives total variance $1/(1+r_i)+c_i=1$. Components far from the prior mean, $|\delta_i|\gg\sqrt{1+r_i}$, can nevertheless lie outside posterior intervals more often than nominal. \hfill$\square$

\SIfreezecounter{subsection}{3}
\subsection{Detection laws and uncertainty treatments}
\label{supp:detector-analysis}
Assume \cref{eq:model}, the exact healthy reference $r=b$, $K\geq2$, and $w\neq0$. The template is fixed, so we omit its position argument in $a_j(p)$. Put $w_F=\Pi w$ and
\[
\tau^2=\sum_{i\in\Hs}n_i\frac{\langle v_i,w\rangle^2}{\langle w,w\rangle^2}.
\]
For a held-only template, stochastic anchoring under a fixed anomaly gives a noncentral $t_{K-1}$ score with
\[
\delta=\frac{\langle w,\ell\rangle}{\langle w,w\rangle\tau\sqrt{1+1/K}}.
\]
Under the null $\ell=0$, the following mixed-template Gaussian calculation is a diagnostic rather than a guarantee for a learned generator. Let uncentered free sample amplitudes be i.i.d.\ $\N(\mu_F,v_F)$ and the true free amplitude be an independent $\N(\mu_F,\lambda v_F)$ draw, independent also of held acquisition noise and anchor draws. With $v_F>0$, $\lambda\geq0$, and $\rho=v_F/(\tau^2+v_F)$, stochastic anchoring gives $t=\sqrt g\,T$, $T\sim t_{K-1}$, where
\[
g=1+\rho(\lambda-1)K/(K+1).
\]
Deterministic anchoring without analytic held variance instead is variance divided by expected squared denominator
\[
g_{\rm det}=1+\left((1-\rho)/\rho+\lambda-1\right)K/(K+1).
\]
Adding $\tau^2$ to $(1+1/K)s_a^2$ changes this ratio to
\[
g_{\rm corr}=1+\rho(\lambda-1)/(1+\rho/K).
\]
The corrected mixed denominator contains a constant plus a scaled chi-squared variable and does not establish an exact Student law. Held-only deterministic scores instead have the known-variance normal law proved below.

\paragraph{Thresholds and analytic-variance implementation.}
Let $\mathcal P$ be the finite set of valid position--size pairs, $N_{\rm tests}=|\mathcal P|$, and $\alpha=0.05$.
For held-only templates and an exact healthy reference, the two-sided Bonferroni rules are
\begin{equation*}
\begin{aligned}
 |t_{\rm stoch}|&>F^{-1}_{t_{K-1}}\!\left(1-\frac{\alpha}{2N_{\rm tests}}\right),\\
 |z_{\rm det}|&>\Phi^{-1}\!\left(1-\frac{\alpha}{2N_{\rm tests}}\right).
\end{aligned}
\end{equation*}
The held-template null laws give image-level false-alarm control by the union bound, without requiring independence across tests.

For the mixed-template CT analytic-variance comparison, write $a_j=a_j^H+a_j^F$. Let $s_F^2$ and $s_{HF}$ be the unbiased sample variance of $a_j^F$ and sample covariance of $a_j^H,a_j^F$.
Here $a_j^H=\langle P_\Hs w,x_j-r\rangle/\langle w,w\rangle$ and $a_j^F=\langle w_F,x_j-r\rangle/\langle w,w\rangle$ are scalar held and free residual amplitudes for sample $j=1,\ldots,K$.
The implemented estimated and total variances and effective degrees of freedom are
\begin{equation*}
\begin{aligned}
 v_{\rm est}&=\max\{(1+1/K)(s_F^2+2s_{HF}),0\},\\
 v_{\rm tot}&=v_{\rm est}+c_H\tau^2,\\
 \nu&=(K-1)\left(\frac{v_{\rm tot}}{v_{\rm est}}\right)^2\quad(v_{\rm est}>0),
\end{aligned}
\end{equation*}
where $c_H=1$ for deterministic anchoring and $c_H=1+1/K$ for stochastic anchoring with analytic held variance.
At fixed positive held variance, $v_{\rm est}\to0$ gives the known-variance normal limit.
Deterministic held amplitudes are constant across samples, so $s_{HF}=0$ and $v_{\rm est}=(1+1/K)s_a^2$.
CT maps $|\bar a|/\sqrt{v_{\rm tot}}$ through its $t_\nu$ cumulative distribution to the corresponding standard-normal quantile and uses the normal Bonferroni threshold above.
This Satterthwaite convention is an approximation for mixed templates; the exact laws concern held-only templates.
MRI instead uses the $t_{K-1}$ score normalization and healthy-validation-frozen thresholds specified in \cref{supp:fixed-reference}.

\SIfreezecounter{remark}{1}
\begin{remark}[Propagation of a held perturbation]
\label{supp:remark:propagation}
For fixed $y$, let $h$ be the random held anchor, $a^H$ its held residual amplitude, and $a^F=g_F(h,Z)$ the free residual amplitude, with independent generator randomness $Z$ and finite second moments. Total variance gives
\begin{align*}
\Var(a^H+a^F)={}&\Var(a^H)+\E[\Var(a^F\mid h)]\\
&+\Var(\E[a^F\mid h])+2\operatorname{Cov}(a^H,\E[a^F\mid h]).
\end{align*}
Generating at a fixed anchor and adding an independent held perturbation afterward does not reproduce the last two terms in general. The terms vanish for independent held and free frequencies in a stationary Gaussian Fourier prior, and may be small for low-noise held directions.
\end{remark}

\SIfreezecounter{subsection}{4}
\subsection{Proof of Proposition 2}
The Satterthwaite degrees of freedom are given in \cref{supp:detector-analysis}.
\label{supp:proof-test}
Write $a_j=\langle w,x_j-r\rangle/\langle w,w\rangle=a_j^H+a_j^F$ and $\beta_i=\langle v_i,w\rangle/\langle w,w\rangle$. Under the held-only null, independent acquisition and anchoring noises give
\[
a_j^H=\sum_{i\in\Hs}\beta_i(e_i+e'_{ij})=\tau(\epsilon_0+\epsilon_j),
\]
with independent standard normal $\epsilon_0,\epsilon_1,\ldots,\epsilon_K$; $e'_{ij}$ is sample $j$'s independent copy of $e'_i$. Set $\bar\epsilon=K^{-1}\sum_{j=1}^K\epsilon_j$ and $S^2=(K-1)^{-1}\sum_{j=1}^K(\epsilon_j-\bar\epsilon)^2$. Thus $\bar a=\tau(\epsilon_0+\bar\epsilon)$ and $s_a^2=\tau^2 S^2$. Cochran's theorem gives $(K-1)S^2\sim\chi^2_{K-1}$ independently of $\epsilon_0+\bar\epsilon$, whose variance is $1+1/K$. Hence $\bar a/(s_a\sqrt{1+1/K})\sim t_{K-1}$. An anomaly adds $\langle w,\ell\rangle/\langle w,w\rangle$, yielding the noncentrality above. Deterministic anchoring uses the fixed held readings, whose null amplitude is $\tau\epsilon_0$; division by known $\tau$ gives $\N(0,1)$.

For the mixed Gaussian diagnostic under the null, define uncentered free amplitudes $\widetilde a_j^F=\langle w_F,x_j\rangle/\langle w,w\rangle$ and $\widetilde a_\star^F=\langle w_F,x\rangle/\langle w,w\rangle$. Since $r=b=x$, the residual amplitude is $a_j^F=\widetilde a_j^F-\widetilde a_\star^F$, so $a_j=\tau(\epsilon_0+\epsilon_j)+\widetilde a_j^F-\widetilde a_\star^F$. The numerator variance is $\tau^2(1+1/K)+v_F/K+\lambda v_F$, while $(K-1)s_a^2/(\tau^2+v_F)$ is an independent $\chi^2_{K-1}$. Their ratio yields $g$. Deterministic anchoring removes $\epsilon_j$, leaving numerator variance $\tau^2+v_F/K+\lambda v_F$ and mean squared sample denominator $(1+1/K)v_F$. Adding $\tau^2$ to that denominator yields $g_{\rm corr}$. \hfill$\square$

\SIfreezecounter{section}{3}
\section{Metrics}
\label{supp:metrics}
Contrast is lesion-profile recovery relative to truth in CT and PAT, and tumor-to-ring contrast recovery in MRI; full recovery is 100\%. Reconstruction metrics are root-mean-square error (RMSE), in Hounsfield units (HU) for CT, and the structural similarity index (SSIM; \cref{supp:ssim}). At one pixel, let $x_\star$, $\bar x$ and $s_{\rm pix}$ denote its true intensity, ensemble mean and sample standard deviation (from the unbiased sample variance). For the labeled stochastic studies, its nominal 90\% interval covers the truth if $|x_\star-\bar x|<t_{0.95,K-1}\sqrt{1+1/K}\,s_{\rm pix}$. Lesion coverage (Cov.) averages over the expert MRI mask or the inserted lesion support above 5\% of peak magnitude in CT and photoacoustic tomography (PAT). Deterministic held-coordinate intervals instead use analytic acquisition variance (\cref{supp:held-coverage,supp:tab:held-cover}).

Sens.$_{\rm val}$ and FA$_{\rm val}$ denote sensitivity and false-alarm rate (FA) at the healthy-validation-frozen threshold; partial area under the curve (pAUC$_5$) is reported as mean localized sensitivity over 0--5\% FA. An image score is its maximum threshold-normalized $|t|$; the area under the receiver operating characteristic curve (AUROC) counts ties as one half. Det$_5$ uses a retrospective cutoff giving 5\% healthy-test FA. F1 uses the preset threshold and requires localization near the anomaly, with prevalence one anomalous to four healthy images: $F_1=2p_{\rm det}/(1+p_{\rm det}+4p_{\rm FA})$, where $p_{\rm det},p_{\rm FA}\in[0,1]$ are the localized anomalous-image detection rate and healthy-image false-alarm rate. It is averaged over anomaly types. Confidence intervals (CIs) use 2000 paired bootstrap draws of patients (anatomies for PAT). Metric differences are signed; intervals including zero do not establish an improvement. Free-response receiver operating characteristic (FROC) and practical MRI threshold and inference details are in \cref{supp:froc-size,supp:fixed-reference}.

\SIfreezecounter{section}{4}
\section{Samplers and settings}
\label{supp:baselines}
\paragraph{Shared prior as a denoiser.}
For the shared unanchored flow prior, $x_t=(1-t)x_0+tx_1$, where $x_0\sim\N(0,I_d)$ is independent of the clean image $x_1$ and $0<t\leq1$. Set $\sigma_{\rm den}=(1-t)/t$ and $x_{\rm den}=x_t/t=x_1+\sigma_{\rm den}x_0$. InverseBench~\SIcite{inversebench} receives $D_{\rm den}(x_{\rm den},\sigma_{\rm den})=D(t x_{\rm den},t)$, with $t=1/(1+\sigma_{\rm den})$ and the main text's clean predictor $D$. Here $\sigma_{\rm den}$ is denoising noise; acquisition noise remains $\sigma$. Samplers operate on $x-\mu$ and $y-A\mu$; $\mu$ is added back afterward.

\paragraph{Validation grids and adaptations.}
Settings are selected on the validation partition (part 0), sweeping DPS guidance $\{0.3,1,3,10,30,100\}$; $\Pi$GDM $\eta\in\{0,0.2,0.5,0.85,1\}$; DDRM $\eta\in\{0.5,0.85,1\}$; DDNM $\eta\in\{0.5,0.85\}$; DiffPIR $\lambda\in\{1/16,1/4,1,8,32\}$; P-DDIM $\eta_1\in\{1,4,16\}$; and EDM steps $\{18,40\}$. The baseline-specific $\eta$ controls injected sampling noise; P-DDIM's $\eta_0,\eta_1$ apply to observed and null-space directions, respectively. DiffPIR's $\lambda$ controls proximal regularization. cFM and AnchoredFM have no analogous main setting. PAT uses DPS guidance $\{1,3,10\}$, DDNM $\eta\in\{0.5,0.85\}$ and AnchoredFM $\kappa\in\{3,30\}$. DDRM retains $s_i \sigma_{\rm den,max}>\sigma$ as in its original code, where $\sigma_{\rm den,max}$ is the maximum denoising noise in the sampler schedule, distinct from the operator singular values $s_i$. DiffPIR uses its exact linear data step at noise $\sigma$. P-DDIM's CT correction is capped at the reading to prevent overshoot and divergence; this stabilizes sampling but leaves too little spread in strongly measured directions.

\SIfreezecounter{section}{5}
\section{Additional results}
\label{supp:tests}

\SIfreezecounter{subsection}{1}
\subsection{Teaser selection}
\label{supp:teaser}
For the main paper's opening figure (the teaser), the lesion example is the median true contrast among detections gained after anchoring. The healthy example is the median unanchored score among removed false alarms whose maximum is inside valid anatomy and at least \res{fig.teaser.free.margin} pixels from every edge. Both show full slices.

\SIfreezecounter{subsection}{2}
\subsection{CT}
\label{supp:problem}
CT256 and CT128 denote CT reconstruction on the 256 and 128 grids, respectively.
The twelve lesion types are 4, 8 and 16\,mm discs at $+50$, $+150$ and $+400$\,HU, an 8\,mm disc at $-100$\,HU, a 10\,mm ring and a 10\,mm lobulated lesion. The 256-grid, 64-view operator has \res{ct.problem.rows.64} rows and rank \res{ct.problem.rank.64}; measurements use the finer simulation described in \cref{supp:setup-details}.

\SIfreezecounter{subsection}{3}
\subsection{CT detection controls}
The Gaussian posterior has detection comparable to stochastic AnchoredFM (difference \res{ct.anchorflow.vs.gp.64} points \res{ct.anchorflow.vs.gp.ci.64}), trails deterministic AnchoredFM by about 6--11 Det$_5$ points across exact and sample-based tests, and has worse lesion RMSE and whole-image SSIM than cFM and AnchoredFM (\cref{supp:tab:ct}). In the stochastic CT128 control on part 1, increasing four anchored ensembles from 16 to 64 samples raises sensitivity from \res{ct128.k16.det5} to \res{ct128.k64.det5}.

\SIfreezecounter{table}{1}
\begin{table*}[!tp]
\caption{CT256 at 64 views: \res{tests.ct.anomalous} lesion and \res{tests.ct.free} healthy slices from \res{tests.ct.patients} patients, against the ideal reference. Left: default deterministic anchoring. Right: secondary stochastic comparison; arrows denote base$\to$stochastic, except Det$_5$, which is stochastic only. RMSE is in HU, SSIM dimensionless and other values in \%. Bold and italic denote better or worse paired results after Holm adjustment; deterministic detection columns and control rows are unmarked point estimates. Coverage is for nominal 90\% lesion-pixel intervals. AnchoredFM is compared with cFM.}
\label{supp:tab:ct}
\centering\footnotesize\setlength{\tabcolsep}{2.5pt}
\begin{minipage}[t]{0.55\textwidth}\centering
\textbf{Deterministic anchoring}\par\smallskip
\SIfitwidth{\linewidth}{
\begin{tabular}{lrrrrrr}
\toprule
Sampler & Lesion RMSE & Contrast & AUROC & Det$_5$ & F1 & FA \\
\midrule
cFM & 64.4 & 60.7 & 70.9 & 33.1 & 25.8 & 48.6 \\
\quad + anchoring & \textbf{59.1} & \textbf{78.2} & 85.5 & 66.8 & 62.1 & 7.1 \\
EDM & 63.7 & 62.3 & 72.5 & 34.9 & 30.2 & 30.7 \\
\quad + anchoring & \textbf{59.3} & \textbf{79.2} & 85.1 & 66.9 & 64.8 & 4.4 \\
DDRM & 65.2 & 81.2 & 82.5 & 59.1 & 31.9 & 100.0 \\
\quad + anchoring & \textbf{63.1} & \textbf{82.6} & 84.3 & 65.6 & 61.5 & 6.8 \\
DDNM & 81.5 & 85.3 & 81.3 & 59.6 & 32.4 & 100.0 \\
\quad + anchoring & \textbf{77.3} & \textbf{85.7} & 84.6 & 66.3 & 65.2 & 3.6 \\
DiffPIR & 61.1 & 74.9 & 83.0 & 55.9 & 36.8 & 44.3 \\
\quad + anchoring & \textbf{59.8} & \textbf{76.6} & 84.7 & 65.6 & 65.2 & 1.9 \\
P-DDIM & 59.0 & 72.2 & 82.0 & 57.7 & 28.2 & 77.8 \\
\quad + anchoring & 59.1 & \textbf{74.7} & 84.3 & 65.5 & 64.3 & 3.2 \\
AnchoredFM & \textbf{59.5} & \textbf{78.9} & 85.1 & 66.6 & 62.8 & 6.0 \\
\bottomrule
\end{tabular}

}
\end{minipage}\hfill
\begin{minipage}[t]{0.43\textwidth}\centering
\textbf{Stochastic diagnostics}\par\smallskip
\SIfitwidth{\linewidth}{
\begin{tabular}{lrrr}
\toprule
Sampler & SSIM & Cov. & Det$_5$ \\
\midrule
cFM & 0.871$\to$\textit{0.849} & 69.7$\to$\textbf{82.8} & \textbf{57.2} \\
EDM & 0.871$\to$\textit{0.849} & 76.7$\to$\textbf{87.3} & \textbf{59.2} \\
DDRM & 0.812$\to$\textit{0.806} & 63.4$\to$\textbf{87.8} & 55.5 \\
DDNM & 0.747$\to$\textit{0.740} & 61.2$\to$\textbf{91.7} & 57.2 \\
DiffPIR & 0.842$\to$\textbf{0.847} & 85.5$\to$\textbf{90.7} & 56.5 \\
P-DDIM & 0.856$\to$\textit{0.850} & 81.1$\to$\textbf{88.6} & 56.2 \\
AnchoredFM & 0.871$\to$\textit{0.846} & 69.7$\to$\textbf{85.6} & \textbf{57.3} \\
\bottomrule
\end{tabular}

}
\end{minipage}
\par\smallskip

\begin{tabular}{lrrrrrr}
\toprule
Control & Lesion RMSE & SSIM & Cov. & AUROC & Det$_5$ & FA \\
\midrule
Gaussian posterior & 68 & 0.724 & 93.7 & 82.0 & 55.2 & 2.8 \\
\quad exact test & -- & -- & -- & 82.3 & 60.7 & 9.7 \\
Ideal observer & -- & -- & -- & 89.7 & 76.4 & 0.8 \\
\bottomrule
\end{tabular}

\end{table*}

\SIfreezecounter{subsection}{4}
\subsection{Structural fidelity}
\label{supp:ssim}
SSIM~\SIcite{ssim} compares the ensemble mean with truth before display windowing, in native image units. It uses valid $11\times11$ Gaussian windows (standard deviation 1.5 pixels), $K_1=0.01$, $K_2=0.03$, and each true image's maximum-minus-minimum range. Whole-image values average all parts 1--3 slices, with paired patient-bootstrap comparisons in \cref{supp:tab:ct,supp:tab:brain-det-reconstruction} and a noise diagnostic in \cref{supp:tab:ssim-noise}.

\SIfreezecounter{subsection}{5}
\subsection{CT on the 128 grid}
\label{supp:ct128}
Settings transfer to CT256 from part 0 of the 128 grid (2\,mm pixels). \Cref{supp:tab:ct128} reports the 64-/256-view stochastic comparisons, including DPS and $\Pi$GDM. The adverse data-step reconstruction results use the CT128 noise convention in \cref{supp:readings}.
\SIfreezecounter{table}{2}
\begin{table*}[!tp]
\caption{Stochastic CT128, parts 1--3: \res{tests.ct128.anomalous} lesion and \res{tests.ct128.free} healthy slices from \res{tests.ct128.patients} patients. Subscripts give view count. RMSE is in HU; coverage, Det$_5$ and FA are in \%. Marks follow \cref{supp:tab:ct}; $\Pi$GDM at 256 views was not run because of cost.}
\label{supp:tab:ct128}
\centering\footnotesize\setlength{\tabcolsep}{3pt}

\begin{tabular}{lrrrrrrrr}
\toprule
Sampler & RMSE$_{64}$ & Cov.$_{64}$ & Det$_{5,64}$ & FA$_{64}$ & RMSE$_{256}$ & Cov.$_{256}$ & Det$_{5,256}$ & FA$_{256}$ \\
\midrule
cFM & 62 & 63.9 & 24.8 & 69.3 & 40 & 73.8 & 56.2 & 61.3 \\
\quad + anchoring & \textbf{57} & \textbf{81.3} & \textbf{48.3} & \textbf{4.4} & \textit{43} & \textbf{89.6} & \textbf{73.3} & \textbf{5.6} \\
EDM & 63 & 69.0 & 25.6 & 38.1 & 40 & 81.0 & 55.1 & 22.5 \\
\quad + anchoring & \textbf{57} & \textbf{84.6} & \textbf{48.5} & \textbf{3.9} & \textbf{38} & \textbf{86.8} & \textbf{73.8} & \textbf{4.3} \\
DPS & 75 & 71.7 & 19.0 & 1.7 & 65 & 72.6 & 28.4 & 4.3 \\
\quad + anchoring & \textbf{55} & \textbf{88.9} & \textbf{49.1} & 1.5 & \textbf{47} & \textbf{86.3} & \textbf{73.3} & \textbf{2.2} \\
$\Pi$GDM & 75 & 82.7 & 47.2 & 100.0 & -- & -- & -- & -- \\
\quad + anchoring & \textit{91} & \textbf{89.8} & 48.1 & \textbf{6.2} & -- & -- & -- & -- \\
DDRM & 81 & 53.4 & 51.2 & 100.0 & 97 & 22.0 & 71.7 & 100.0 \\
\quad + anchoring & \textit{144} & \textbf{93.5} & 48.8 & \textbf{2.4} & \textit{169} & \textbf{90.6} & 73.2 & \textbf{3.3} \\
DDNM & 111 & 49.9 & 48.7 & 100.0 & 131 & 14.7 & 73.0 & 100.0 \\
\quad + anchoring & \textit{187} & \textbf{94.1} & 48.1 & \textbf{2.1} & \textit{193} & \textbf{90.3} & 72.7 & \textbf{4.0} \\
DiffPIR & 61 & 84.0 & 46.3 & 51.4 & 44 & 92.7 & 66.9 & 1.0 \\
\quad + anchoring & \textit{69} & \textit{96.3} & \textbf{52.5} & \textbf{1.1} & \textit{47} & \textit{98.2} & \textbf{74.8} & 0.7 \\
P-DDIM & 58 & 79.4 & 47.0 & 80.1 & 51 & 81.5 & 75.0 & 95.7 \\
\quad + anchoring & \textit{65} & \textbf{95.8} & \textbf{51.7} & \textbf{1.1} & \textit{73} & \textbf{95.9} & 73.8 & \textbf{2.2} \\
AnchoredFM & \textbf{56} & \textbf{84.6} & \textbf{50.7} & \textbf{2.8} & \textbf{39} & \textbf{86.8} & \textbf{73.7} & \textbf{3.6} \\
Gaussian posterior & 61 & 94.6 & 48.8 & 2.5 & 45 & 94.2 & 73.3 & 4.4 \\
\quad exact test & -- & -- & 59.1 & 5.0 & -- & -- & 77.5 & 7.2 \\
Ideal observer & -- & -- & 71.6 & 1.9 & -- & -- & 85.6 & 2.2 \\
\bottomrule
\end{tabular}

\end{table*}

\SIfreezecounter{subsection}{6}
\subsection{Brain MRI and photoacoustics}
\label{supp:brain-pat}
Deterministic MRI fidelity and analytic held coverage are reported in \cref{supp:tab:brain-det-reconstruction,supp:held-coverage}. \Cref{supp:tab:brain-full,supp:tab:pat} report lesion-pixel coverage for the secondary stochastic MRI and PAT comparisons. Practical detection uses the reference protocol of \cref{supp:fixed-reference}.

\SIfreezecounter{table}{3}
\begin{table*}[!tp]
\caption{Nominal 90\% tumor-pixel coverage (\%) under stochastic anchoring in 64-line MRI: 661 tumor and 701 healthy slices from 60 patients. This secondary P-DDIM comparison uses its implicit default $\eta_0$, whereas the primary matched cohort uses $\eta_0=0$. Marks indicate movement toward or away from 90\% after Holm adjustment. AnchoredFM is compared with cFM.}
\label{supp:tab:brain-full}
\centering\footnotesize

\begin{tabular}{lrr}
\toprule
Sampler & Base & Stochastic \\
\midrule
cFM & 62.9 & \textbf{71.2} \\
EDM & 77.9 & \textbf{79.2} \\
DPS & 85.1 & \textit{84.2} \\
$\Pi$GDM & 71.3 & \textbf{72.1} \\
DDRM & 61.7 & \textbf{72.0} \\
DDNM & 67.6 & \textbf{72.7} \\
DiffPIR & 54.6 & \textbf{68.2} \\
P-DDIM & 53.0 & \textbf{59.2} \\
AnchoredFM & 62.9 & \textbf{68.2} \\
\bottomrule
\end{tabular}

\end{table*}

\SIfreezecounter{table}{4}
\begin{table*}[!tp]
\caption{MRI reconstruction at 64 of 256 lines: default deterministic anchoring and a 16-sample Gaussian-posterior control. RMSE (\% of median brain intensity) and contrast metrics (\%) use 661 tumor slices; whole-image SSIM uses all 1362 test slices. Bold and italic denote better or worse paired reconstruction after Holm adjustment; contrast-recovery marks use absolute contrast error. AnchoredFM is compared with cFM; the Gaussian row is unmarked.}
\label{supp:tab:brain-det-reconstruction}
\centering\small\setlength{\tabcolsep}{5pt}

\begin{tabular}{lrrrr}
\toprule
Sampler & Tumor RMSE & Contrast recovery & Contrast error & SSIM \\
\midrule
cFM & 13.9 & 88.9 & 11.1 & 0.947 \\
\quad + anchoring & \textbf{12.2} & \textbf{91.8} & \textbf{8.4} & \textit{0.741} \\
EDM & 11.7 & 90.7 & 9.4 & 0.954 \\
\quad + anchoring & \textbf{10.8} & \textbf{93.0} & \textbf{7.1} & \textit{0.749} \\
DPS & 11.0 & 92.9 & 7.2 & 0.946 \\
\quad + anchoring & 11.0 & \textit{92.4} & \textit{7.7} & \textit{0.886} \\
$\Pi$GDM & 10.9 & 92.5 & 7.6 & 0.953 \\
\quad + anchoring & \textbf{10.8} & 92.5 & 7.6 & \textit{0.855} \\
DDRM & 11.6 & 90.8 & 9.3 & 0.946 \\
\quad + anchoring & \textbf{11.2} & \textbf{92.4} & \textbf{7.8} & \textit{0.782} \\
DDNM & 11.1 & 93.1 & 7.0 & 0.944 \\
\quad + anchoring & 11.0 & \textbf{93.6} & \textbf{6.6} & \textit{0.814} \\
DiffPIR & 11.0 & 92.4 & 7.7 & 0.952 \\
\quad + anchoring & \textbf{10.6} & \textbf{93.9} & \textbf{6.3} & \textit{0.788} \\
P-DDIM & 11.0 & 91.8 & 8.3 & 0.954 \\
\quad + anchoring & \textbf{10.8} & \textbf{92.8} & \textbf{7.3} & \textit{0.815} \\
AnchoredFM & \textbf{12.3} & \textbf{90.8} & \textbf{9.3} & \textit{0.729} \\
Eight-sampler mean & 11.5 & 91.6 & 8.4 & 0.950 \\
\quad + anchoring & 11.0 & 92.8 & 7.3 & 0.804 \\
\midrule
Gaussian posterior & 16.9 & 84.8 & 15.5 & 0.546 \\
\bottomrule
\end{tabular}

\end{table*}

\SIfreezecounter{table}{5}
\begin{table}[!htbp]
\caption{Analytic held-coordinate coverage (\%) of nominal 90\% intervals in 64-line MRI. Intervals use known acquisition variance; patient-bootstrap 95\% intervals assess this coordinate-level acquisition diagnostic.}
\label{supp:held-coverage}
\centering\scriptsize\setlength{\tabcolsep}{3pt}
\SIfitwidth{\linewidth}{
\begin{tabular}{lrrrr}
\toprule
$\kappa$ & Held directions & All slices [95\% CI] & Tumor slices & Healthy slices \\
\midrule
1 & 4833 & 90.01 [89.99, 90.03] & 89.99 & 90.02 \\
3 & 2663 & 90.00 [89.97, 90.03] & 90.00 & 90.00 \\
10 & 1401 & 90.00 [89.96, 90.04] & 89.98 & 90.02 \\
30 & 757 & 90.02 [89.96, 90.08] & 90.00 & 90.04 \\
100 & 347 & 90.06 [89.98, 90.14] & 90.05 & 90.07 \\
\bottomrule
\end{tabular}

}
\end{table}

\SIfreezecounter{table}{6}
\begin{table*}[!tp]
\caption{Stochastic PAT at 64 detectors: \res{tests.pat.anomalous} lesion phantoms and healthy twins from \res{tests.pat.patients} anatomies. Each entry is base$\to$anchored; AnchoredFM uses cFM as its base. RMSE is in \% of median initial pressure; other values are in \%, with marks as in \cref{supp:tab:ct}. Coverage is for nominal 90\% lesion-pixel intervals.}
\label{supp:tab:pat}
\centering\footnotesize\setlength{\tabcolsep}{5pt}

\begin{tabular}{lrrrr}
\toprule
Sampler & Lesion RMSE & Cov. & Det$_5$ & FA \\
\midrule
cFM & 49.4$\to$\textbf{24.3} & 22.3$\to$\textbf{64.7} & 20.1$\to$\textbf{98.1} & 73.1$\to$\textbf{5.1} \\
DPS & 52.1$\to$\textbf{25.6} & 37.3$\to$\textbf{69.4} & 0.0$\to$\textbf{98.1} & 100.0$\to$\textbf{3.8} \\
DDNM & 22.8$\to$22.4 & 24.8$\to$\textbf{69.5} & 99.4$\to$98.7 & 100.0$\to$\textbf{2.6} \\
AnchoredFM & 49.4$\to$\textbf{24.1} & 22.3$\to$\textbf{64.8} & 20.1$\to$\textbf{98.7} & 73.1$\to$\textbf{5.1} \\
\bottomrule
\end{tabular}

\end{table*}

\SIfreezecounter{subsection}{7}
\subsection{Tumor exposure in BraTS training}
\label{supp:brain-prior}
This secondary stochastic control trains separate models with tumors in \res{brain.prior.exposure} of training slices, keeping training-set size, patient pool, flips, readings, selection surrogate and sampler settings fixed. Exposure improves unread-component recovery and generally reduces the additional contrast benefit of anchoring. At four-times nominal noise, anchoring increases tumor RMSE for all four tested samplers despite better pixel coverage (\cref{supp:tab:brain-prior}). Tumor contrast is mask mean minus the surrounding four-pixel-ring mean. Unread recovery is the pooled coefficient inner product with the true change, relative to mask-inpainted references, divided by the true change's pooled squared norm.
At nominal noise with the exposed prior, AnchoredFM's contrast change is $-1.5$ points $[-2.0,-1.0]$ and its tumor RMSE rises by $0.5\%$ of median brain intensity (\cref{supp:tab:brain-prior}).

\SIfreezecounter{table}{7}
\begin{table*}[!tp]
\caption{Stochastic-anchoring robustness comparison: training on slices with tumors in the 64-line BraTS test. Noise standard deviation $\sigma$ is relative
  to median brain intensity. Contrast changes are in percentage points, with paired 95\% patient-bootstrap
  intervals. Exposure is tumor-exposed minus tumor-free for the base
  sampler. The anchoring columns are anchored minus base under the stated prior. Unread recovery of the base
  sampler is shown before and after tumor exposure, in \%. With exposure, tumor RMSE changes are in \% of
  median brain intensity. Coverage of 90\% tumor-pixel intervals is shown before and after anchoring. The
  AnchoredFM row compares it with cFM.}
\label{supp:tab:brain-prior}
\centering\footnotesize\setlength{\tabcolsep}{2.5pt}
\SIfitwidth{\textwidth}{
\begin{tabular}{@{}rlcccccc@{}}
\toprule
& & \multicolumn{3}{c}{Change in contrast recovery} & & \multicolumn{2}{c}{Tumor-exposed prior} \\
\cmidrule(lr){3-5}\cmidrule(lr){7-8}
$\sigma$ & Sampler & Exposure & Anchor, tumor-free & Anchor, exposed & Unread recovery & $\Delta$ tumor RMSE & Cov. \\
\midrule
0.088 & DDRM & $+2.4\;[1.9,\,2.8]$ & $+1.4\;[1.1,\,1.7]$ & $+1.0\;[0.8,\,1.2]$ & $59\to66$ & $-0.1\;[-0.1,\,0.0]$ & $67\to77$ \\
 & DDNM & $+2.3\;[2.0,\,2.7]$ & $+0.4\;[0.3,\,0.5]$ & $+0.1\;[0.0,\,0.2]$ & $70\to76$ & $+0.2\;[0.1,\,0.2]$ & $72\to77$ \\
 & $\Pi$GDM & $+2.4\;[2.0,\,2.8]$ & $0.0\;[-0.2,\,0.1]$ & $-0.2\;[-0.3,\,0.0]$ & $68\to74$ & $+0.1\;[0.1,\,0.1]$ & $74\to76$ \\
 & Posterior-DDIM & $+2.4\;[2.0,\,2.8]$ & $+0.6\;[0.4,\,0.7]$ & $+0.1\;[0.0,\,0.3]$ & $67\to74$ & $+0.1\;[0.0,\,0.1]$ & $58\to64$ \\
 & cFM & $+5.0\;[4.2,\,5.9]$ & $+2.7\;[2.3,\,3.2]$ & $+1.4\;[1.1,\,1.6]$ & $57\to72$ & $-0.7\;[-0.9,\,-0.5]$ & $78\to82$ \\
 & AnchoredFM & $+5.0\;[4.2,\,5.9]$ & $+1.5\;[0.9,\,2.0]$ & $-1.5\;[-2.0,\,-1.0]$ & $57\to72$ & $+0.5\;[0.2,\,0.9]$ & $78\to80$ \\
\addlinespace
0.352 & DDRM & $+5.8\;[5.0,\,6.6]$ & $+5.9\;[4.8,\,7.1]$ & $+2.4\;[1.8,\,3.1]$ & $42\to50$ & $+0.7\;[0.6,\,0.8]$ & $55\to74$ \\
 & DDNM & $+4.0\;[3.5,\,4.5]$ & $+1.8\;[1.3,\,2.3]$ & $+0.2\;[-0.2,\,0.6]$ & $55\to63$ & $+0.4\;[0.2,\,0.6]$ & $62\to74$ \\
 & $\Pi$GDM & $+6.5\;[5.5,\,7.5]$ & $+5.9\;[4.6,\,7.3]$ & $+1.9\;[1.2,\,2.6]$ & $47\to55$ & $+0.6\;[0.5,\,0.7]$ & $68\to79$ \\
 & Posterior-DDIM & $+5.9\;[5.2,\,6.5]$ & $+4.9\;[4.1,\,5.7]$ & $+1.2\;[0.7,\,1.7]$ & $51\to59$ & $+0.3\;[0.3,\,0.4]$ & $48\to69$ \\
\bottomrule
\end{tabular}

}
\end{table*}

\SIfreezecounter{subsection}{8}
\subsection{Acquisition noise in BraTS}
\label{supp:brain-noise}
The tumor-free-prior stochastic sweep scales the same Gaussian draw per slice at 1, 2, 4 and 8 times nominal noise, with the true level supplied to each sampler. Base settings and $\kappa$ minimize tumor RMSE on part 0 and are frozen for testing; wrappers retain base settings. This secondary sweep uses independently selected P-DDIM settings with implicit default $\eta_0$. \Cref{supp:tab:brain-noise} gives the nominal and eight-times endpoints. All four samplers gain contrast at eight-times noise after Holm adjustment, with larger gains than at nominal noise; DDRM's contrast gain grows from nominal to eight-times noise by \res{brain.noise.ddrm.growth}. At eight-times noise tumor RMSE falls for all four, while normal-tissue RMSE rises for three. Coverage improves but remains below 90\%.
\SIfreezecounter{table}{8}
\begin{table*}[!tp]
\caption{Stochastic MRI noise sweep endpoints. Contrast and coverage are in \%; changes are anchored minus base. Contrast gains have paired patient-bootstrap 95\% intervals; stars denote Holm-adjusted $p<0.05$ across four samplers at that noise level. Coverage is base$\to$anchored. RMSE changes are in \% of median brain intensity; normal tissue excludes a four-pixel tumor margin.}
\label{supp:tab:brain-noise}
\centering\scriptsize\setlength{\tabcolsep}{3pt}

\begin{tabular}{lrrrrrrr}
\toprule
$\sigma$ & Sampler & $\kappa$ & Base contrast & $\Delta$ contrast & Cov. & $\Delta$ RMSE$_{\rm tumor}$ & $\Delta$ RMSE$_{\rm normal}$ \\
\midrule
0.088 & DDRM & 3 & 90.8 & $+1.4\;[1.1,\,1.7]$\textsuperscript{*} & 61.7$\to$72.0 & -0.30 & +0.12 \\
 & DDNM & 10 & 93.1 & $+0.4\;[0.3,\,0.5]$\textsuperscript{*} & 67.6$\to$72.7 & +0.04 & +0.23 \\
 & $\Pi$GDM & 30 & 92.5 & $0.0\;[-0.2,\,0.1]$ & 71.3$\to$72.1 & -0.06 & +0.04 \\
 & Posterior-DDIM & 10 & 92.4 & $+0.6\;[0.4,\,0.7]$\textsuperscript{*} & 53.0$\to$59.2 & -0.09 & +0.12 \\
0.704 & DDRM & 1 & 71.5 & $+10.0\;[8.7,\,11.5]$\textsuperscript{*} & 38.6$\to$67.2 & -1.46 & +1.30 \\
 & DDNM & 1 & 83.5 & $+1.8\;[1.1,\,2.6]$\textsuperscript{*} & 52.8$\to$63.6 & -1.80 & -1.10 \\
 & $\Pi$GDM & 1 & 81.9 & $+3.2\;[2.4,\,4.2]$\textsuperscript{*} & 82.1$\to$84.3 & -0.99 & +0.10 \\
 & Posterior-DDIM & 1 & 74.6 & $+8.6\;[7.7,\,9.6]$\textsuperscript{*} & 65.7$\to$75.1 & -1.73 & +0.18 \\
\bottomrule
\end{tabular}

\end{table*}

\SIfreezecounter{section}{6}
\section{Detection controls and robustness}
\label{supp:froc-size}
\begin{figure*}[p]
\centering
\includegraphics[width=0.9\textwidth]{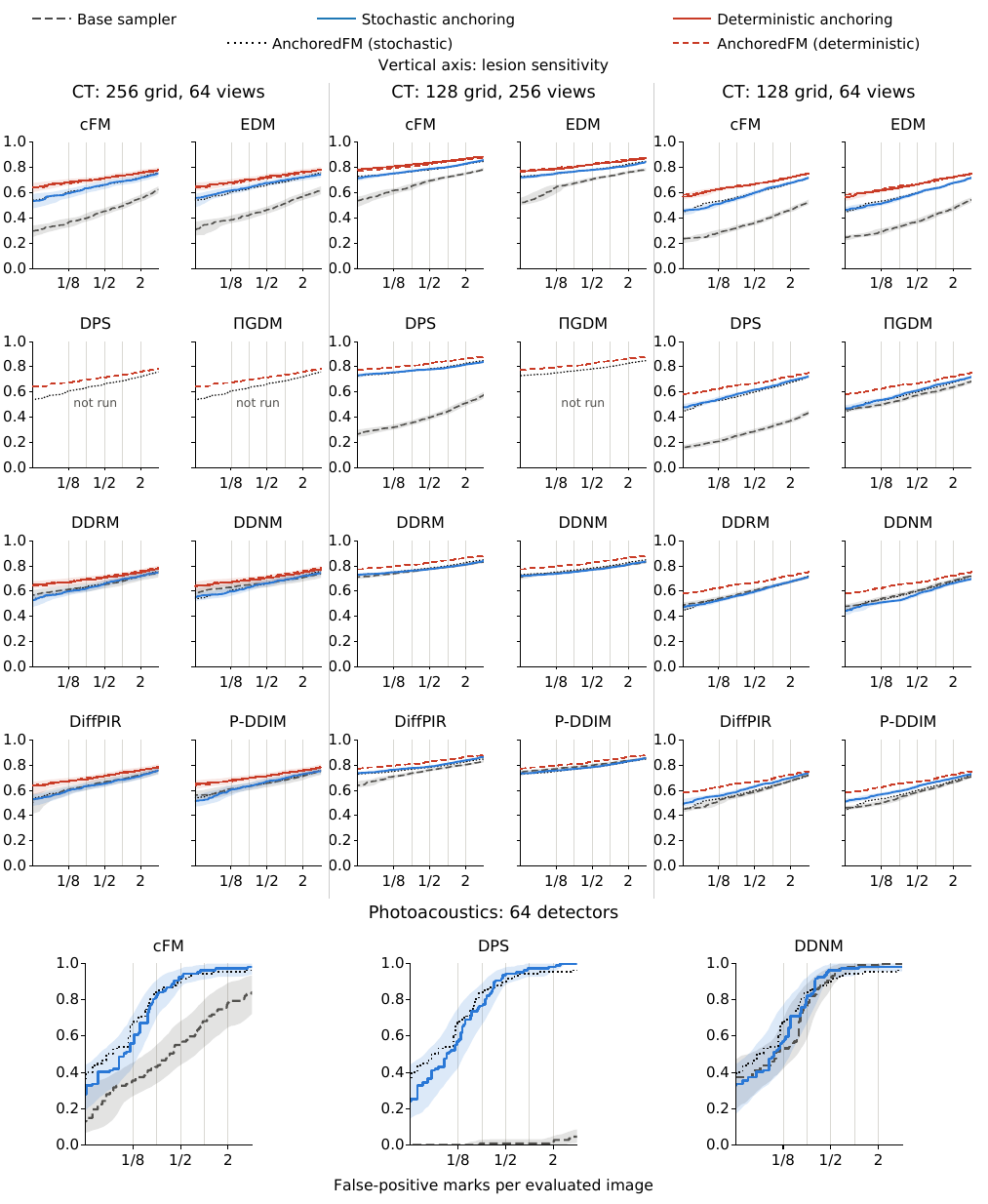}
\caption{CT and PAT FROC curves. Deterministic CT scores include analytic held variance (\cref{supp:detector-analysis}): AnchoredFM and six wrappers for CT256, AnchoredFM, cFM and EDM for CT128. PAT shows stochastic anchoring. Bands are 95\% patient-bootstrap (CT) or anatomy-bootstrap (PAT) intervals; burdens are per evaluated image.}
\label{supp:fig:froc}
\end{figure*}

\paragraph{FROC protocol.}
FROC uses threshold-normalized score maps. Candidate marks are $3\times3$ local maxima, retained in descending score order up to 32 per image. After each mark, candidates within a center-to-center distance of 16\,mm (CT), 8\,mm (PAT) or 32\,mm (MRI) are suppressed; these are radii of the exclusion discs. Truth components widened by the original localization tolerance define lesions. Greedy one-to-one matching counts unmatched and duplicate marks as false positives. Sensitivity is matched lesions divided by all lesions; burden is false marks per evaluated image, or selected 2D slice in MRI, not per 3D scan. MRI tables report 0.25 marks per slice; CT and PAT curves span 0.125--2 marks per image. Intervals use 2000 paired patient or anatomy bootstrap draws (\cref{supp:fig:froc}).

\SIfreezecounter{table}{9}
\begin{table*}[!tp]
\caption{CT calibration controls at 64 views: each entry is median preset FA / Det$_5$ (\%) across samplers with or without an explicit measurement update. Healthy-part-0 spread rescaling changes preset alarms but leaves ranking and Det$_5$ unchanged. Deterministic anchoring includes analytic held variance. Gaussian and ideal controls use the same exact healthy reference.}
\label{supp:tab:summary}
\centering\small\setlength{\tabcolsep}{5pt}

\begin{tabular}{lrrrrrr}
\toprule
Base sampler & Base & Rescaled & Det. + var. & Stochastic & Gaussian & Ideal \\
\midrule
No data step & 40/34 & 6/34 & 6/67 & 2/58 & 3/55 & 1/76 \\
Data step & 89/58 & 6/58 & 3/66 & 2/56 & 3/55 & 1/76 \\
\bottomrule
\end{tabular}

\end{table*}

\SIfreezecounter{figure}{1}

\SIfreezecounter{subsection}{1}
\subsection{Sensitivity by lesion size}
Every run uses one global FROC threshold across size groups. The primary deterministic checkerboard results are in \cref{tab:size,supp:tab:checkerboard-size64}; the eight-sampler gain at 64 lines is 8.8 points at 0.25 false marks per slice, with low absolute sensitivity for the smallest tumors.

EDM's overall interval and several per-group intervals, including the smallest-group intervals for cFM, EDM, DiffPIR and P-DDIM, include zero (\cref{supp:tab:checkerboard-size64}).

\SIfreezecounter{subsection}{2}
\subsection{Photoacoustic operator error}
\label{supp:pat-error}
In this stochastic failure control, the 128-grid reconstruction operator cannot reproduce skin and vessel detail in 256-grid simulated slabs. Operator error is about ten times the assumed noise, and anchoring worsens lesion reconstruction for all three samplers. Simulating on the reconstruction grid reduces the mismatch. The main 256-grid experiment instead simulates with four-times denser per-pixel quadrature; its operator error is one fifth of the noise.

\SIfreezecounter{subsection}{3}
\subsection{Noise misspecification}
\label{supp:sigma}
Four stochastic CT128 samplers at 64 views use validation-F1 settings and shared data-step readings. With dimensionless assumed-noise multiplier $q_\sigma>0$, the wrapper selects $r_i/q_\sigma^2>\kappa$ and adds $q_\sigma\sigma\eta_j$; AnchoredFM keeps its trained held set and scales the perturbation. Other steps retain the true noise. At $q_\sigma=0.8$, healthy FA rises to \res{ct128.sigma.fa.low} and F1 changes by \res{ct128.sigma.f1.low} points; at $q_\sigma=1.2$, FA falls and F1 changes by \res{ct128.sigma.f1.high} points. Matched-rate sensitivity changes by at most \res{ct128.sigma.det5.max} points. Separately, the CT256 residual outside the range of $A$ estimates noise at \res{ct.sigma.est.64} times truth, with 90\% of images in \res{ct.sigma.est.range.64}.

\SIfreezecounter{subsection}{4}
\subsection{AnchoredFM ablation}
\label{supp:ablations}
The same network without a stochastic draw \emph{and without analytic held variance} flags \res{ct.nodraw.fa.64} of healthy CT256 images (\cref{supp:tab:ablations}). The comparison with deterministic anchoring using analytic held variance appears in \cref{supp:tab:draw}.

\SIfreezecounter{table}{10}
\begin{table*}[!tp]
\caption{Ablations of stochastic AnchoredFM in CT on the 256 grid, 64 views, parts 1--3, in \%: the same network without the noise draw and without analytic held variance,
  compared at AnchoredFM's $\kappa$; cFM with and without the anchoring wrapper. Where shown, bold (italic) marks better (worse) paired results than
  AnchoredFM; unmarked entries have no reported significance designation.}
\label{supp:tab:ablations}
\centering\footnotesize\setlength{\tabcolsep}{3pt}
\SIfitwidth{\linewidth}{
\begin{tabular}{lrrr}
\toprule
 & \multicolumn{3}{c}{64 views} \\
\cmidrule(lr){2-4}
 & Cov. & F1 & FA \\
\midrule
AnchoredFM & 85.6 & 56.0 & 1.7 \\
\quad no draw & \textit{78.4} & \textit{29.6} & \textit{99.3} \\
\midrule
cFM + anchoring & 82.8 & 56.8 & 2.6 \\
cFM & 69.7 & 25.8 & 48.6 \\
\bottomrule
\end{tabular}

}
\end{table*}

\SIfreezecounter{subsection}{5}
\subsection{Held directions of the operators}
\label{supp:spectra}
\begin{figure*}[!tp]
\begin{minipage}[t]{0.48\textwidth}
\vspace{0pt}
\SIfreezecounter{figure}{2}
\centering
\includegraphics[width=\linewidth]{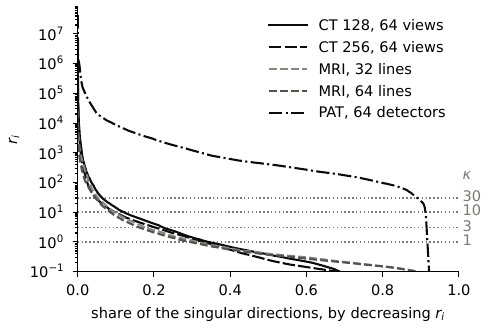}

\caption{Information ratio $r_i$, in decreasing order, with cutoffs $\kappa=1,3,10,30$.}
\label{supp:fig:spectra}
\end{minipage}\hfill
\begin{minipage}[t]{0.48\textwidth}
\vspace{0pt}
\SIfreezecounter{figure}{3}
\centering
\includegraphics[width=\linewidth]{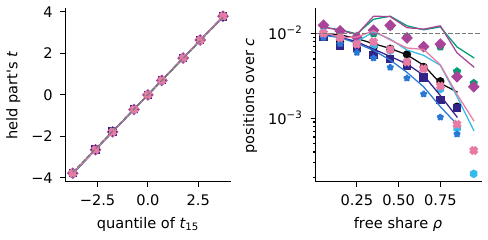}

\caption{Secondary stochastic CT diagnostic. Left: held-template score quantiles versus $t_{15}$. Right: exceedance of the per-position 1\% cutoff $c=F^{-1}_{t_{15}}(0.995)$ versus free variance share $\rho$, with the Gaussian diagnostic's predictions (lines) and empirical rates (markers). Markers: star, AnchoredFM; circle, cFM; square, EDM; diamond, DDRM; pentagon, DDNM; hexagon, DiffPIR; cross, P-DDIM.}
\label{supp:fig:held}
\end{minipage}
\end{figure*}

The information-ratio spectra in \cref{supp:fig:spectra} show how many singular directions each cutoff selects across CT, MRI and PAT.

\SIfreezecounter{section}{7}
\section{Anchoring variants and reference controls}
\label{supp:controls}

\SIfreezecounter{subsection}{1}
\subsection{Held uncertainty treatments}
\label{supp:draw}
\Cref{supp:tab:draw} compares deterministic and stochastic anchoring with sample-estimated or analytic held variance under common settings. The analytic correction changes the detector, not reconstructed means or free components. Without it, cFM, EDM and AnchoredFM flag \res{ct.lock.fa.64} of healthy slices; with it, they flag \res{ct.lockknown.fa.task.64} and obtain higher Det$_5$ than stochastic anchoring. For the four data-step samplers, deterministic FA is 1.9--6.8\% and Det$_5$ 65.5--66.3\%, versus stochastic Det$_5$ 55.5--57.2\%. DDRM's deterministic FA is 6.8\% [5.0,8.8], at or above the 5\% target. Adding analytic held variance to stochastic anchoring gives 16--47\% FA. Unresolved-component uncertainty remains empirical.

As a simpler control, healthy-validation spread rescaling (\cref{supp:tab:summary}) gives base CT FA of \res{ct.recal.fa.base}; it does not consistently meet 5\%, alter ranking, or establish coverage at other thresholds. Held-only stochastic null scores agree with $t_{15}$ (\cref{supp:fig:held}); mixed scores depend on free uncertainty as in \cref{supp:detector-analysis}.

\SIfreezecounter{table}{11}
\begin{table*}[!tp]
\caption{CT on the 256 grid, 64 views, parts 1--3: deterministic anchoring (every sample holds $\lift(y)$) and stochastic anchoring, with held variance estimated from samples or supplied analytically as $\tau^2$.
  F1, false alarms and Det$_5$ in \%.
  Where shown, bold (italic) marks better (worse) paired results than deterministic anchoring + analytic variance, Holm-adjusted over samplers; unmarked entries have no reported significance designation.}
\label{supp:tab:draw}
\centering\footnotesize\setlength{\tabcolsep}{2.5pt}
\SIfitwidth{\textwidth}{
\begin{tabular}{l*{12}{r}}
\toprule
 & \multicolumn{12}{c}{64 views: held variance} \\
\cmidrule(lr){2-13}
 & \multicolumn{3}{c}{\shortstack{deterministic,\\sample var.}} & \multicolumn{3}{c}{\shortstack{deterministic\\+ analytic var.}} & \multicolumn{3}{c}{\shortstack{stochastic\\anchoring}} & \multicolumn{3}{c}{\shortstack{stochastic\\+ analytic var.}} \\
\cmidrule(lr){2-4}\cmidrule(lr){5-7}\cmidrule(lr){8-10}\cmidrule(lr){11-13}
Sampler & F1 & FA & Det$_5$ & F1 & FA & Det$_5$ & F1 & FA & Det$_5$ & F1 & FA & Det$_5$ \\
\midrule
cFM & 30.3 & 99.9 & 47.4 & 62.1 & 7.1 & 66.8 & \textit{56.8} & \textbf{2.6} & \textit{57.2} & \textit{49.3} & \textit{21.5} & \textit{62.1} \\
EDM & 29.9 & 99.7 & 50.1 & 64.8 & 4.4 & 66.9 & \textit{57.8} & \textbf{1.9} & \textit{59.2} & \textit{52.6} & \textit{16.1} & \textit{63.1} \\
DDRM & 27.7 & 98.5 & 53.4 & 61.5 & 6.8 & 65.6 & \textit{56.2} & \textbf{2.2} & \textit{55.5} & \textit{35.8} & \textit{45.4} & \textit{56.3} \\
DDNM & 27.2 & 97.4 & 52.5 & 65.2 & 3.6 & 66.3 & \textit{57.0} & \textbf{1.7} & \textit{57.2} & \textit{35.4} & \textit{47.4} & \textit{57.0} \\
DiffPIR & 27.2 & 97.8 & 52.9 & 65.2 & 1.9 & 65.6 & \textit{53.7} & 1.5 & \textit{56.5} & \textit{51.1} & \textit{16.1} & \textit{59.0} \\
P-DDIM & 28.0 & 98.6 & 54.4 & 64.3 & 3.2 & 65.5 & \textit{54.0} & 1.7 & \textit{56.2} & \textit{43.5} & \textit{28.2} & \textit{57.5} \\
AnchoredFM & 29.6 & 99.3 & 51.5 & 62.8 & 6.0 & 66.6 & \textit{56.0} & \textbf{1.7} & \textit{57.3} & \textit{52.7} & \textit{15.7} & \textit{61.8} \\
\bottomrule
\end{tabular}

}
\end{table*}

\SIfreezecounter{subsection}{2}
\subsection{Data updates from remaining readings}
\label{supp:readings}
Original network conditioning is retained, including images computed from all readings. Additional data steps constrain the free component using $A\Pi=QA$, where $Q=I_m-U_\Hs U_\Hs\T\in\mathbb R^{m\times m}$ projects measurements onto the remaining directions.

Data steps use shared $Qy$ or per-sample $Q(y+\sigma\eta_j)$, with $\eta_j\sim\N(0,I_m)$ independent of acquisition noise and across samples. CT256 uses per-sample readings for both anchor variants at noise level $\sqrt2\sigma$, accounting for acquisition and added noise. MRI, PAT and supplementary CT $\Pi$GDM use shared readings; CT128 retains assumed $\sigma$.

Stochastic CT128 validation gives DDRM and DDNM healthy FA of \res{ct128.readings.shared.fa} for shared readings versus \res{ct128.readings.own.fa} for per-sample readings across cutoffs. CT256 transfers $\kappa=10$ for cFM, EDM and AnchoredFM, and $\kappa=30$ for the four data-step samplers, with healthy-part-0 FA of 1.7--2.5\% and 0.4--2.5\%, respectively.

Using $\sigma$ for perturbed CT256 readings raises DDRM and DDNM median body RMSE from \res{ct.quality.ddrm.base.64} to \res{ct.quality.ddrm.anchored.64}\,HU. Using $\sqrt2\sigma$ gives 56 and 70\,HU, with stochastic data-step FA of 1.5--2.2\% on parts 1--3. This noise correction is empirical for approximate samplers.

\SIfreezecounter{subsection}{3}
\subsection{Reference misregistration}
\label{supp:shifted-reference}
The stochastic CT control displaces the same healthy reference for every sampler using white noise smoothed at 20\,mm standard deviation and scaled to a specified root-mean-square (RMS) shift. At 0.5\,mm, Det$_5$ remains 53.1--56.3\% with FA 2.1--4.7\%; at 1\,mm the ranges are 43.3--47.1\% and 15.0--23.2\%; at 2\,mm they are 26.5--29.9\% and 75.3--84.0\%. Reference error can therefore dominate preserved reconstruction evidence.

\SIfreezecounter{subsection}{4}
\subsection{Reference-free KLIP control}
\label{supp:klip}
The KLIP detection adaptation~\cite{kheirandish2026klip} (\cref{supp:tab:klip}) uses DPS with the same unconditional prior, measurements, operators and 16 samples. The detection procedure is reimplemented with our shared prior. Its 1000-step variance-preserving (VP) readout squares each likelihood update before averaging, following the paper's denoising diffusion probabilistic model (DDPM) convention; released CT code averages before squaring. Anchored MRI includes both the held anchor correction and free-direction DPS guidance.

Labeled part 0 selects guidance, block size and scoring window by Det$_5$, with AUROC breaking ties; healthy part 0 fixes the 95th-percentile threshold. Held-out Det$_5$ recomputes the healthy-test cutoff within each of 2000 paired patient-bootstrap draws; part-0 selection is not resampled. KLIP selects MRI guidance 1, while subtraction DPS selects 3, so these are separately tuned pipelines.

\SIfreezecounter{table}{12}
\begin{table*}[!tp]
\caption{KLIP adaptation, with deterministic MRI anchoring at $\kappa=30$. Test counts are anomalous and healthy images; validation counts are images and patients. Settings give guidance, block side and VP scoring-window fraction. Metrics and patient-bootstrap 95\% intervals are in \%; FA uses the validation-frozen threshold.}
\label{supp:tab:klip}
\centering\scriptsize\setlength{\tabcolsep}{3pt}
\begin{tabular}{lcccrrr}
\toprule
Test & Test/patients & Validation & Setting & AUROC & Det$_5$ & FA$_{\rm val}$ \\
\midrule
MRI, 64 lines & 661/701; 60 & 491/21 & 1, 2, 25--45\% & 83.7 [80.2,87.0] & 55.1 [45.8,63.9] & 6.7 [3.5,10.8] \\
\quad + deterministic & 661/701; 60 & 491/21 & 1, 8, 75--80\% & 93.7 [91.7,95.5] & 71.2 [59.6,79.1] & 4.9 [2.2,8.3] \\
CT, 64 views & 1080/720; 45 & 600/15 & 30, 8, 80--100\% & 70.5 [67.6,73.6] & 32.8 [28.2,36.8] & 4.9 [3.2,6.7] \\
PAT, 64 detectors & 78/78; 27 & 50/9 & 30, 16, 35--40\% & 62.4 [54.8,68.3] & 0.0 [0.0,2.6] & 6.4 [1.4,12.3] \\
\bottomrule
\end{tabular}
\end{table*}

Deterministic MRI KLIP gains are $+10.0$ AUROC points [7.2,13.1] and $+16.2$ Det$_5$ points [5.9,25.1]. The stochastic counterpart gives 93.8\% AUROC, 71.2\% Det$_5$ and 4.6\% FA. With checkerboard seed 0, subtraction DPS gives 75.4\% AUROC and 23.3\% Det$_5$; deterministic anchoring gives 83.7\% and 45.4\%, a within-pipeline Det$_5$ gain of 22.1 points [14.1,28.4]. Both subtraction Det$_5$ estimates remain below unanchored KLIP, with no reported paired cross-pipeline interval. CT and PAT subtraction additionally accesses an exact healthy reference. Reference availability and tuning differ between these pipelines. MRI and CT base KLIP, PAT KLIP and anchored MRI KLIP ran on different GPUs (L40S, H200 and H100), precluding a wall-clock comparison.

\SIfreezecounter{section}{8}
\section{Numerical implementation}
\label{supp:implementation}
CT and PAT compute a dense-operator singular value decomposition (SVD) once in double precision. MRI uses orthonormal Hartley rows, with $U=I_m$ and $s_i=1$. Each prior variance $v_i\T C v_i$ uses one stationary-covariance FFT product. The surrogate measures broad normal-image variability, not anatomical knowledge; degenerate singular subspaces can give solver-dependent held bases, so matched dense-operator experiments use the same GPU type (excluding the KLIP control). Networks run in single precision and projections in double precision. With $T\geq1$ Euler steps, AnchoredFM updates $x\leftarrow x+T^{-1}\Pi v_\theta(x,n/T,\xmap(y))$ for $n=0,\ldots,T-1$; its distinction between training and inference is given in \cref{supp:AnchoredFM-training}.

\SIfreezecounter{section}{9}
\section{AnchoredFM runtime and quality}
\label{supp:efficiency-details}
The stochastic CT runtime study (\cref{supp:tab:runtime-quality}) uses one H100, the 128 grid, 64 views and 16 samples on 432 lesion images from 45 test patients. Contrast error is absolute departure from full lesion recovery; coverage error is departure of mean lesion-pixel coverage from 90\%. The reported trade-off is not monotonic in every metric, and anchored cFM has lower lesion RMSE than one-step AnchoredFM.
Times are median repeated, GPU-synchronized reconstruction calls after network and operator setup and warm-up; setup and training are excluded.
\SIfreezecounter{table}{13}
\begin{table*}[!tp]
\caption{Stochastic CT128 runtime and quality: the AnchoredFM and anchored-cFM budgets plotted in the main paper, full-budget learned comparators, and the Gaussian control. NFE is the number of network evaluations per sample; time is in seconds per 16-sample ensemble. RMSE is in HU, contrast error in \%, and coverage error in percentage points.}
\label{supp:tab:runtime-quality}
\centering\small

\begin{tabular}{lrrrrr}
\toprule
Sampler & NFE & Time (s) & Lesion RMSE & Contrast err. & Coverage err. \\
\midrule
AnchoredFM & 1 & 0.00589 & 65.5 & 26.8 & 9.4 \\
AnchoredFM & 2 & 0.0103 & 65.3 & 26.8 & 9.3 \\
AnchoredFM & 5 & 0.0236 & 65.6 & 26.8 & 6.5 \\
AnchoredFM & 10 & 0.0457 & 65.8 & 27.0 & 2.5 \\
AnchoredFM & 20 & 0.0898 & 66.5 & 27.0 & 1.1 \\
AnchoredFM & 100 & 0.441 & 67.3 & 27.1 & 4.1 \\
\addlinespace[2pt]
cFM + anchoring & 1 & 0.00506 & 59.3 & 28.6 & 18.4 \\
cFM + anchoring & 2 & 0.00942 & 59.3 & 28.5 & 19.5 \\
cFM + anchoring & 5 & 0.0225 & 59.3 & 28.3 & 16.7 \\
cFM + anchoring & 10 & 0.044 & 59.5 & 27.9 & 12.2 \\
cFM + anchoring & 20 & 0.0878 & 59.7 & 27.9 & 7.5 \\
cFM + anchoring & 100 & 0.436 & 60.1 & 27.9 & 2.5 \\
\addlinespace[2pt]
cFM & 100 & 0.411 & 62.9 & 48.4 & 26.4 \\
\addlinespace[2pt]
DDRM + anchoring & 100 & 0.582 & 73.9 & 29.5 & 0.0 \\
\addlinespace[2pt]
DDNM + anchoring & 199 & 1.55 & 127.7 & 32.4 & 14.3 \\
\addlinespace[2pt]
DPS + anchoring & 1000 & 9.83 & 68.1 & 28.7 & 5.3 \\
\addlinespace[2pt]
$\Pi$GDM + anchoring & 100 & 19.5 & 81.9 & 26.1 & 2.5 \\
\addlinespace[2pt]
DiffPIR + anchoring & 1000 & 5.86 & 75.0 & 26.1 & 3.2 \\
\addlinespace[2pt]
P-DDIM + anchoring & 100 & 0.583 & 71.3 & 26.8 & 2.8 \\
\addlinespace[2pt]
EDM + anchoring & 35 & 0.717 & 67.9 & 27.5 & 3.2 \\
Gaussian posterior & 0 & 0.0012 & 61.0 & 28.8 & 4.8 \\
\bottomrule
\end{tabular}

\end{table*}

\SIfreezecounter{section}{10}
\section{Experimental setup details}
\label{supp:setup-details}
\paragraph{Acquisitions.}
All acquisitions are simulated; detection models train on normal images. CT uses LoDoPaB-CT chest slices~\cite{leuschner2021lodopab}, a 256 grid with 1\,mm pixels, 64 parallel-beam views, and Gaussian noise at 1\% of the median central reading. Lesions are specified in \cref{supp:problem}; additional tests use the 128 grid at 64/256 views. Readings are simulated from 362-grid images with a finer projector than reconstruction. Brain Tumor Segmentation (BraTS)/Medical Segmentation Decathlon (MSD) fluid-attenuated inversion recovery (FLAIR) MRI~\cite{antonelli2022msd} uses 256-grid, 1\,mm images at 32/64 of 256 k-space lines (8$\times$/4$\times$ acceleration), noise of a 1\,mm scan, and released real tumors with expert masks. Training uses tumor-free slices of training patients. Virtual Imaging Clinical Trials for Regulatory Evaluation (VICTRE)-family PAT~\cite{park2026pat,badano2018victre} uses 256-grid breast phantoms with 0.66\,mm pixels, initial pressure at 800\,nm, 64 ring detectors and matched healthy twins.

\paragraph{Selection and statistics.}
All methods return 16 samples. MRI has eight paired base and anchored samplers; main CT256 has six, with DPS and $\Pi$GDM in CT128; PAT has cFM, DPS and DDNM. cFM and EDM condition on the surrogate MAP image, while data-step methods share an unconditional prior. Patient part 0 selects settings, with parts 1--3 held out. Base settings minimize lesion error in MRI and PAT and maximize preset F1 in CT; anchored pairs retain them. MRI and PAT select $\kappa$ by labeled lesion error. CT selects the largest $\kappa\in\{1,3,10,30\}$ with at most 5\% healthy-validation FA, or the fewest FA if none qualifies, then transfers settings from 128 to 256 grid. All stochastic CT256 variants remain within 5\% on healthy part 0. Table marks use paired bootstrap $p<0.05$, Holm-adjusted across sampler pairs; text intervals are unadjusted. Practical MRI inference follows \cref{supp:fixed-reference}.

\paragraph{Controls.}
The direct measurement statistic is $T(p)=\langle Aw_p,y-Ar\rangle/(\sigma\|Aw_p\|)$ for $p\in\mathcal P$ with $\|Aw_p\|>0$, using the same Bonferroni family. It is optimal for a known healthy reference, template and position under the Gaussian model. Gaussian-surrogate posterior samples use Matheron's rule~\cite{hoffman1991constrained,wilson2020efficient}, with an additional exact-covariance test. Spread rescaling fits one factor per base sampler on healthy part 0 and freezes it for testing. KLIP's separately tuned adaptation is in \cref{supp:klip}.

\paragraph{Data and ethics.} All experiments use existing public resources: LoDoPaB-CT~\cite{leuschner2021lodopab}, the BraTS FLAIR data distributed with the Medical Segmentation Decathlon~\cite{antonelli2022msd}, and computer-generated breast phantoms~\cite{park2026pat,badano2018victre}. We collected no new data from human participants and had no access to identifying information, so no institutional review was sought for this secondary analysis. Approval and consent for the original acquisitions are described by the dataset providers~\cite{leuschner2021lodopab,antonelli2022msd}. All measurements are simulated, and the method has not been validated for clinical use.

\SIfreezecounter{section}{11}
\section{MRI references and detection protocol}
\label{supp:fixed-reference}
\label{supp:brain-references}
\label{supp:checkerboard-main}

\paragraph{Checkerboard reference.}
On the normalized 256-by-256, 1\,mm grid, the zero-filled inverse Hartley image initializes the minimization of
\[
\begin{aligned}
&\tfrac12\|A_Vx_{\rm TV}-y\|^2\\
&\quad+0.05\sum_{u,v}\sqrt{(D_xx_{\rm TV})_{u,v}^2+(D_yx_{\rm TV})_{u,v}^2+0.01^2}
\end{aligned}
\]
using 400 double-precision Adam~\SIcite{adam} steps at learning rate 0.02 and default moments. Here $V\in\{32,64\}$ counts acquired MRI lines, $x_{\rm TV}$ is the candidate image, and $A_V\in\mathbb R^{256V\times256^2}$ is the main operator $A$ for that Hartley-line acquisition, with measurements $y\in\mathbb R^{256V}$. Indices $u,v\in\{0,\ldots,255\}$ denote row and column. $D_x,D_y$ take forward column and row differences, respectively, with zero differences at the last column or row; the norm is Euclidean. This procedure is applied separately to each measurement setting (noise 0.088) to obtain its initial context.

The grid is partitioned into 16-by-16 squares. The first mask has even $\lfloor u/16\rfloor+\lfloor v/16\rfloor$; the second is its complement. Each of 16 draws inpaints the first mask with the tumor-free unconditional flow prior, then the second using the first-pass output as context. Each pass starts from fresh standard normal noise $\xi$ and takes 100 Euler steps at $t=0,0.01,\ldots,0.99$. Before each velocity evaluation, unmasked pixels are set to $(1-t)\xi+tx_{\rm ctx}$, where $x_{\rm ctx}$ is that pass's context. The state updates by $x_{t+0.01}\leftarrow x_t+0.01v_\theta(x_t,t)$, and the unmasked context is restored exactly at the end. The second pass retains generated first-pass squares. Networks use single precision and state updates double precision.

The mean of 16 second-pass images is the fixed reference shared by all reconstruction samplers. Reference construction uses neither tumor masks nor a tumor finder and treats tumor and healthy slices identically. Results average ten reference seeds (0--9).

\paragraph{Oracle control.}
The oracle reference averages 16 flow inpaintings of the true image inside the expert tumor mask dilated by square radius 16 pixels. A healthy slice is its own reference. This privileged reference nearly saturates AUROC and Det$_5$ even without anchoring; scores also respond to changes in healthy surroundings. FROC additionally counts false marks on tumor slices.

\paragraph{Cohort and settings.}
Part 0 has 491 slices from 21 patients (252 healthy); parts 1--3 have 1362 slices from 60 patients (661 tumor, 701 healthy). Primary comparisons use eight matched samplers with P-DDIM $\eta_0=0$, $\eta_1=16$ across base, deterministic and stochastic treatments. AnchoredFM is compared separately with cFM and excluded from eight-sampler means.

\paragraph{Scores and inference.}
Scores use reconstruction variance times $1+1/K$, $K=16$, plus analytic $\tau^2$ for deterministic anchoring. Reference error remains in the residual. Templates are 8, 16 and 32\,mm discs. The common evaluation mask is derived from skull-stripped truth by thresholding at 0.15, filling holes, eroding four times with cross connectivity, and excluding a four-pixel square neighborhood of the exterior. Each absolute score is normalized by the two-sided Bonferroni $t_{15}$ threshold over tested position--size pairs. Null laws for mixed templates and estimated references remain approximate.

For each sampler and treatment, the 95th-percentile multiplier is fitted to pooled healthy-validation scores over ten reference seeds and frozen for test parts 1--3. The unadjusted Bonferroni rule flags all healthy checkerboard test slices and is not used as a calibrated operating point. Det$_5$ instead uses a retrospective healthy-test cutoff; AUROC ranks global scores. Partial AUC averages localized sensitivity over 201 rates from 0 to 5\% FA. MRI AUROC and slice sensitivity average four predefined tumor-size classes; FROC averages lesions.

Intervals use 2000 paired bootstrap draws of the 60 test patients, with reference seeds resampled within each replicate and shared draws across samplers. Point estimates average ten seeds, with validation thresholds fixed throughout the test bootstrap. Intervals are exploratory and unadjusted. Global spread rescaling changes score scale and preset alarms, not ranking or reconstruction samples.

\SIfreezecounter{subsection}{1}
\subsection{Checkerboard results}
Deterministic anchoring raises mean frozen-threshold sensitivity from 20.2\% to 44.1\% with FA 3.2\% and 3.3\%; matched stochastic anchoring reaches 37.6\% sensitivity. \Cref{supp:tab:practical-fixed64,supp:tab:practical-fixed-contrasts64,supp:tab:checkerboard-size64} provide treatment means, deterministic paired changes and lesion-size results. The deterministic oracle control is reported in \cref{supp:tab:oracle64}. EDM's three overall stochastic intervals and several per-group intervals, including the smallest-group cFM and EDM intervals, include zero (\cref{supp:tab:checkerboard-stochastic-contrasts64,supp:tab:checkerboard-stochastic-size64}).

\SIfreezecounter{table}{14}
\begin{table*}[!tp]
\caption{Stochastic checkerboard MRI comparison at 64 lines with default P-DDIM $\eta_0$: stochastic anchoring minus the matched base sampler, in percentage points with paired patient and seed bootstrap 95\% intervals. AnchoredFM is compared with cFM. These exploratory intervals are not multiplicity-adjusted.}
\label{supp:tab:checkerboard-stochastic-contrasts64}
\centering\footnotesize\setlength{\tabcolsep}{3pt}
\SIfitwidth{\textwidth}{
\begin{tabular}{lrrr}
\toprule
Sampler & $\Delta$ AUROC & $\Delta$ Sens.$_{\rm val}$ & $\Delta$ FROC@0.25 \\
\midrule
cFM & +4.2 [+2.1, +6.2] & +19.1 [+15.6, +22.8] & +4.5 [+2.1, +6.8] \\
EDM & +0.9 [-0.4, +2.3] & -1.1 [-3.6, +1.2] & -0.5 [-2.5, +1.6] \\
DPS & +6.8 [+5.3, +8.4] & +15.6 [+12.1, +18.8] & +8.4 [+5.4, +11.5] \\
$\Pi$GDM & +3.8 [+2.5, +5.2] & +15.7 [+12.1, +19.4] & +3.4 [+1.0, +5.9] \\
DDRM & +6.1 [+4.6, +7.7] & +21.3 [+17.2, +25.6] & +6.1 [+3.6, +8.3] \\
DDNM & +13.3 [+11.3, +15.5] & +35.7 [+29.8, +41.6] & +15.4 [+11.8, +18.8] \\
DiffPIR & +3.8 [+2.6, +5.0] & +7.8 [+4.9, +10.9] & +2.7 [+0.3, +5.0] \\
P-DDIM & +15.1 [+12.6, +17.7] & +35.4 [+29.3, +41.6] & +18.8 [+14.3, +23.3] \\
AnchoredFM & +4.3 [+2.2, +6.4] & +22.2 [+18.5, +25.9] & +6.1 [+3.8, +8.4] \\
Eight-sampler mean & +6.8 [+5.8, +7.8] & +18.7 [+15.6, +21.7] & +7.3 [+5.5, +9.2] \\
\bottomrule
\end{tabular}

}
\end{table*}

\SIfreezecounter{table}{15}
\begin{table*}[!tp]
\caption{Stochastic checkerboard MRI comparison at 64 lines with default P-DDIM $\eta_0$: lesion sensitivity (\%) at 0.25 false-positive marks per slice, by predefined tumor size. Each run uses one global threshold across groups. Base, stochastic anchoring, and their paired differences are averaged over ten reference seeds. Differences are in points with paired patient and seed bootstrap 95\% intervals. The separate cFM--AnchoredFM comparison is excluded from the eight-sampler mean.}
\label{supp:tab:checkerboard-stochastic-size64}
\centering\fontsize{7.5}{8.7}\selectfont\setlength{\tabcolsep}{2.5pt}
\SIfitwidth{\textwidth}{
\begin{tabular}{lrrrrr}
\toprule
Sampler & 11--23\,mm & 23--34\,mm & 34--45\,mm & $\geq$45\,mm & All \\
\midrule
cFM & 9.1 & 34.8 & 55.8 & 72.7 & 40.8 \\
\quad + anchoring & 9.4 & 38.8 & 65.7 & 76.7 & 45.3 \\
\quad difference & +0.4 [-2.7, +3.6] & +4.0 [-0.6, +8.9] & +9.9 [+4.6, +15.1] & +4.0 [+0.3, +8.1] & +4.5 [+2.1, +6.8] \\
EDM & 10.3 & 38.3 & 65.1 & 78.5 & 45.6 \\
\quad + anchoring & 8.7 & 37.1 & 68.4 & 76.1 & 45.1 \\
\quad difference & -1.6 [-4.4, +1.7] & -1.2 [-5.9, +3.7] & +3.3 [-0.0, +6.8] & -2.4 [-5.5, +0.8] & -0.5 [-2.5, +1.6] \\
DPS & 5.1 & 26.9 & 50.2 & 65.0 & 34.6 \\
\quad + anchoring & 7.7 & 36.2 & 62.1 & 75.5 & 43.0 \\
\quad difference & +2.7 [-0.0, +5.3] & +9.3 [+4.7, +14.3] & +11.9 [+6.8, +17.1] & +10.4 [+4.5, +17.2] & +8.4 [+5.4, +11.5] \\
$\Pi$GDM & 6.1 & 32.9 & 61.0 & 76.9 & 41.7 \\
\quad + anchoring & 10.5 & 36.0 & 65.7 & 77.8 & 45.0 \\
\quad difference & +4.4 [+1.6, +7.4] & +3.1 [-2.0, +8.4] & +4.7 [+0.8, +9.1] & +0.8 [-3.2, +5.6] & +3.4 [+1.0, +5.9] \\
DDRM & 6.4 & 27.1 & 54.4 & 74.8 & 38.1 \\
\quad + anchoring & 8.8 & 37.6 & 64.5 & 75.3 & 44.2 \\
\quad difference & +2.4 [-1.0, +5.5] & +10.5 [+5.6, +15.1] & +10.1 [+4.9, +15.7] & +0.4 [-3.9, +4.6] & +6.1 [+3.6, +8.3] \\
DDNM & 4.4 & 20.0 & 40.9 & 58.8 & 29.0 \\
\quad + anchoring & 9.2 & 37.9 & 63.7 & 76.1 & 44.4 \\
\quad difference & +4.7 [+1.5, +7.7] & +17.9 [+12.4, +23.3] & +22.8 [+15.8, +29.2] & +17.3 [+9.4, +25.1] & +15.4 [+11.8, +18.8] \\
DiffPIR & 8.5 & 30.8 & 62.5 & 75.1 & 41.7 \\
\quad + anchoring & 8.8 & 36.8 & 66.3 & 75.6 & 44.5 \\
\quad difference & +0.3 [-2.1, +2.6] & +6.0 [+1.2, +10.2] & +3.8 [-1.1, +8.9] & +0.5 [-4.0, +4.9] & +2.7 [+0.3, +5.0] \\
P-DDIM & 2.7 & 13.7 & 36.1 & 59.3 & 25.7 \\
\quad + anchoring & 9.0 & 38.1 & 62.9 & 78.0 & 44.6 \\
\quad difference & +6.3 [+2.9, +10.3] & +24.4 [+17.9, +29.9] & +26.8 [+17.9, +38.2] & +18.7 [+11.2, +26.1] & +18.8 [+14.3, +23.3] \\
\midrule
cFM & 9.1 & 34.8 & 55.8 & 72.7 & 40.8 \\
AnchoredFM & 11.2 & 40.8 & 67.1 & 77.9 & 46.9 \\
\quad difference & +2.2 [-1.0, +5.7] & +6.0 [+1.4, +10.1] & +11.3 [+5.9, +16.5] & +5.3 [+1.1, +10.1] & +6.1 [+3.8, +8.4] \\
\midrule
Eight-sampler mean & 6.6 & 28.1 & 53.3 & 70.2 & 37.2 \\
\quad + anchoring & 9.0 & 37.3 & 64.9 & 76.4 & 44.5 \\
\quad difference & +2.5 [+0.9, +4.1] & +9.2 [+6.3, +12.1] & +11.7 [+8.3, +15.4] & +6.2 [+3.0, +9.3] & +7.3 [+5.5, +9.2] \\
\bottomrule
\end{tabular}

}
\end{table*}

\SIfreezecounter{table}{16}
\begin{table*}[!tp]
\caption{Matched MRI treatment means (\%) at 32/64 lines over ten reference seeds. Healthy-part-0 thresholds are frozen for testing. P-DDIM uses $\eta_0=0$ in all treatments. AnchoredFM is excluded from these eight-sampler means.}
\label{supp:tab:practical-fixed64}\label{supp:tab:practical-fixed32}
\centering\footnotesize

\begin{tabular}{lrrrrrrr}
\toprule
Lines & Treatment & AUROC & pAUC$_5$ & Det$_5$ & Sens.$_{\rm val}$ & FA$_{\rm val}$ & FROC@0.25 \\
\midrule
32 & base & 67.1 & 5.9 & 11.4 & 8.0 & 2.9 & 21.9  \\
32 & deterministic & 79.2 & 29.2 & 36.8 & 29.5 & 2.3 & 39.4  \\
32 & stochastic & 77.2 & 18.7 & 28.5 & 23.8 & 3.4 & 36.3  \\
64 & base & 78.4 & 14.8 & 26.6 & 20.2 & 3.2 & 38.8  \\
64 & deterministic & 84.6 & 36.4 & 49.0 & 44.1 & 3.3 & 47.5  \\
64 & stochastic & 84.1 & 32.3 & 43.0 & 37.6 & 3.4 & 44.5  \\
\bottomrule
\end{tabular}

\end{table*}
The stochastic-minus-base difference in EDM's frozen-threshold sensitivity at 32 lines is $-3.8$ points $[-6.5,-1.1]$ (\cref{supp:tab:practical-fixed-contrasts64}).

\SIfreezecounter{table}{17}
\begin{table*}[!tp]
\caption{Primary matched checkerboard MRI cohort: frozen-threshold sensitivity differences from each matched base sampler, in points with paired patient and seed bootstrap 95\% intervals. Det. includes analytic held variance. AnchoredFM is compared with cFM.}
\label{supp:tab:practical-fixed-contrasts64}
\centering\footnotesize\setlength{\tabcolsep}{3pt}
\SIfitwidth{\textwidth}{
\begin{tabular}{lrrrr}
\toprule
Sampler & \shortstack{32 lines\\deterministic} & \shortstack{32 lines\\stochastic} & \shortstack{64 lines\\deterministic} & \shortstack{64 lines\\stochastic} \\
\midrule
cFM & +22.2 [+17.7, +26.6] & +19.6 [+15.5, +23.6]   & +23.3 [+18.8, +27.8] & +19.1 [+15.6, +22.8]   \\
EDM & +1.1 [-1.4, +3.4] & -3.8 [-6.5, -1.1]   & +3.9 [+1.5, +6.4] & -1.1 [-3.6, +1.2]   \\
DPS & +20.4 [+15.5, +25.1] & +14.9 [+10.8, +18.8]   & +24.2 [+19.7, +28.8] & +15.6 [+12.1, +18.8]   \\
$\Pi$GDM & +20.9 [+16.3, +25.6] & +14.9 [+11.3, +18.5]   & +21.4 [+17.2, +25.8] & +15.7 [+12.1, +19.4]   \\
DDRM & +19.0 [+14.5, +23.6] & +12.6 [+9.1, +16.2]   & +27.4 [+21.9, +32.6] & +21.3 [+17.2, +25.6]   \\
DDNM & +27.7 [+22.6, +32.9] & +19.2 [+15.4, +23.1]   & +41.2 [+34.6, +47.5] & +35.7 [+29.8, +41.6]   \\
DiffPIR & +30.3 [+24.9, +35.5] & +21.2 [+16.8, +25.6]   & +16.1 [+12.1, +20.0] & +7.8 [+4.9, +10.9]   \\
P-DDIM & +30.1 [+24.6, +35.8] & +27.2 [+22.1, +32.2]   & +33.4 [+27.0, +39.6] & +25.0 [+19.5, +30.3]   \\
AnchoredFM & +21.3 [+17.0, +25.5] & +18.4 [+14.5, +21.9]   & +23.0 [+18.5, +27.4] & +22.2 [+18.5, +25.9]   \\
Eight-sampler mean & +21.5 [+17.7, +25.3] & +15.7 [+12.8, +18.5]   & +23.8 [+19.6, +27.8] & +17.4 [+14.3, +20.4]   \\
\bottomrule
\end{tabular}

}
\end{table*}

\SIfreezecounter{table}{18}
\begin{table*}[!tp]
\caption{Deterministic MRI oracle control, 64 lines, seed 0. Entries are base$\to$anchored (\%); AnchoredFM is compared with cFM. AUROC is 100.0\% throughout and Det$_5$ is 99.9--100.0\%, illustrating oracle saturation. Saved per-slice scores independently confirm 278/701 preset false alarms (39.7\%) for both cFM and DPS; their alarm sets differ.}
\label{supp:tab:oracle64}
\centering\footnotesize\setlength{\tabcolsep}{3pt}
\SIfitwidth{\textwidth}{
\begin{tabular}{lrr}
\toprule
Sampler & FROC@0.25 & FA \\
\midrule
cFM & 65.7$\to$85.8 & 39.7$\to$23.7 \\
EDM & 66.9$\to$83.4 & 14.6$\to$4.4 \\
DPS & 67.1$\to$72.4 & 39.7$\to$2.0 \\
$\Pi$GDM & 66.0$\to$80.8 & 59.6$\to$5.3 \\
DDRM & 69.3$\to$83.1 & 92.9$\to$5.6 \\
DDNM & 67.1$\to$82.4 & 99.7$\to$4.4 \\
DiffPIR & 67.6$\to$86.0 & 98.3$\to$11.7 \\
P-DDIM & 69.8$\to$83.8 & 99.1$\to$14.7 \\
AnchoredFM & 65.7$\to$86.6 & 39.7$\to$49.8 \\
Eight-sampler mean & 67.4$\to$82.2 & 67.9$\to$9.0 \\
\bottomrule
\end{tabular}

}
\end{table*}

\SIfreezecounter{table}{19}
\begin{table*}[!tp]
\caption{Checkerboard MRI with deterministic anchoring at 64 lines: per-method FROC sensitivity at 0.25 false marks per slice by predefined tumor size. Every run shares one global threshold across size groups. The separate cFM--AnchoredFM comparison is excluded from the eight-sampler mean.}
\label{supp:tab:checkerboard-size64}
\centering\fontsize{7.5}{8.7}\selectfont\setlength{\tabcolsep}{2.5pt}
\SIfitwidth{\textwidth}{
\begin{tabular}{lrrrrr}
\toprule
Sampler & 11--23\,mm & 23--34\,mm & 34--45\,mm & $\geq$45\,mm & All \\
\midrule
cFM & 9.1 & 34.8 & 55.8 & 72.7 & 40.8 \\
\quad + anchoring & 11.6 & 45.5 & 68.2 & 79.9 & 49.0 \\
\quad difference & +2.5 [-0.7, +6.0] & +10.7 [+4.7, +16.1] & +12.4 [+7.4, +17.3] & +7.3 [+3.3, +11.6] & +8.2 [+5.7, +10.4] \\
EDM & 10.3 & 38.3 & 65.1 & 78.5 & 45.6 \\
\quad + anchoring & 10.7 & 43.2 & 67.5 & 78.8 & 47.7 \\
\quad difference & +0.5 [-2.9, +4.1] & +4.9 [-0.3, +9.1] & +2.5 [-0.3, +5.2] & +0.3 [-3.5, +3.0] & +2.1 [-0.6, +4.2] \\
DPS & 5.1 & 26.9 & 50.2 & 65.0 & 34.6 \\
\quad + anchoring & 8.8 & 38.4 & 65.2 & 77.9 & 45.1 \\
\quad difference & +3.7 [+1.0, +6.3] & +11.5 [+6.6, +16.2] & +15.0 [+9.9, +20.4] & +12.8 [+6.7, +20.2] & +10.5 [+7.9, +13.3] \\
$\Pi$GDM & 6.1 & 32.9 & 61.0 & 76.9 & 41.7 \\
\quad + anchoring & 9.4 & 42.1 & 66.6 & 78.8 & 46.8 \\
\quad difference & +3.3 [+0.6, +6.0] & +9.2 [+4.7, +14.0] & +5.6 [+2.2, +9.9] & +1.9 [-1.9, +7.1] & +5.2 [+2.7, +8.0] \\
DDRM & 6.4 & 27.1 & 54.4 & 74.8 & 38.1 \\
\quad + anchoring & 10.0 & 41.9 & 67.8 & 79.0 & 47.3 \\
\quad difference & +3.6 [+0.2, +6.8] & +14.8 [+10.0, +19.8] & +13.4 [+8.5, +19.5] & +4.2 [-0.1, +8.4] & +9.2 [+6.5, +11.7] \\
DDNM & 4.4 & 20.0 & 40.9 & 58.8 & 29.0 \\
\quad + anchoring & 10.4 & 43.7 & 67.4 & 79.5 & 47.9 \\
\quad difference & +5.9 [+2.6, +9.8] & +23.7 [+18.1, +29.3] & +26.5 [+19.8, +33.6] & +20.6 [+12.5, +28.2] & +18.9 [+15.5, +22.2] \\
DiffPIR & 8.5 & 30.8 & 62.5 & 75.1 & 41.7 \\
\quad + anchoring & 10.8 & 45.2 & 67.8 & 80.1 & 48.6 \\
\quad difference & +2.3 [-0.2, +5.1] & +14.4 [+9.3, +19.2] & +5.3 [+0.7, +10.4] & +5.0 [+0.9, +8.6] & +6.9 [+4.2, +9.3] \\
P-DDIM & 5.7 & 29.6 & 55.7 & 73.6 & 38.7 \\
\quad + anchoring & 9.4 & 44.4 & 67.5 & 80.1 & 48.0 \\
\quad difference & +3.7 [-0.0, +8.1] & +14.8 [+9.3, +20.4] & +11.8 [+7.5, +17.5] & +6.5 [+1.8, +11.4] & +9.3 [+6.9, +12.0] \\
\midrule
cFM & 9.1 & 34.8 & 55.8 & 72.7 & 40.8 \\
AnchoredFM & 12.5 & 46.0 & 68.7 & 80.4 & 49.6 \\
\quad difference & +3.4 [+0.1, +6.4] & +11.3 [+6.0, +16.3] & +12.9 [+7.8, +17.5] & +7.7 [+3.5, +12.3] & +8.8 [+6.3, +11.0] \\
\midrule
Eight-sampler mean & 6.9 & 30.1 & 55.7 & 71.9 & 38.8 \\
\quad + anchoring & 10.1 & 43.0 & 67.3 & 79.3 & 47.5 \\
\quad difference & +3.2 [+1.0, +5.5] & +13.0 [+9.5, +16.5] & +11.6 [+8.3, +15.2] & +7.3 [+4.3, +10.4] & +8.8 [+6.9, +10.6] \\
\bottomrule
\end{tabular}

}
\end{table*}

\SIfreezecounter{section}{12}
\section{Qualitative examples and diagnostic checks}
\label{supp:qualitative-examples}
\begin{figure*}[!tp]
\centering\includegraphics[width=0.965\textwidth]{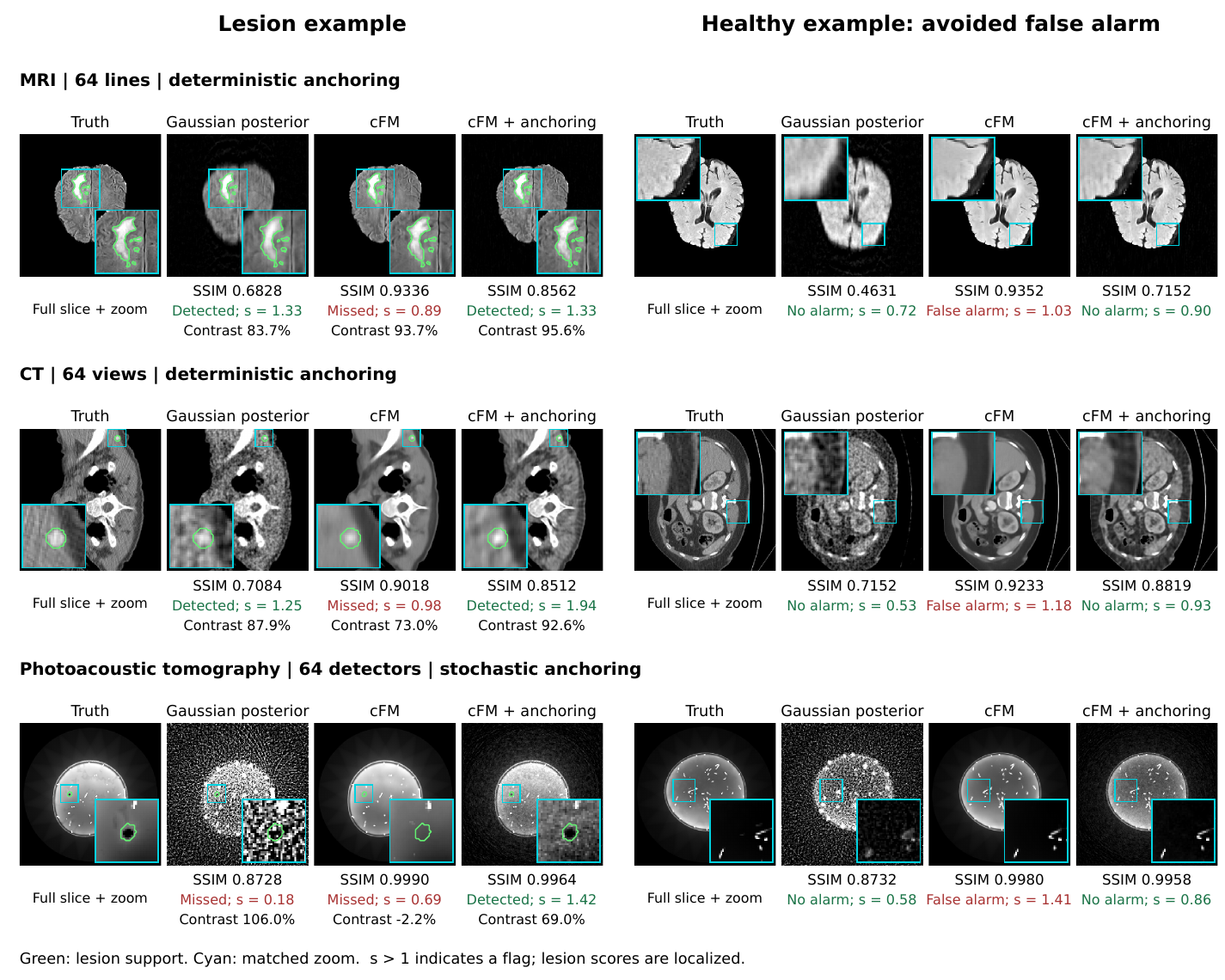}
\caption{Truth and 16-sample means at 64 MRI lines, CT views or PAT detectors. MRI and CT use deterministic anchoring ($\kappa=1$ and 10, respectively); PAT uses stochastic anchoring ($\kappa=30$). Green outlines show lesions; cyan boxes give matched lesion or healthy-score-maximum zooms. Within each case, display windows and zooms are identical. SSIM precedes display windowing. Contrast is tumor-to-ring recovery in MRI or true lesion-profile recovery against the exact healthy reference in CT and PAT. Scores are threshold-normalized maxima near the lesion (support plus four pixels) or over healthy search positions; $s>1$ flags a detection or alarm. MRI thresholds are healthy-validation-frozen over ten seeds; CT and PAT use preset Bonferroni thresholds over position--template pairs. Deterministic scores include analytic held variance.}
\label{supp:fig:qualitative-examples}

\par
\begin{minipage}[t]{0.48\textwidth}
\vspace{0pt}\captionsetup{type=table,skip=2pt}\renewcommand{\arraystretch}{0.88}
\SIfreezecounter{table}{20}
\caption{Source of the SSIM decrease in 64-line MRI (1362 slices). Noise only: the true
image with its held coefficients replaced by the noisy readings. Swap: the unanchored
ensemble mean with the same replacement. Whole-image SSIM, settings as in
\cref{supp:ssim}. For the noise-only image, mean SSIM is 0.990 inside the brain and 0.810 in the background.}
\label{supp:tab:ssim-noise}
\centering\footnotesize\setlength{\tabcolsep}{3pt}
\SIfitwidth{\linewidth}{\begin{tabular}{lrrrrr}
\toprule
Sampler & $\kappa$ & Base & Noise only & Swap & Anchored \\
\midrule
cFM      &   1 & 0.947 & 0.778 & 0.722 & 0.741 \\
EDM      &   1 & 0.954 & 0.778 & 0.738 & 0.749 \\
DPS      & 100 & 0.946 & 0.938 & 0.886 & 0.886 \\
$\Pi$GDM &  30 & 0.953 & 0.895 & 0.850 & 0.855 \\
DDRM     &   3 & 0.946 & 0.819 & 0.773 & 0.782 \\
DDNM     &  10 & 0.944 & 0.857 & 0.812 & 0.814 \\
DiffPIR  &   3 & 0.952 & 0.819 & 0.780 & 0.788 \\
P-DDIM   &  10 & 0.954 & 0.857 & 0.816 & 0.815 \\
\midrule
Mean     &     & 0.950 & 0.843 & 0.797 & 0.804 \\
\bottomrule
\end{tabular}}
\end{minipage}\hfill
\begin{minipage}[t]{0.48\textwidth}
\vspace{0pt}\captionsetup{type=table,skip=2pt}\renewcommand{\arraystretch}{0.88}
\SIfreezecounter{table}{21}
\caption{Coverage (\%) of the held coefficients by nominal 90\% ensemble intervals in
64-line MRI, parts 1--3, with patient-bootstrap 95\% intervals for the base samplers.
Anchored values are from stochastic anchoring; deterministic anchoring uses the
analytic intervals of \cref{supp:held-coverage}.}
\label{supp:tab:held-cover}
\centering\footnotesize\setlength{\tabcolsep}{3pt}
\SIfitwidth{\linewidth}{\begin{tabular}{lrr}
\toprule
Sampler & Base & Anchored \\
\midrule
cFM      & 84.2 [83.8, 84.6] & 90.0 \\
EDM      & 90.6 [90.4, 90.8] & 90.0 \\
DPS      & 76.1 [75.7, 76.5] & 90.0 \\
$\Pi$GDM & 75.3 [75.0, 75.7] & 90.0 \\
DDRM     & 64.3 [64.0, 64.7] & 90.0 \\
DDNM     & 32.0 [31.6, 32.4] & 90.0 \\
DiffPIR  & 66.7 [66.5, 67.0] & 90.0 \\
P-DDIM   & 57.2 [56.7, 57.7] & 90.0 \\
\midrule
Mean     & 68.3 & 90.0 \\
\bottomrule
\end{tabular}}
\end{minipage}

\end{figure*}

The selected lesion examples (\cref{supp:fig:qualitative-examples}) show detections gained through preserved contrast; healthy examples show false alarms removed. They also illustrate lower global SSIM despite better lesion decisions. 

\paragraph{Selection.}
All examples compare Gaussian posterior, cFM and anchored cFM with 16 samples; Gaussian results do not enter selection. MRI uses checkerboard seed 0. Among tumors detected only after anchoring with lower SSIM and contrast error, selection takes median unanchored contrast recovery. Among removed healthy alarms with lower SSIM and all original flags inside brain tissue, it takes median unanchored score. CT follows \cref{supp:teaser}. PAT takes median unanchored score among gained lesion detections or removed healthy alarms with lower SSIM, requiring improved lesion contrast. Even-sized sets use the upper middle case.

\section*{Supplementary references}
\begin{enumerate}
\renewcommand{\theenumi}{S\arabic{enumi}}
\renewcommand{\labelenumi}{[\theenumi]}
\setlength{\itemsep}{3pt}
\setlength{\parskip}{0pt}
\item\label{supp:ref:inversebench}
Hongkai Zheng, Wenda Chu, Bingliang Zhang, Zihui Wu, Austin Wang, Berthy T. Feng, Caifeng Zou, Yu Sun, Nikola Kovachki, Zachary E. Ross, Katherine L. Bouman, and Yisong Yue.
InverseBench: Benchmarking Plug-and-Play Diffusion Priors for Inverse Problems in Physical Sciences.
In \emph{International Conference on Learning Representations}, 2025.
\item\label{supp:ref:ssim}
Zhou Wang, Alan C. Bovik, Hamid R. Sheikh, and Eero P. Simoncelli.
Image Quality Assessment: From Error Visibility to Structural Similarity.
\emph{IEEE Transactions on Image Processing}, 13(4):600--612, 2004.
\item\label{supp:ref:adam}
Diederik P. Kingma and Jimmy Ba. Adam: A Method for Stochastic Optimization.
In \emph{International Conference on Learning Representations}, 2015.
\end{enumerate}

\SIfreezecounter{figure}{4}

\end{document}